\documentclass{article}

\usepackage{arxiv}

\usepackage[utf8]{inputenc} 
\usepackage[T1]{fontenc}    
\usepackage{hyperref}       
\usepackage{url}            
\usepackage{underscore}     
\usepackage{booktabs}       
\usepackage{amsmath}
\usepackage{amsfonts}       
\usepackage{amssymb}
\usepackage{nicefrac}       
\usepackage{microtype}      
\usepackage{graphicx}
\usepackage{xcolor}
\usepackage{colortbl}
\definecolor{toolshade}{RGB}{31,119,180}
\usepackage{subcaption}
\usepackage{caption}
\usepackage{multirow}
\usepackage{array}
\usepackage{pdflscape}      
\usepackage[numbers]{natbib}
\usepackage{doi}

\allowdisplaybreaks

\usepackage{listings}
\lstdefinestyle{prompt}{
    basicstyle=\ttfamily\scriptsize,
    breaklines=true,
    breakautoindent=false,
    breakindent=0pt,
    postbreak={},
    columns=fullflexible,
    keepspaces=true,
    aboveskip=0pt,
    belowskip=0pt,
}
\newenvironment{promptbox}[1]{%
    \par\medskip\noindent\textbf{#1}\par\smallskip\hrule\smallskip
    \begingroup
}{%
    \endgroup\smallskip\hrule\medskip
}

\usepackage[nohyperlinks,nolist]{acronym}
\acrodef{LLM}{Large Language Model}
\acrodef{ML}{Machine Learning}
\acrodef{RAG}{Retrieval-Augmented Generation}
\acrodef{HPC}{High Performance Computing}
\acrodef{MCP}{Model Context Protocol}
\acrodef{MAS}{Multi-Agent System}
\acrodef{FEA}{Finite Element Analysis}
\acrodef{FEM}{Finite Element Methods}
\acrodef{FDM}{Fused Deposition Modeling}
\acrodef{AM}{Additive Manufacturing}
\acrodef{DfM}{Design for Manufacturing}
\acrodef{STL}{Standard Tessellation Language}
\acrodef{MMD}{Maximum Mean Discrepancy}
\acrodef{DPP}{Determinantal Point Process}
\acrodef{RVC}{Ratio of Violated Constraints}
\acrodef{IOG}{Intermediate Optimality Gap}
\acrodef{COG}{Cumulative Optimality Gap}
\acrodef{FOG}{Final Optimality Gap}
\acrodef{IoU}{Intersection over Union}
\acrodef{API}{Application Programming Interface}
\acrodef{cGAN}{conditional Generative Adversarial Network}
\acrodef{SLURM}{Simple Linux Utility for Resource Management}
\acrodef{AHP}{Analytic Hierarchy Process}

\title{EngiAI: Capability-Based Evaluation of Tool-Connected LLM Agents for Engineering Design}

\date{}

\author{
    \href{https://orcid.org/0000-0001-6115-217X}{\includegraphics[scale=0.06]{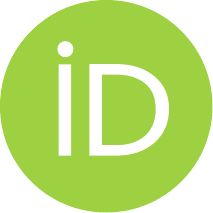}\hspace{1mm}Gioele Molinari}\thanks{Corresponding author.} \\
    IDEAL \\
    Chair of Artificial Intelligence in Engineering Design \\
    ETH Zurich \\
    Zurich, Switzerland \\
    \texttt{gioelemo@ethz.ch} \\
    \And
    \href{https://orcid.org/0000-0002-2874-3645}{\includegraphics[scale=0.06]{orcid.pdf}\hspace{1mm}Florian Felten} \\
    IDEAL, ETH Zurich, Zurich, Switzerland \\
    herve.review, Brussels, Belgium \\
    \And
    \href{https://orcid.org/0000-0002-6763-3625}{\includegraphics[scale=0.06]{orcid.pdf}\hspace{1mm}Soheyl Massoudi} \\
    IDEAL \\
    Chair of Artificial Intelligence in Engineering Design \\
    ETH Zurich \\
    Zurich, Switzerland \\
    \And
    \href{https://orcid.org/0000-0003-3819-8895}{\includegraphics[scale=0.06]{orcid.pdf}\hspace{1mm}Mark Fuge} \\
    IDEAL \\
    Chair of Artificial Intelligence in Engineering Design \\
    ETH Zurich \\
    Zurich, Switzerland \\
}

\renewcommand{\headeright}{Preprint}
\renewcommand{\undertitle}{Preprint}

\hypersetup{
    hidelinks,
    pdftitle={EngiAI: Capability-Based Evaluation of Tool-Connected LLM Agents for Engineering Design},
    pdfsubject={Capability-based evaluation of tool-connected LLM agents for engineering design},
    pdfauthor={Gioele Molinari, Florian Felten, Soheyl Massoudi, Mark Fuge},
    pdfkeywords={engineering agents, agent evaluation, large language models, engineering design, high-performance computing},
}

\begin{document}
\maketitle

\begin{abstract}

Engineering-agent systems are proliferating, but differences in tasks, tools, and success criteria make demonstrations difficult to compare and failures difficult to diagnose. We introduce a capability-based evaluation framework for tool-connected engineering agents. The framework separately evaluates workflow execution, retrieval-assisted parameter selection, high-performance computing (HPC) orchestration, and training-code authoring using execution traces and resulting engineering artifacts. We evaluate four LLM backends on the EngiBench Beams2D and Photonics2D problems using \textsc{EngiAI}, a LangGraph reference implementation. On Beams2D, the two proprietary models complete 96--97\% of workflow tasks, compared with 55--78\% for the two open-source models. Workflows requiring tool-based decision-making perform worse on Photonics2D, reaching 20--53\% completion. Indexed retrieval improves parameter selection. For HPC orchestration, Gemini-3-Flash completes every tested pipeline, whereas GPT-5-mini completes 50--70\% through the final evaluation step. When the agents must write the training code themselves, they fill small gaps reliably but diverge as larger regions are left open. In the open-synthesis tier, both proprietary models select conditional variational autoencoders instead of the reference \ac{cGAN}; Gemini-3-Flash achieves lower maximum mean discrepancy than the reference in seven of ten Beams2D trials. The results support capability-specific evaluation, with separate scores for distinct skills, rather than assessment through successful end-to-end demonstrations alone. The framework provides common evaluation dimensions for identifying failure mechanisms and structuring comparisons of models, agent architectures, and tool interfaces.\end{abstract}


\acresetall

\section{Introduction}
\label{sec:introduction}

Engineering design workflows span problem definition, concept generation, analysis and optimization, and preparation for manufacturing~\cite{pahl2007engineering,wynn2018process}. These stages are usually supported by separate software tools and require users to transfer data and decisions between them. Deep generative models support engineering design-space exploration and optimization, including quality- and diversity-aware generation~\cite{chen2021padgan} and design under manufacturing uncertainty~\cite{chen2023ganduf}. Their use adds further workflow steps, including data preparation, model training, and evaluation~\cite{regenwetter2022deep}.

\acp{LLM} agents can coordinate external tools from natural-language instructions by selecting an appropriate tool, generating its arguments, observing its output, and using that output in subsequent decisions~\cite{yao_react_2023,schick_toolformer_2023}. However, recent benchmarks report substantial failures in multi-step tool use and scientific workflows~\cite{yao_tau-bench_2024,chen_scienceagentbench_2025}. It therefore remains unclear whether current agents can perform these operations reliably across complete engineering workflows, and for which steps they break down.

LLM-agent systems are being proposed for a growing range of engineering activities, including conceptual design, simulation, optimization, manufacturing, and design coordination~\cite{ni_mechagents_2024,pradasgomez_ductile_2026,chen_designagent_2026,massoudi_agentic_2025}. However, these systems generally use different engineering problems, tool interfaces, prompts, stopping conditions, and success criteria. A successful end-to-end demonstration therefore does not reveal whether performance arose from the underlying LLM, the orchestration logic, retrieved information, the tool interface, or the engineering program being executed. These differences also prevent direct comparison across agent systems and make it difficult to translate individual demonstrations into general development guidance.

Here, we use the term \emph{agent harness} for the components that connect an LLM to an engineering workflow: prompts, tool schemas, routing and state management, retrieval, execution control, and output validation. Existing benchmarks evaluate several of these components in general-purpose settings, but they do not establish how their failures propagate into engineering artifacts, remote training jobs, or trained models. A useful engineering-agent evaluation must therefore assess both the execution trace and the resulting engineering outcome, while separating the capabilities required at different stages of the workflow.

This paper introduces a benchmark suite for evaluating \acs{LLM}-driven engineering workflows, together with \textsc{EngiAI}, a multi-agent reference implementation that operationalizes it. Accessible through a web-based interface, \textsc{EngiAI} coordinates multiple engineering tools through natural-language interaction, with a modular architecture that allows new capabilities to be added via additional agents or tool APIs. The benchmark suite and reference implementation together address the following research question:

\vspace{-.4cm}
\begin{quote}\itshape
How can tool-connected \acs{LLM} agents be evaluated by capability, and what do these evaluations reveal about failures in workflow execution, document-grounded parameter selection, remote training-job orchestration, and training code authoring?
\end{quote}

We evaluate these four capabilities separately so that failures can be attributed to a specific part of the workflow. This paper makes four contributions. First, we formulate an evaluation framework that separates four capabilities of tool-connected engineering agents: workflow execution, retrieval-assisted parameter selection, remote training-job orchestration, and training-code authoring. Second, we instantiate this framework as a benchmark that combines controlled prompt variations and retrieval ablations with capability-specific metrics, scoring both tool traces and resulting engineering artifacts. Third, we apply the benchmark to four LLM backends and two engineering problems, identifying failures in semantic grounding, conditional decisions, long-horizon completion, checkpoint compatibility, sampling behavior, and trained-model quality. Fourth, we provide \href{https://github.com/gioelemo/EngiAI-Public}{\textsc{EngiAI}} as an open reference implementation for comparing model backends within its multi-agent architecture. Applying the same tasks and scorers to another agent architecture requires a compatible tool, trace, and artifact interface (\href{https://youtu.be/QbQVZFCq3X0}{video demonstration}).

\section{Related Work}
\label{sec:relatedwork}

Pahl and Beitz~\cite{pahl2007engineering} organize engineering design into task clarification, conceptual design, embodiment design, detail design, and preparation of production documentation. Other models represent design and development processes differently~\cite{wynn2018process}, but iteration is common across them: analysis, testing, and new information can require earlier decisions to be revisited~\cite{wynn2017perspectives}. Decision-based design provides a complementary view in which design involves selecting alternatives under objectives, constraints, and uncertainty~\cite{hazelrigg1998framework}.

\acs{ML}-augmented design adds infrastructure for training surrogate and generative models, as provided by systems such as EngiOpt~\cite{felten_engibench_2025}. It also introduces coordination among heterogeneous software tools, which Jiang \textit{et al.}~\cite{jiang_intelligent_2025} describe within their Intelligent Design~4.0 framework.

The following subsections review related work on multi-agent orchestration, engineering-tool integration and manufacturing, retrieval-augmented engineering, agent benchmarking, and machine-learning code authoring.

\subsection{Multi-agent orchestration}
Recent surveys describe the use of LLM-based agents for multi-step tasks across scientific domains~\cite{acharya_agentic_2025,gridach_agentic_2025}. General-purpose frameworks such as AutoGen~\cite{wu_autogen_2024}, CrewAI~\cite{moura_crewai_2024}, the OpenAI Agents SDK~\cite{openai_agents_sdk_2025}, and LangGraph~\cite{noauthor_langchain-ailanggraph_2026} support communication and tool routing among multiple agents. The AI Co-Scientist applies structured agent coordination to research tasks~\cite{gottweis_towards_2025}. Jiang \textit{et al.}~\cite{jiang_intelligent_2025} place such systems within their Intelligent Design~4.0 framework for multi-agent design automation.

Engineering applications have used agent systems for topology optimization and finite-element analysis~\cite{ni_mechagents_2024,qi_feagpt_2025,deotale_all-fem_2026}, aerospace structural analysis~\cite{pradasgomez_ductile_2026}, automotive and product design~\cite{elrefaie_ai_2025,chen_designagent_2026}, material selection~\cite{ying2026aligning}, iterative battery-pack design~\cite{xu2026supervising}, mechatronics~\cite{wang_llm-enabled_2025}, and design-state coordination across \acs{LLM} backends~\cite{massoudi_agentic_2025}. TopOptAgents~\cite{park2026selfrefiningtopologyoptimizationllmbased} coordinates six agents to formulate, execute, and refine a topology-optimization problem. These systems address specific design, analysis, or coordination tasks. Our experiments additionally evaluate document retrieval, remote training-job management, and training-code authoring within one tool-connected workflow.

\subsection{Tool integration and manufacturing}
Tool interfaces determine which engineering operations an agent can perform. The \ac{MCP} provides a common interface for external services~\cite{anthropic_model_2025}, although tool-selection accuracy declines as tasks and tool sets become more complex~\cite{wang_mcp-bench_2025}. Jadhav and Farimani~\cite{jadhav_llm3dprint_2025} use a hierarchical agent system to monitor fused-deposition-modeling processes. EngiBench and EngiOpt provide Python interfaces for engineering problems, optimization, and generative design~\cite{felten_engibench_2025}. \textsc{EngiAI} connects these interfaces so that the benchmark can evaluate both the selected tools and the engineering outputs they produce.

\subsection{Retrieval-augmented engineering}
\acs{RAG} supplies an \acs{LLM} with passages retrieved from an external document collection, reducing its reliance on parametric knowledge alone~\cite{lewis_retrieval-augmented_2021,gao_retrieval-augmented_2024}. This is particularly relevant when an engineering workflow requires numerical parameters from technical documents, because an incorrect value can produce an invalid or misleading design. Chandrasekhar \textit{et al.}~\cite{chandrasekhar_amgpt_2024} combine Llama2-7B with indexed \ac{AM} papers and report performance competitive with GPT-4 on metal additive-manufacturing questions. Khanghah \textit{et al.}~\cite{khanghah_zero-shot_2025} apply multimodal \acs{RAG} to zero-shot anomaly detection in laser powder bed fusion and report a 12\% improvement in classification accuracy over non-retrieval baselines. These studies evaluate question answering or classification. They do not examine whether an agent can retrieve numerical parameters and use them correctly in subsequent simulation or design-tool calls. Our \acs{RAG} experiment evaluates this latter task.

\subsection{Agent benchmarking}
Surveys of agent evaluation distinguish capabilities such as tool selection, argument accuracy, multi-step planning, and task completion~\cite{mohammadi_evaluation_2025}. ACEBench~\cite{chen_acebench_2025}, ToolSandbox~\cite{lu_toolsandbox_2025}, BFCL~\cite{patil_gorilla_2024}, and ToolBench~\cite{qin_toolllm_2024} focus primarily on tool calls and report performance differences between open-source and proprietary models. AgentBench~\cite{liu_agentbench_2025}, AgentBoard~\cite{ma_agentboard_2024}, and ScienceAgentBench~\cite{chen_scienceagentbench_2025} evaluate longer reasoning and execution sequences across multiple environments, primarily through single-agent tasks. For multi-turn professional tasks, $\tau$-bench~\cite{yao_tau-bench_2024} reports task-success rates below 50\% even for state-of-the-art models.

Engineering-specific benchmarks address more specialized capabilities. FDM-Bench~\cite{eslaminia_fdm-bench_2024} evaluates \acsp{LLM} on fused-deposition-modeling tasks, while Zhou \textit{et al.}~\cite{zhou_engibench_2025} introduce a hierarchical engineering benchmark and show that current models struggle with open-ended modeling problems. These benchmarks cover important aspects of tool use, reasoning, and engineering knowledge, but they do not jointly evaluate a multi-agent workflow that combines simulation, document retrieval, and remote \acs{HPC} execution.

ScienceAgentBench is especially relevant because it decomposes scientific workflows into executable tasks and evaluates generated programs, execution results, and cost~\cite{chen_scienceagentbench_2025}. EngDesign evaluates functional engineering outputs through simulation across nine engineering domains~\cite{guo_toward_2025}. These studies establish task-level scientific evaluation and simulation-based engineering verification, respectively. Our benchmark examines a different combination: failures propagated across workflow execution, document-grounded parameter selection, remote training-job management, and training-code authoring.

Our benchmark adds three requirements that arise when these capabilities are connected. First, numerical outputs from one tool must be used correctly in later operations, such as selecting an export branch from a simulation result. Second, the agent must distinguish between competing plausible parameter values and preserve their numerical precision. Third, it must retain separate parameter sets across sequential tool calls. We evaluate these behaviors using both the tool-call trace and the resulting engineering artifact or trained model.

\subsection{Machine-learning engineering agents}
Software-engineering benchmarks such as SWE-bench evaluate whether agents can resolve issues in existing repositories~\cite{jimenez2024swebench}. A related group of benchmarks evaluates agents that write and revise machine-learning code. MLAgentBench tests iterative modifications to training programs and reports near-zero success on its more recent datasets~\cite{3692070.3692884}. MLE-bench evaluates agents through Kaggle competitions~\cite{chan2025mlebench}, while MLGym and RE-Bench cover more open-ended machine-learning research tasks~\cite{nathani2025mlgymnewframeworkbenchmark,wijk2025rebench}. MLGym finds that agents can tune hyperparameters more reliably than they can implement model architectures that differ from the supplied baseline. The AI Scientist generates and iterates on complete research ideas, but evaluates the resulting paper with an automated reviewer rather than evaluating the engineering output of the trained model~\cite{lu_ai_2024}. AutoML-Agent instead coordinates specialist \acs{LLM} agents across stages of an AutoML workflow~\cite{trirat2025automlagent}.

These studies show that agents can be evaluated through the code they produce and, in several cases, through the performance of the resulting model. Our benchmark applies this approach to generative models of engineering designs. It separates training-code authoring into progressively more open tasks. The configuration tier modifies parameters in an otherwise fixed program; the code-completion tiers require selected regions to be implemented; and the open-synthesis tier supplies only the data and evaluation interfaces, allowing the agent to select and implement a model architecture. Here, ``open synthesis'' means selecting an architecture that is not prescribed by the supplied reference program; it does not imply an algorithm that is novel to the research literature.

The authored program must then be submitted to an \acs{HPC}/\acs{SLURM} cluster, complete training, and produce a model whose generated designs are scored by a functional engineering evaluator. The benchmark therefore measures both code authoring and completion of the remote training workflow. Among the studies reviewed above, this combination of progressively open training-code authoring, functional evaluation of generated engineering designs, and cluster-level execution is not otherwise assessed.

The reviewed studies cover different parts of tool-connected engineering workflows. \textsc{EngiAI} implements seven system capabilities: multi-agent orchestration, physics simulation and topology optimization, document retrieval, manufacturing export, remote \acs{HPC} orchestration, training-code authoring, and structured benchmarking. We organize the experiments into four evaluation dimensions: workflow execution, retrieval-assisted parameter selection, \acs{HPC} orchestration, and training-code authoring. Each evaluation examines both the execution trace and the resulting engineering artifact. Table~\ref{tab:literature_comparison} compares \textsc{EngiAI} with the systems and benchmarks reviewed above.

\begin{table}[ht]
\centering
\caption[Comparison of related work.]{Representative selection of \acs{LLM}-relevant engineering capabilities covered by \textsc{EngiAI} versus prior work, grouped into \textit{Capabilities}, \textit{Infrastructure}, and \textit{Evaluation}. \checkmark\ = supported, $\times$ = not supported. Systems are split into engineering agent systems (top) and evaluation benchmarks (bottom).}
\label{tab:literature_comparison}
\footnotesize
\setlength{\tabcolsep}{3pt}
\resizebox{\columnwidth}{!}{%
\begin{tabular}{@{}l c c c c c c c@{}}
\toprule
 & \multicolumn{5}{c}{\textbf{Capabilities}}
 & \textbf{Infra.}
 & \textbf{Eval.} \\
\cmidrule(lr){2-6}\cmidrule(lr){7-7}\cmidrule(lr){8-8}
 & \rotatebox{70}{Multi-Agent}
 & \rotatebox{70}{RAG}
 & \rotatebox{70}{Simulation}
 & \rotatebox{70}{Mfg.\ Export}
 & \rotatebox{70}{Code Gen.}
 & \rotatebox{70}{HPC}
 & \rotatebox{70}{Benchmark} \\
\midrule
\multicolumn{8}{@{}l}{\textit{Engineering Agent Systems}} \\[1pt]
DSG-MAS \cite{massoudi_agentic_2025}
  & \checkmark & $\times$   & \checkmark & $\times$   & \checkmark & $\times$ & \checkmark \\
MechAgents \cite{ni_mechagents_2024}
  & \checkmark & $\times$   & \checkmark & $\times$   & \checkmark & $\times$ & $\times$ \\
FEAGPT \cite{qi_feagpt_2025}
  & \checkmark & \checkmark & \checkmark & $\times$   & \checkmark & $\times$ & $\times$ \\
LLM3DPrint \cite{jadhav_llm3dprint_2025}
  & \checkmark & $\times$   & $\times$   & \checkmark & $\times$ & $\times$ & $\times$ \\
DUCTILE \cite{pradasgomez_ductile_2026}
  & $\times$   & $\times$   & \checkmark & $\times$   & \checkmark & $\times$ & \checkmark \\
AMGPT \cite{chandrasekhar_amgpt_2024}
  & $\times$   & \checkmark & $\times$   & $\times$   & $\times$ & $\times$ & $\times$ \\
Wang et al.\ \cite{wang_llm-enabled_2025}
  & \checkmark & $\times$   & \checkmark & $\times$   & \checkmark & $\times$ & $\times$ \\
DesignAgent \cite{chen_designagent_2026}
  & \checkmark & \checkmark & \checkmark & $\times$ & $\times$ & $\times$ & $\times$ \\
ALL-FEM \cite{deotale_all-fem_2026}
  & \checkmark & \checkmark & \checkmark & $\times$ & \checkmark & $\times$ & \checkmark \\
TopOptAgents \cite{park2026selfrefiningtopologyoptimizationllmbased}
  & \checkmark & $\times$ & \checkmark & $\times$ & \checkmark & $\times$ & $\times$ \\
\midrule
\multicolumn{8}{@{}l}{\textit{Evaluation Benchmarks}} \\[1pt]
EngiBench/EngiOpt \cite{felten_engibench_2025}
  & $\times$   & $\times$   & \checkmark & $\times$   & $\times$ & $\times$ & \checkmark \\
EngiBench \cite{zhou_engibench_2025}
  & $\times$   & $\times$   & $\times$   & $\times$   & $\times$ & $\times$ & \checkmark \\
FDM-Bench \cite{eslaminia_fdm-bench_2024}
  & $\times$   & $\times$   & $\times$   & $\times$   & $\times$ & $\times$ & \checkmark \\
EngDesign \cite{guo_toward_2025}
  & $\times$ & $\times$ & \checkmark & $\times$ & $\times$ & $\times$ & \checkmark \\
General benchmarks \cite{chen_acebench_2025,lu_toolsandbox_2025,patil_gorilla_2024,qin_toolllm_2024,wang_mcp-bench_2025,liu_agentbench_2025,ma_agentboard_2024,chen_scienceagentbench_2025,yao_tau-bench_2024,he_webvoyager_2024,deng_mind2web_2023}
  & $\times$   & $\times$   & $\times$   & $\times$   & \checkmark & $\times$ & \checkmark \\
Code-authoring benchmarks~\cite{nathani2025mlgymnewframeworkbenchmark,3692070.3692884,chan2025mlebench,jimenez2024swebench,trirat2025automlagent,wijk2025rebench}
  & \checkmark & \checkmark & $\times$ & $\times$ & \checkmark & $\times$ & \checkmark \\
\midrule
\textbf{\textsc{EngiAI} (ours)}
  & \checkmark & \checkmark & \checkmark & \checkmark & \checkmark & \checkmark & \checkmark \\
\bottomrule
\end{tabular}%
}
\end{table}


\section{Methods}
\label{sec:methods}

\subsection{System Architecture}

\begin{figure}
    \centering
    \includegraphics[width=\linewidth]{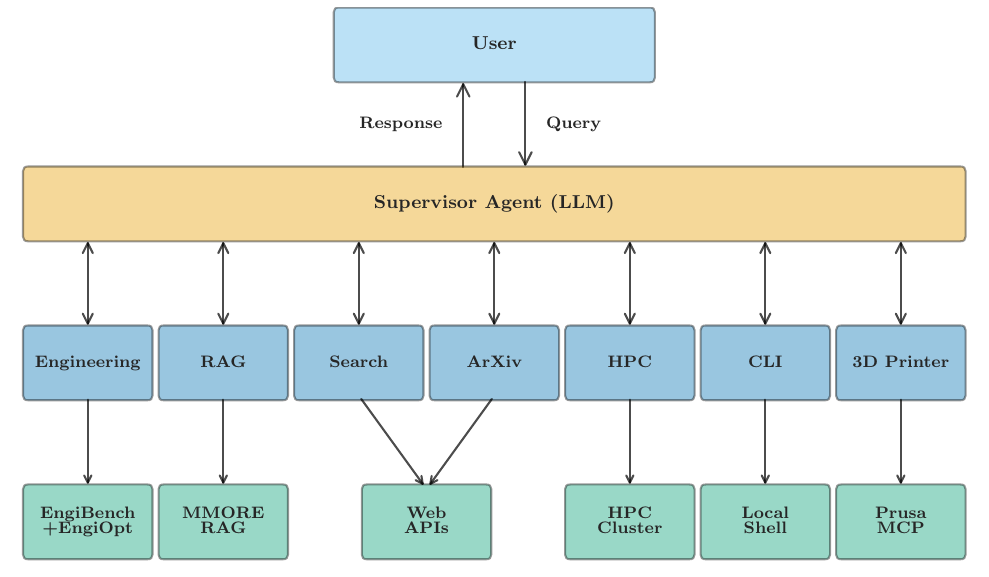}
    \caption[Multi-agent architecture]{Multi-agent architecture. From top to bottom: the user interface, the orchestration layer (supervisor agent routing prompts), the specialization layer (seven domain agents), and the execution layer (external services and tools).}
    \label{fig:multiagentarch}
\end{figure}

The \textsc{EngiAI} system follows a hierarchical supervisor pattern (Figure~\ref{fig:multiagentarch}). A central supervisor receives each user prompt, classifies its intent, and routes it to the appropriate specialist agent. Each specialist is implemented as a LangGraph state machine that receives a task, invokes its domain-specific tools, and returns the result to the supervisor.

We use this decomposition to limit the number of tools exposed to each specialist. This design choice is motivated by evidence that \acsp{LLM} become less accurate at tool selection as the number of available tools increases~\cite{wang_mcp-bench_2025}. The present study does not compare this supervisor architecture with a single agent having access to the complete tool set.

The supervisor can invoke several specialists sequentially. For example, it can first route a request to the \acs{RAG} agent to retrieve parameters and then pass those parameters to the engineering agent. LangGraph stores the workflow state, binds tools to agents, supports human-interruption points, and persists checkpoints~\cite{chase_langchain_2022}. \textsc{EngiAI} terminates a workflow when it detects repeated rerouting.

A typical design session illustrates the pattern: when a user requests a topology optimization, the supervisor routes to the \emph{engineering agent}, which calls \emph{create\_problem} $\to$ \emph{optimize\_design} $\to$ \emph{simulate\_design} $\to$ \emph{render\_design} (or \emph{convert\_design\_to\_stl} for manufacturing export). If the user first asks about a parameter from a paper, the supervisor routes to the \emph{\acs{RAG} agent}, which retrieves the relevant information, then re-routes to the \emph{engineering agent} with the extracted values.

\subsection{Agent Capabilities}

Table~\ref{tab:agent_overview} summarizes the seven specialist agents and their tools. We organize them by their role in the design-to-manufacturing lifecycle.

\paragraph{Engineering agent.} The core design agent drives topology optimization, simulation, \acs{ML}-based inverse design, and \acs{STL} export through the EngiBench/EngiOpt~\cite{felten_engibench_2025} stack. EngiBench provides a unified Python \acs{API} for accessing engineering design problems across multiple physics domains; EngiOpt extends it with \acs{ML}-based generative design algorithms (\textit{e.g.}, \ac{cGAN}s, diffusion models) that can be trained and evaluated against dataset baselines. Both model families are established for engineering design: \acsp{cGAN}~\cite{mirza2014conditionalgenerativeadversarialnets} have been applied to structural topology optimization~\cite{10.1115/1.4049533}, and diffusion models were subsequently shown to outperform them on performance- and manufacturability-aware topology generation~\cite{Mazé_Ahmed_2023}.

\paragraph{Information retrieval agents.} Three agents provide complementary retrieval capabilities. The \emph{\acs{RAG} agent} and \emph{ArXiv agent} share the MMORE (Multimodal Open Retrieval Engine)~\cite{sallinen_mmore_2025} framework, which indexes and retrieves information from heterogeneous document types (PDFs, images, tables)\textemdash the former for document Q\&A over user-indexed collections, the latter for fetching and indexing papers from arXiv. The \emph{search agent} provides web search via the Tavily \acs{API}.

\paragraph{Infrastructure agents.} The \emph{\acs{HPC} agent} manages \acs{SLURM} jobs on remote clusters via SSH. The \emph{CLI agent} executes local shell commands (with a whitelisted set of GUI applications and human-in-the-loop confirmation step before running commands), and the \emph{3D printer agent (Prusa)} controls 3D printers through a dedicated \acs{MCP}~\cite{anthropic_model_2025} server.

\begin{table}[ht]
\centering
\caption{\textsc{EngiAI} multi-agent architecture and tool integration.}
\label{tab:agent_overview}
\footnotesize
\begin{tabular}{llc}
\toprule
\textbf{Agent} & \textbf{Primary Function} & \textbf{\# Tools} \\
\midrule
Supervisor   & Request routing              & -- \\
Engineering  & Optimization, ML, STL export & 14 \\
Search       & Web retrieval                & 1 \\
RAG          & Document Q\&A                & 5 \\
ArXiv        & Paper retrieval              & 5 \\
HPC          & Remote job submission        & 8 \\
CLI          & Command execution            & 2 \\
3D Printer (Prusa MCP)  & 3D printer control           & 8 \\
\bottomrule
\end{tabular}
\end{table}

\subsection{Experimental Setup}

\subsubsection{Evaluation Problems}

We evaluate on two EngiBench~\cite{felten_engibench_2025} problems spanning different physics domains.

\textit{Beams2D.} A $50 \times 100$ cantilever beam topology optimization problem minimizing compliance subject to a volume fraction constraint~\cite{sigmund_99_2001, andreassen_efficient_2011}. This well-studied structural optimization benchmark provides ground-truth solutions for evaluating agent correctness. Key parameters are volume fraction (\emph{volfrac}), force distance (\emph{forcedist}), and filter radius (\emph{rmin}).

\textit{Photonics2D.} A $120 \times 120$ photonic device inverse design problem maximizing electromagnetic field overlap for specified wavelengths, testing whether agent capabilities transfer across physics domains.

\subsubsection{Models and Configuration}

We evaluate two proprietary models, GPT-5-mini (\texttt{gpt-5-mini-2025-08-07})~\cite{singh2025openaigpt5card} and Gemini-3-Flash (\texttt{gemini-3-flash-preview}), and two open-source 4B-parameter models served through Ollama, Qwen3-4B (\texttt{qwen3:4b-instruct-2507-q8_0})~\cite{yang_qwen3_2025} and Qwen3.5-4B (\texttt{qwen3.5:4b-q8_0}). This selection permits a comparison between proprietary and locally served models and between two Qwen generations at the same parameter count. All models support structured tool calls. We set the sampling temperature to zero and the API seed to 42. Each workflow model--style cell contains 15 runs: three random seeds and five dataset samples. Beams2D uses all seven prompt styles; Photonics2D uses three. Appendix~\ref{app:compute} reports the compute and inference settings.

\subsection{Benchmark Design}
The benchmark addresses four questions: (1)~\emph{workflow execution}\textemdash can agents reliably chain engineering tools when prompts impose different cognitive demands? (2)~\emph{retrieval-augmented parameter selection}\textemdash does document retrieval improve the selection of engineering parameters? (3)~\emph{\acs{HPC} orchestration}\textemdash can agents submit, monitor, and evaluate long-running training jobs on remote infrastructure? and (4)~\emph{training-code authoring}\textemdash can agents write and execute training code rather than only run a supplied program? Sections~\ref{sec:workflow_eval}--\ref{sec:tc_eval} define the tasks and success criteria for each evaluation.

Table~\ref{tab:evaluation_framework} summarizes how each evaluation is conducted, what evidence it records, and which types of failure it can identify.

\begin{table*}[t]
\centering
\caption[Capability-based evaluation framework]{
Operationalization of the capability-based evaluation framework. Each evaluation varies the information, instructions, or implementation responsibility given to the agent and collects evidence needed to distinguish execution, grounding, interface, and outcome failures.
}
\label{tab:evaluation_framework}
\footnotesize
\setlength{\tabcolsep}{4pt}
\renewcommand{\arraystretch}{1.15}

\begin{tabular}{@{}
p{0.16\textwidth}
p{0.28\textwidth}
p{0.27\textwidth}
p{0.23\textwidth}
@{}}
\toprule
\textbf{Evaluation dimension}
&
\textbf{Experimental contrast}
&
\textbf{Evidence collected}
&
\textbf{Failures distinguished}
\\
\midrule

Workflow execution
&
Seven prompt styles vary explicitness, ambiguity, distractors, derived parameters, conditional decisions, and retention of multiple parameter sets.
&
Expected and observed tool calls, task completion, parameter validation, and the resulting engineering design.
&
Missing or redundant calls, failure to request clarification, incorrect numerical values, branch inversion, and interference between parameter sets.
\\

Retrieval-assisted parameter selection
&
\acs{RAG}-on and Empty \acs{RAG} hold tool access constant while varying indexed document content. \acs{RAG}-off separately verifies behavior without retrieval access.
&
Retrieval invocation, selected parameter values, and the values passed to downstream design-tool calls.
&
Failure to retrieve, reliance on unsupported defaults, incomplete parameter sets, and incorrect values passed downstream.
\\

\acs{HPC} orchestration
&
Explicit prompts enumerate the required operations, whereas natural prompts require the agent to infer the remote training workflow.
&
\acs{SLURM} configuration, script acceptance, job submission, monitoring, final evaluation, and extracted model metrics.
&
Invalid configurations, failed submission or monitoring, premature workflow termination, and omission of the final evaluation.
\\

Training-code authoring
&
T1--T6v2 progressively increase the code authored by the agent. Canonical and functional evaluators separate reference-interface compatibility from functional model behavior.
&
Syntax and structural validation, local dry run, \acs{SLURM} acceptance, job completion, checkpoint loading, sampler behavior, and \acs{MMD}.
&
Syntax or interface errors, execution failure, checkpoint incompatibility, degenerate or condition-insensitive sampling, and poor trained-model quality.
\\

\bottomrule
\end{tabular}
\end{table*}

\subsubsection{Workflow Evaluation}
\label{sec:workflow_eval}

We define seven prompt styles spanning distinct cognitive demands (Table~\ref{tab:workflow_prompts}; representative prompts in Appendix~\ref{app:prompts}). The first two establish baselines: \textsc{Full} provides all numerical parameters for a direct optimize$\to$simulate$\to$render pipeline, while \textsc{Natural} uses qualitative descriptions (\textit{e.g.}, ``lightweight'' instead of $\text{volfrac}=0.2$), requiring the agent to detect ambiguity and request human clarification.

The five W-styles extend \textsc{Full} with an \acs{STL} export step. Each modifies only the export instructions while keeping the optimization and simulation steps identical, isolating the effect of instruction phrasing on task completion. The styles target three categories of failure mode that standard tool-calling benchmarks do not capture.

The first category is \emph{numerical fidelity}: can the agent propagate precise values from prompt to tool call? \textsc{W-Rand} tests this directly with explicit randomized export parameters (density threshold, XY scale, Z extrusion, mirror flag). \textsc{W-Derived} raises the bar by specifying parameters as derivation rules (\textit{e.g.}, ``use the volume fraction as the density threshold''), requiring arithmetic over values the agent passed to earlier tools.

The second is \emph{semantic grounding}: can the agent select the correct value when the prompt contains competing alternatives? \textsc{W-Distract} presents two plausible values for the same parameter\textemdash one in a preview context, one for manufacturing export\textemdash both of which the tool accepts without error. Only the parameter-validation scorer distinguishes correct from incorrect choices, making this failure mode invisible to benchmarks that check only function signatures.

The third is \emph{stateful reasoning}: can the agent condition its actions on runtime information? \textsc{W-Cond} requires the agent to read a compliance value from \emph{simulate\_design} and select the corresponding parameter branch via an if/then rule. \textsc{W-Multi} requires two sequential export calls with entirely different parameter sets, testing whether the agent can maintain distinct configurations in working memory rather than merging or dropping one.

\begin{table}[ht]
\centering
\caption[Workflow evaluation prompt styles.]{Workflow evaluation prompt styles. \textsuperscript{$\dagger$}Randomized STL parameters (\emph{mirror\_y}, \emph{scale\_xy}, \emph{scale\_z}, \emph{threshold}); validated within $\pm 0.05$ absolute tolerance on each parameter value to account for \ac{LLM} rounding. \textsuperscript{$\ddagger$}STL parameters derived from optimization inputs.}
\label{tab:workflow_prompts}
\footnotesize
\begin{tabular}{@{}llll@{}}
\toprule
\textbf{Style} & \textbf{Input} & \textbf{Capability} & \textbf{Sequence} \\
\midrule
\textsc{Full}       & All num.\ params          & Direct tool use       & opt$\to$sim$\to$ren \\
\textsc{Natural}    & Text only                 & Ambiguity detect.     & ask\_human \\
\midrule
\textsc{W-Rand}     & + rand.\ STL\textsuperscript{$\dagger$}           & Instr.\ adherence     & opt$\to$sim$\to$stl \\
\textsc{W-Derived}  & + deriv.\ rules\textsuperscript{$\ddagger$}       & Arithmetic            & opt$\to$sim$\to$stl \\
\textsc{W-Distract} & + competing vals\textsuperscript{$\dagger$}       & Semantic disamb.      & opt$\to$sim$\to$stl \\
\textsc{W-Cond}     & + branching\textsuperscript{$\dagger$}            & Conditional           & opt$\to$sim$\to$stl \\
\textsc{W-Multi}    & + two exports\textsuperscript{$\dagger$}          & Working memory        & opt$\to$sim$\to$stl$\to$stl \\
\bottomrule
\end{tabular}
\end{table}

\subsubsection{\acs{RAG} Evaluation}
\label{sec:rag_eval}

We design four fixed prompts with different retrieval requirements (Table~\ref{tab:rag_prompts}; full text in Appendix~\ref{app:prompts}). Unlike the workflow benchmarks, which sample from EngiBench datasets, the \acs{RAG} evaluation instructs the agent to retrieve specific optimization parameters from indexed research papers and use them in a design task. The prompts range from retrieving a single parameter (\textsc{P0}) to combining parameters from two papers (\textsc{P3}).

\begin{table}[ht]
\centering
\caption[\acs{RAG} evaluation prompts with escalating retrieval difficulty.]{\acs{RAG} evaluation prompts with escalating retrieval difficulty. Scoring weights are given in Table~\ref{tab:rag_weights}.}
\label{tab:rag_prompts}
\footnotesize
\begin{tabular}{@{}clccc@{}}
\toprule
& & \multicolumn{3}{c}{\textbf{Target Parameters}} \\
\cmidrule(lr){3-5}
\textbf{ID} & \textbf{Source} & \texttt{volfrac} & \texttt{forcedist} & \texttt{rmin} \\
\midrule
\textsc{P0} & EngiBench paper \cite{felten_engibench_2025} & 0.35 & --- & --- \\
\textsc{P1} & EngiBench  \acs{API} ex. \cite{felten_engibench_2025} & 0.70 & 0.30 & --- \\
\textsc{P2} & SOPTX paper \cite{he_soptx_2025} & 0.40 & --- & 6.0 \\
\textsc{P3} & EngiBench \cite{felten_engibench_2025} + SOPTX \cite{he_soptx_2025} & 0.70 & 0.30 & 6.0 \\
\bottomrule
\end{tabular}
\end{table}

We evaluate under three conditions to isolate the contribution of document retrieval. \acs{RAG}-on provides reference documents indexed and retrievable via \emph{search\_documents}. \acs{RAG}-off removes retrieval tools entirely, forcing the agent to rely on parametric knowledge. Empty \acs{RAG} is the critical ablation: the retrieval tools are available but the index is empty, testing whether the mere availability of a retrieval tool changes agent behavior\textemdash for example, by inducing the agent to trust empty results and omit parameters it would otherwise guess from training data.

To ensure the only retrieval path is \emph{search\_documents}, we disable the web search and ArXiv agents for all \acs{RAG} evaluation runs.

The RAG service (MMORE) synthesizes retrieved passages at temperature 0.8; this applies only to retrieval-answer synthesis, and the evaluated agent remains at temperature zero. Additional details of the RAG service are provided in Appendix~\ref{app:compute}.

\subsubsection{\acs{HPC} Training Evaluation}
\label{sec:hpc_eval}
The HPC benchmark requires four operations. The agent must generate a SLURM script with the requested configuration, submit it, monitor the job until it terminates, and evaluate the trained model. Evaluation uses the EngiOpt metrics defined in Appendix~\ref{app:gen_metrics}: \ac{MMD}, \ac{DPP}, \ac{RVC}, and optimality gaps \ac{IOG}, \ac{COG}, and \ac{FOG}.

We define 10 prompts per style (seeds 1--10) with 100 fixed training epochs, using two prompt formats (Appendix~\ref{app:prompts}): \textsc{Explicit} (step-by-step instructions with tool names) and \textsc{Natural} (high-level description requiring the agent to infer the workflow). We report results for the \acs{cGAN} algorithm; diffusion model results are provided in Appendix~\ref{app:diffusion}.

\subsubsection{Training-Code Authoring}
\label{sec:tc_eval}

The training-code benchmark extends the HPC benchmark by requiring the agent to write part or all of the training program. T1 uses a fixed program and varies only its launch configuration. T2--T5 remove progressively larger code regions from the EngiOpt cGAN program. T6 removes the model and training implementation and supplies only the required command-line and evaluation interfaces. The evaluated implementation of this final tier is denoted T6v2. Neither construction is specific to the \acs{cGAN}: T2--T5 blank regions of any supplied reference implementation, a conditional diffusion template is also provided, and T6 imposes no algorithm at all.
Figure~\ref{fig:tc_tiers} lists the code supplied at each tier. Representative prompts are provided in Appendix~\ref{app:tc_prompts}.

\begin{figure*}[ht]
    \centering
    \includegraphics[width=0.9\linewidth]{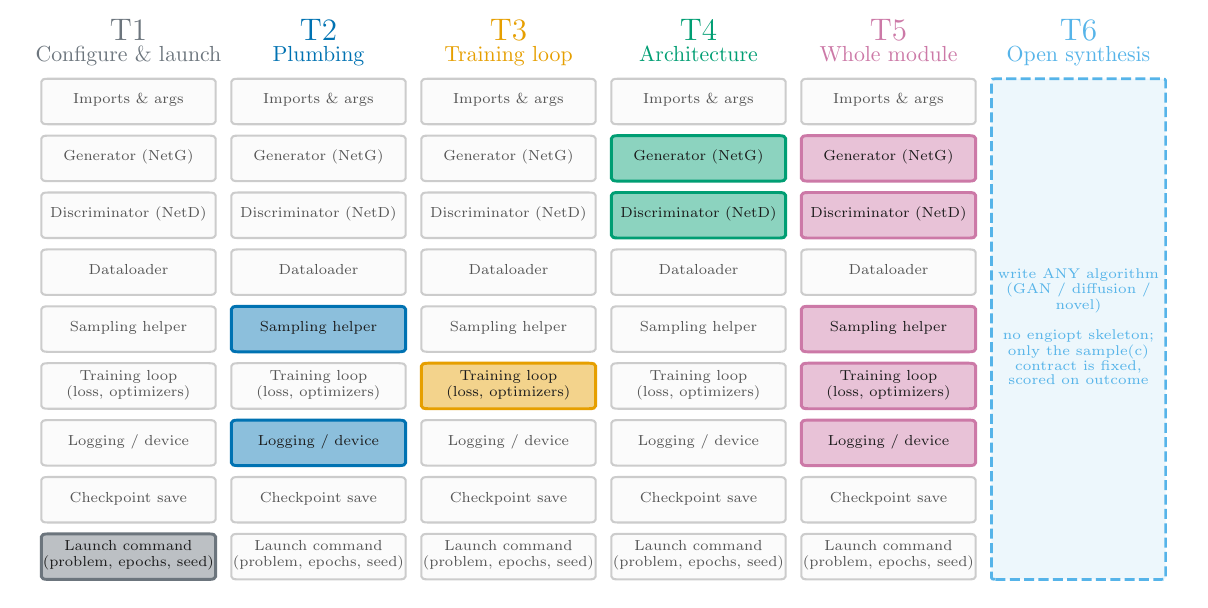}
    \caption[Training-code authoring tiers]{Training-code authoring tiers. Shaded blocks are written by the agent; unshaded blocks are supplied. T1 configures a fixed program, T2--T5 remove progressively larger code regions, and T6 supplies only the required interfaces. The data loader is supplied in T1--T5.}
    \label{fig:tc_tiers}
\end{figure*}

We evaluate checkpoints in two ways. The canonical evaluator loads the checkpoint with the reference \acs{cGAN} class. The functional evaluator loads the class and sampling function written by the agent, allowing architectures that do not match the reference class. Both evaluators compute \acs{MMD} from generated designs. In the functional evaluation, a sampler receives zero model-quality credit only if its outputs are both nearly constant across latent samples and insensitive to changes in conditioning.

\subsection{Scoring Methodology}
\label{subsec:scoring_methodology}

The workflow score combines design quality (65\%), tool use (20\%), and task completion (15\%). Design quality includes geometric agreement, engineering objectives and constraints, and, where applicable, manufacturability (2D connectivity, 3D watertightness). Tool use compares the observed calls with the expected calls. Task completion records whether all required operations succeeded. Appendix~\ref{app:metrics} defines each term and reports its weight.

Design quality receives the largest weight (65\%) because the benchmark primarily evaluates the engineering output. Tool use accounts for 20\% and distinguishes correct but inefficient traces by penalizing missing or redundant calls. Task completion accounts for the remaining 15\% and records whether the agent executed the full requested sequence. Appendix~\ref{app:raw_scores} reports the individual sub-scores, allowing the results to be examined under alternative weighting schemes.

Two limitations affect the interpretation of design quality. First, the watertightness sub-score is low across all models because the current threshold-based \acs{STL} extraction produces non-manifold meshes. This reflects a limitation of the extraction pipeline rather than agent capability, as noted in Appendix Tables~\ref{tab:raw_scores_beams2d} and~\ref{tab:raw_scores_photonics2d}. Second, the manufacturability sub-scores apply only to prompt styles that request \acs{STL} export. Design quality should therefore be compared within each prompt style rather than averaged across styles.

For evaluating generative models (\acs{HPC} training benchmark), we additionally report global distribution metrics\textemdash{}\ac{MMD} for dataset fidelity and \ac{DPP} for design diversity\textemdash{}as well as engineering performance metrics such as \ac{RVC} and optimality gaps (\ac{IOG}, \ac{COG}, \ac{FOG}), following Felten \textit{et al.}~\cite{felten_engibench_2025}; full definitions are given in Appendix~\ref{app:gen_metrics}.

\paragraph{\acs{RAG} Scoring}

The \acs{RAG} evaluation employs a gated scoring mechanism: parameter accuracy dimensions only contribute to the score when the agent invokes the \emph{search\_documents} tool, preventing credit for values reached through alternative means. The score combines per-parameter accuracy with a retrieval indicator (Appendix~\ref{app:rag_metrics}).

\paragraph{\acs{HPC} Workflow Scoring}

The \acs{HPC} training benchmark scores workflow orchestration rather than design quality. The primary metric is a weighted composite of step completion (70\%), configuration correctness (15\%), and evaluation metric extraction (15\%). Appendix~\ref{app:hpc_metrics} provides the full formulation.

\paragraph{Training-Code Scoring}

The training-code score assigns 55\% to model quality, 15\% to code validation, 10\% each to cluster-job completion and the one-epoch local dry run, and 5\% each to syntax validity and \acs{SLURM} acceptance. These terms are not independent. Model quality is scored only when the trial's own training job completes, so the largest term requires a full run that produces a working model, and the smaller terms record which intermediate checks the agent passed along the way. For a scorable model, the model-quality term is $\min(1,\mathrm{MMD}_{\mathrm{ref}}/\mathrm{MMD}_{\mathrm{agent}})$. The reference and agent models are matched by seed because reference \acs{MMD} varies substantially across seeds. Appendix~\ref{app:tc_metrics} defines the gates and complete score; Appendix~\ref{app:gen_metrics} defines the generative-model metrics.

\section{Results}
\label{sec:results}

\begin{table*}[ht]
\centering
\caption[Workflow evaluation results (Beams2D).]{Workflow evaluation results for Beams2D (mean $\pm$ std). TC = Task Completion rate, CO = Combined Overall score. \textbf{Bold} = best model per metric per row.}
\label{tab:workflow_results}
\footnotesize
\setlength{\tabcolsep}{4pt}
\begin{tabular*}{\textwidth}{@{\extracolsep{\fill}}lcccccccc@{}}
\toprule
 & \multicolumn{2}{c}{GPT-5-mini} & \multicolumn{2}{c}{Gemini-3-Flash} & \multicolumn{2}{c}{Qwen3-4B} & \multicolumn{2}{c}{Qwen3.5-4B} \\
\cmidrule(lr){2-3} \cmidrule(lr){4-5} \cmidrule(lr){6-7} \cmidrule(lr){8-9}
\textbf{Style} & \textbf{TC\,$\uparrow$} & \textbf{CO\,$\uparrow$} & \textbf{TC\,$\uparrow$} & \textbf{CO\,$\uparrow$} & \textbf{TC\,$\uparrow$} & \textbf{CO\,$\uparrow$} & \textbf{TC\,$\uparrow$} & \textbf{CO\,$\uparrow$} \\
\midrule
\textsc{Full} & \textbf{1.00\tiny{$\pm$0.00}} & 0.65\tiny{$\pm$0.07} & 0.93\tiny{$\pm$0.25} & \textbf{0.69\tiny{$\pm$0.10}} & 0.00\tiny{$\pm$0.00} & 0.45\tiny{$\pm$0.06} & 0.73\tiny{$\pm$0.44} & 0.59\tiny{$\pm$0.11} \\
\textsc{Natural} & 0.87\tiny{$\pm$0.34} & 0.82\tiny{$\pm$0.30} & \textbf{1.00\tiny{$\pm$0.00}} & \textbf{0.88\tiny{$\pm$0.23}} & 0.00\tiny{$\pm$0.00} & 0.29\tiny{$\pm$0.05} & 0.33\tiny{$\pm$0.47} & 0.48\tiny{$\pm$0.32} \\
\midrule
\textsc{W-Rand} & \textbf{1.00\tiny{$\pm$0.00}} & 0.66\tiny{$\pm$0.07} & \textbf{1.00\tiny{$\pm$0.00}} & 0.65\tiny{$\pm$0.09} & \textbf{1.00\tiny{$\pm$0.00}} & 0.64\tiny{$\pm$0.07} & \textbf{1.00\tiny{$\pm$0.00}} & \textbf{0.69\tiny{$\pm$0.06}} \\
\textsc{W-Derived} & \textbf{1.00\tiny{$\pm$0.00}} & 0.65\tiny{$\pm$0.07} & \textbf{1.00\tiny{$\pm$0.00}} & 0.64\tiny{$\pm$0.07} & 0.47\tiny{$\pm$0.50} & 0.53\tiny{$\pm$0.08} & 0.85\tiny{$\pm$0.36} & \textbf{0.66\tiny{$\pm$0.10}} \\
\textsc{W-Distract} & \textbf{1.00\tiny{$\pm$0.00}} & 0.63\tiny{$\pm$0.06} & \textbf{1.00\tiny{$\pm$0.00}} & 0.63\tiny{$\pm$0.08} & \textbf{1.00\tiny{$\pm$0.00}} & 0.61\tiny{$\pm$0.06} & 0.93\tiny{$\pm$0.25} & \textbf{0.66\tiny{$\pm$0.09}} \\
\textsc{W-Cond} & \textbf{0.93\tiny{$\pm$0.25}} & \textbf{0.66\tiny{$\pm$0.10}} & 0.87\tiny{$\pm$0.34} & 0.65\tiny{$\pm$0.10} & 0.40\tiny{$\pm$0.49} & 0.57\tiny{$\pm$0.12} & 0.60\tiny{$\pm$0.49} & 0.62\tiny{$\pm$0.13} \\
\textsc{W-Multi} & 0.93\tiny{$\pm$0.25} & 0.66\tiny{$\pm$0.09} & \textbf{1.00\tiny{$\pm$0.00}} & \textbf{0.70\tiny{$\pm$0.07}} & \textbf{1.00\tiny{$\pm$0.00}} & 0.65\tiny{$\pm$0.08} & \textbf{1.00\tiny{$\pm$0.00}} & 0.69\tiny{$\pm$0.06} \\
\midrule
\textit{Average} & \textit{0.96} & \textit{0.68} & \textbf{\textit{0.97}} & \textbf{\textit{0.69}} & \textit{0.55} & \textit{0.53} & \textit{0.78} & \textit{0.63} \\
\bottomrule
\end{tabular*}
\end{table*}
\begin{table*}[ht]
\centering
\caption[Workflow evaluation results (Photonics2D).]{Workflow evaluation results for Photonics2D (mean $\pm$ std). TC = Task Completion rate, CO = Combined Overall score. \textbf{Bold} = best model per metric per row. {\color{black!80}Gray text} shows Beams2D results on the same workflow styles for comparison.}
\label{tab:workflow_results_photonics}
\footnotesize
\setlength{\tabcolsep}{4pt}
\begin{tabular*}{\textwidth}{@{\extracolsep{\fill}}lcccccccc@{}}
\toprule
 & \multicolumn{2}{c}{GPT-5-mini} & \multicolumn{2}{c}{Gemini-3-Flash} & \multicolumn{2}{c}{Qwen3-4B} & \multicolumn{2}{c}{Qwen3.5-4B} \\
\cmidrule(lr){2-3} \cmidrule(lr){4-5} \cmidrule(lr){6-7} \cmidrule(lr){8-9}
\textbf{Style} & \textbf{TC\,$\uparrow$} & \textbf{CO\,$\uparrow$} & \textbf{TC\,$\uparrow$} & \textbf{CO\,$\uparrow$} & \textbf{TC\,$\uparrow$} & \textbf{CO\,$\uparrow$} & \textbf{TC\,$\uparrow$} & \textbf{CO\,$\uparrow$} \\
\midrule
\textsc{W-Rand} & \textbf{1.00\tiny{$\pm$0.00}} & 0.53\tiny{$\pm$0.01} & \textbf{1.00\tiny{$\pm$0.00}} & \textbf{0.60\tiny{$\pm$0.02}} & \textbf{1.00\tiny{$\pm$0.00}} & 0.56\tiny{$\pm$0.02} & \textbf{1.00\tiny{$\pm$0.00}} & 0.55\tiny{$\pm$0.01} \\
\textsc{W-Distract} & \textbf{1.00\tiny{$\pm$0.00}} & 0.52\tiny{$\pm$0.02} & \textbf{1.00\tiny{$\pm$0.00}} & 0.55\tiny{$\pm$0.03} & \textbf{1.00\tiny{$\pm$0.00}} & 0.52\tiny{$\pm$0.02} & \textbf{1.00\tiny{$\pm$0.00}} & \textbf{0.57\tiny{$\pm$0.05}} \\
\textsc{W-Cond} & 0.40\tiny{$\pm$0.49} & 0.47\tiny{$\pm$0.08} & \textbf{0.53\tiny{$\pm$0.50}} & \textbf{0.54\tiny{$\pm$0.09}} & 0.20\tiny{$\pm$0.40} & 0.43\tiny{$\pm$0.07} & 0.47\tiny{$\pm$0.50} & 0.49\tiny{$\pm$0.08} \\
\midrule
\textit{Average} & \textit{0.80} & \textit{0.51} & \textbf{\textit{0.84}} & \textbf{\textit{0.56}} & \textit{0.73} & \textit{0.50} & \textit{0.82} & \textit{0.54} \\
\midrule
\multicolumn{9}{l}{{\color{black!80}\textit{Beams2D (same styles, for comparison)}}} \\
{\color{black!80}\textsc{W-Rand}} & {\color{black!80}\textbf{1.00\tiny{$\pm$0.00}}} & {\color{black!80}0.66\tiny{$\pm$0.07}} & {\color{black!80}\textbf{1.00\tiny{$\pm$0.00}}} & {\color{black!80}0.65\tiny{$\pm$0.09}} & {\color{black!80}\textbf{1.00\tiny{$\pm$0.00}}} & {\color{black!80}0.64\tiny{$\pm$0.07}} & {\color{black!80}\textbf{1.00\tiny{$\pm$0.00}}} & {\color{black!80}\textbf{0.69\tiny{$\pm$0.06}}} \\
{\color{black!80}\textsc{W-Distract}} & {\color{black!80}\textbf{1.00\tiny{$\pm$0.00}}} & {\color{black!80}0.63\tiny{$\pm$0.06}} & {\color{black!80}\textbf{1.00\tiny{$\pm$0.00}}} & {\color{black!80}0.63\tiny{$\pm$0.08}} & {\color{black!80}\textbf{1.00\tiny{$\pm$0.00}}} & {\color{black!80}0.61\tiny{$\pm$0.06}} & {\color{black!80}0.93\tiny{$\pm$0.25}} & {\color{black!80}\textbf{0.66\tiny{$\pm$0.09}}} \\
{\color{black!80}\textsc{W-Cond}} & {\color{black!80}\textbf{0.93\tiny{$\pm$0.25}}} & {\color{black!80}\textbf{0.66\tiny{$\pm$0.10}}} & {\color{black!80}0.87\tiny{$\pm$0.34}} & {\color{black!80}0.65\tiny{$\pm$0.10}} & {\color{black!80}0.40\tiny{$\pm$0.49}} & {\color{black!80}0.57\tiny{$\pm$0.12}} & {\color{black!80}0.60\tiny{$\pm$0.49}} & {\color{black!80}0.62\tiny{$\pm$0.13}} \\
\cmidrule{1-9}
{\color{black!80}\textit{Average}} & {\color{black!80}\textbf{\textit{0.98}}} & {\color{black!80}\textit{0.65}} & {\color{black!80}\textit{0.96}} & {\color{black!80}\textit{0.64}} & {\color{black!80}\textit{0.80}} & {\color{black!80}\textit{0.61}} & {\color{black!80}\textit{0.84}} & {\color{black!80}\textbf{\textit{0.66}}} \\
\bottomrule
\end{tabular*}
\end{table*}

\begin{figure*}[ht]
    \centering
    \includegraphics[width=.9\linewidth]{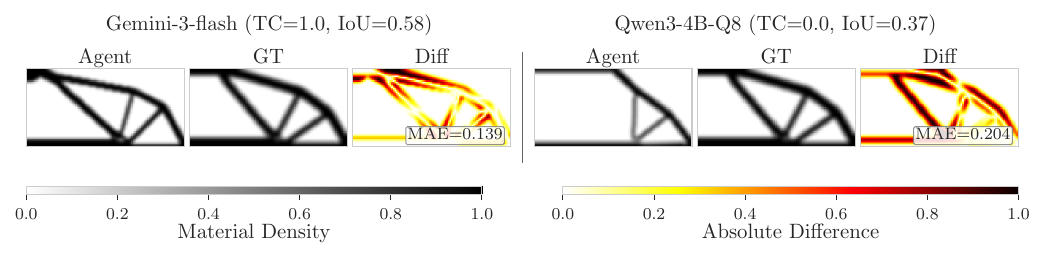}
    \caption[Design comparison, \textsc{W-Cond}]{Design comparison for the \textsc{W-Cond} style on the same problem instance (Beams2D, seed~3, example~3). Each group shows the agent-generated design (left), ground truth (center), and pixelwise absolute difference (right). Gemini-3-Flash: TC\,=\,1.0 and \acs{IoU}\,=\,0.58. Qwen3-4B: TC\,=\,0.0 and \acs{IoU}\,=\,0.37.}
    \label{fig:design_comparison}
\end{figure*}

\begin{table*}[t]
\centering
\caption[Tool-calling counts]{Tool-calling counts for the \textsc{Full} and \textsc{W-Cond} prompt styles: average calls per tool per sample (mean $\pm$ std). Cell shading is a linear interpolation between the smallest and largest average within each prompt style, and encodes how often a tool was called, not how well.}
\label{tab:tool_usage}
\footnotesize
\setlength{\tabcolsep}{4pt}
\resizebox{\textwidth}{!}{%
\begin{tabular}{@{}lcccccccc@{}}
\toprule
 & \multicolumn{4}{c}{\textsc{Full}} & \multicolumn{4}{c}{\textsc{W-Cond}} \\
\cmidrule(lr){2-5} \cmidrule(lr){6-9}
\textbf{Tool} & GPT-5-mini & Gemini-3-Flash & Qwen3-4B-Q8 & Qwen3.5-4B-Q8 & GPT-5-mini & Gemini-3-Flash & Qwen3-4B-Q8 & Qwen3.5-4B-Q8 \\
\midrule
\texttt{simulate\_design} & \cellcolor{toolshade!65}1.0\tiny{$\pm$0.0} & \cellcolor{toolshade!65}1.0\tiny{$\pm$0.0} & \cellcolor{toolshade!65}1.0\tiny{$\pm$0.0} & \cellcolor{toolshade!65}1.0\tiny{$\pm$0.0} & \cellcolor{toolshade!54}1.0\tiny{$\pm$0.0} & \cellcolor{toolshade!65}1.2\tiny{$\pm$0.5} & \cellcolor{toolshade!54}1.0\tiny{$\pm$0.0} & \cellcolor{toolshade!54}1.0\tiny{$\pm$0.0} \\
\texttt{optimize\_design} & \cellcolor{toolshade!65}1.0\tiny{$\pm$0.0} & \cellcolor{toolshade!65}1.0\tiny{$\pm$0.0} & \cellcolor{toolshade!65}1.0\tiny{$\pm$0.0} & \cellcolor{toolshade!65}1.0\tiny{$\pm$0.0} & \cellcolor{toolshade!54}1.0\tiny{$\pm$0.0} & \cellcolor{toolshade!54}1.0\tiny{$\pm$0.0} & \cellcolor{toolshade!54}1.0\tiny{$\pm$0.0} & \cellcolor{toolshade!54}1.0\tiny{$\pm$0.0} \\
\texttt{convert\_design\_to\_stl} & \cellcolor{toolshade!43}0.7\tiny{$\pm$0.5} & \cellcolor{toolshade!0}0.0\tiny{$\pm$0.0} & \cellcolor{toolshade!0}0.0\tiny{$\pm$0.0} & \cellcolor{toolshade!0}0.0\tiny{$\pm$0.0} & \cellcolor{toolshade!65}1.2\tiny{$\pm$0.4} & \cellcolor{toolshade!54}1.0\tiny{$\pm$0.0} & \cellcolor{toolshade!43}0.8\tiny{$\pm$0.5} & \cellcolor{toolshade!54}1.0\tiny{$\pm$0.0} \\
\texttt{render\_design} & \cellcolor{toolshade!65}1.0\tiny{$\pm$0.0} & \cellcolor{toolshade!61}0.9\tiny{$\pm$0.2} & \cellcolor{toolshade!0}0.0\tiny{$\pm$0.0} & \cellcolor{toolshade!52}0.8\tiny{$\pm$0.5} & \cellcolor{toolshade!25}0.5\tiny{$\pm$0.5} & \cellcolor{toolshade!25}0.5\tiny{$\pm$0.5} & \cellcolor{toolshade!14}0.3\tiny{$\pm$0.4} & \cellcolor{toolshade!0}0.0\tiny{$\pm$0.0} \\
\texttt{create\_problem} & \cellcolor{toolshade!26}0.4\tiny{$\pm$0.5} & \cellcolor{toolshade!0}0.0\tiny{$\pm$0.0} & \cellcolor{toolshade!65}1.0\tiny{$\pm$0.0} & \cellcolor{toolshade!65}1.0\tiny{$\pm$0.0} & \cellcolor{toolshade!4}0.1\tiny{$\pm$0.2} & \cellcolor{toolshade!0}0.0\tiny{$\pm$0.0} & \cellcolor{toolshade!32}0.6\tiny{$\pm$0.5} & \cellcolor{toolshade!0}0.0\tiny{$\pm$0.0} \\
\texttt{get\_problem\_details} & \cellcolor{toolshade!13}0.2\tiny{$\pm$0.4} & \cellcolor{toolshade!0}0.0\tiny{$\pm$0.0} & \cellcolor{toolshade!0}0.0\tiny{$\pm$0.0} & \cellcolor{toolshade!0}0.0\tiny{$\pm$0.0} & \cellcolor{toolshade!0}0.0\tiny{$\pm$0.0} & \cellcolor{toolshade!0}0.0\tiny{$\pm$0.0} & \cellcolor{toolshade!0}0.0\tiny{$\pm$0.0} & \cellcolor{toolshade!0}0.0\tiny{$\pm$0.0} \\
\bottomrule
\end{tabular}%
}
\end{table*}

\begin{figure*}[ht]
    \centering
    \begin{subfigure}[t]{0.48\linewidth}
        \centering
        \includegraphics[width=0.70\linewidth]{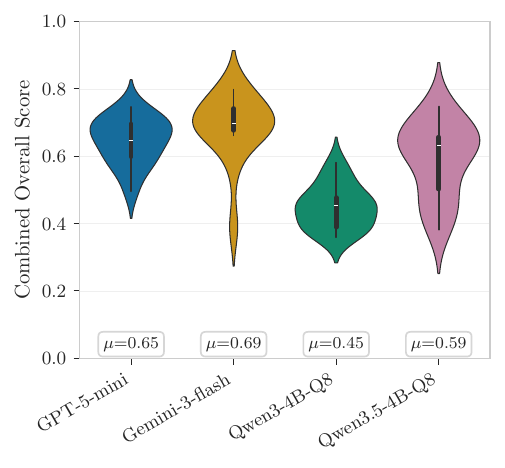}
        \caption{\textsc{Full}}
        \label{fig:co_violin_full}
    \end{subfigure}\hfill
    \begin{subfigure}[t]{0.48\linewidth}
        \centering
        \includegraphics[width=0.70\linewidth]{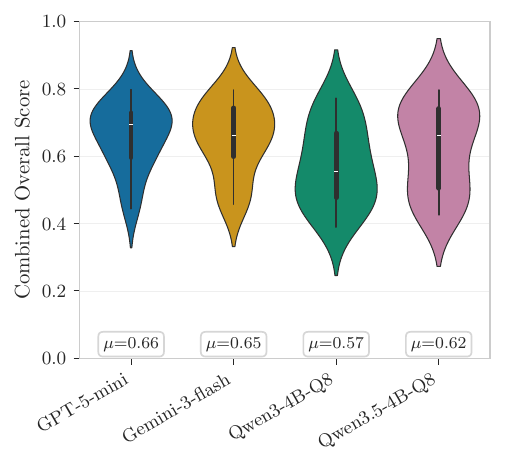}
        \caption{\textsc{W-Cond}}
        \label{fig:co_violin_wcond}
    \end{subfigure}
    \caption[Combined overall score distributions]{Combined overall score distributions for the \textsc{Full} (a) and \textsc{W-Cond} (b) prompt styles by model backend.}
    \label{fig:co_violin}
\end{figure*}

\begin{figure}[ht]
    \centering
    \includegraphics[width=0.7\linewidth]{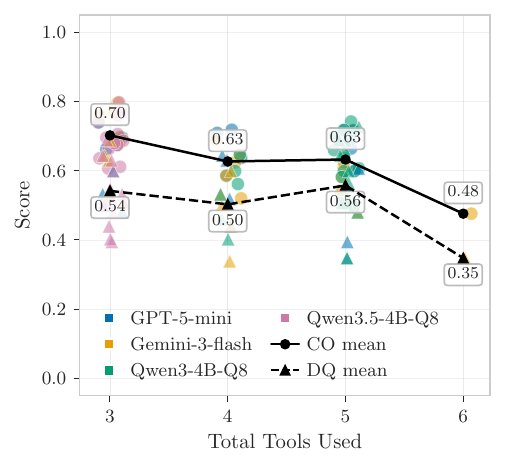}
    \caption[Combined overall score vs.\ design quality score by tool count.]{Mean combined overall score (CO, solid line) and design quality score (DQ, dashed line) for the \textsc{W-Rand} style, grouped by the total number of tool calls.}
    \label{fig:co_vs_dq_tools}
\end{figure}

\begin{figure*}[ht]
    \centering
    \includegraphics[width=0.9\linewidth]{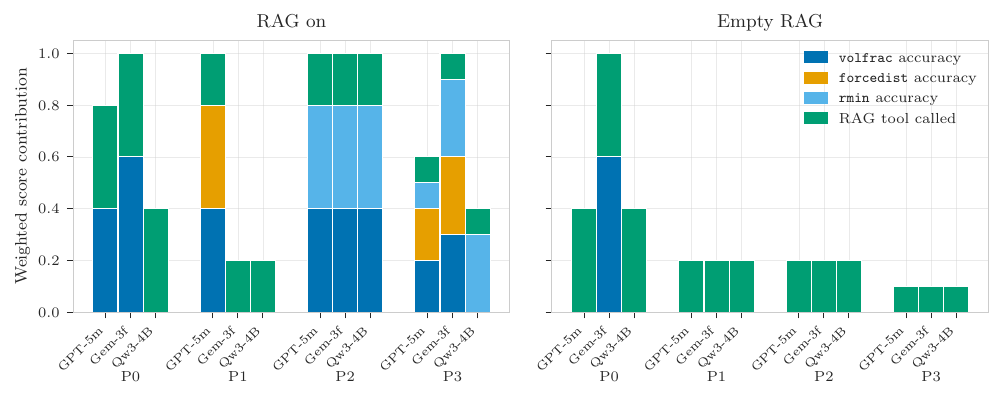}
    \caption[Weighted RAG score contributions]{Weighted \acs{RAG} score contributions by prompt and \acs{LLM} backend under \acs{RAG}-on and Empty \acs{RAG} conditions (three runs per cell). \acs{RAG}-off is omitted because all scores are zero by construction.}
    \label{fig:rag_score_components}
\end{figure*}

\begin{table}[t]
\centering
\caption[Workflow step completion]{Workflow step completion for the \textsc{Explicit} and \textsc{Natural} prompt styles: average completion rate per workflow step (mean $\pm$ std over the runs in that cell). Shading is the completion rate.}
\label{tab:hpc_step_completion}
\footnotesize
\setlength{\tabcolsep}{4pt}
\begin{tabular*}{\columnwidth}{@{\extracolsep{\fill}}lcccc@{}}
\toprule
 & \multicolumn{2}{c}{\textsc{Explicit}} & \multicolumn{2}{c}{\textsc{Natural}} \\
\cmidrule(lr){2-3} \cmidrule(lr){4-5}
\textbf{Step} & GPT-5-mini & Gemini-3-Flash & GPT-5-mini & Gemini-3-Flash \\
\midrule
Generate cmd & \cellcolor{toolshade!65}100\%\tiny{$\pm$0.0} & \cellcolor{toolshade!65}100\%\tiny{$\pm$0.0} & \cellcolor{toolshade!58}90\%\tiny{$\pm$31.6} & \cellcolor{toolshade!65}100\%\tiny{$\pm$0.0} \\
Submit job & \cellcolor{toolshade!65}100\%\tiny{$\pm$0.0} & \cellcolor{toolshade!65}100\%\tiny{$\pm$0.0} & \cellcolor{toolshade!52}80\%\tiny{$\pm$42.2} & \cellcolor{toolshade!65}100\%\tiny{$\pm$0.0} \\
Monitor job & \cellcolor{toolshade!58}90\%\tiny{$\pm$31.6} & \cellcolor{toolshade!65}100\%\tiny{$\pm$0.0} & \cellcolor{toolshade!52}80\%\tiny{$\pm$42.2} & \cellcolor{toolshade!65}100\%\tiny{$\pm$0.0} \\
Evaluate & \cellcolor{toolshade!46}70\%\tiny{$\pm$48.3} & \cellcolor{toolshade!65}100\%\tiny{$\pm$0.0} & \cellcolor{toolshade!32}50\%\tiny{$\pm$52.7} & \cellcolor{toolshade!65}100\%\tiny{$\pm$0.0} \\
\bottomrule
\end{tabular*}
\end{table}

\begin{figure*}[ht]
    \centering
    \includegraphics[width=0.9\linewidth]{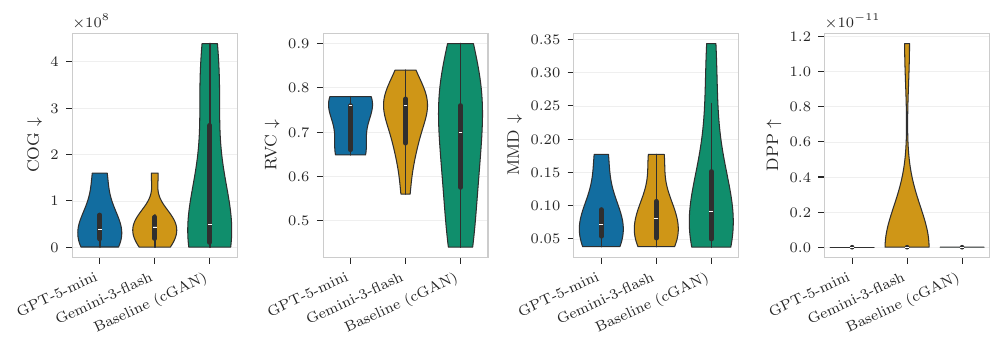}
    \caption[Offline cGAN model-quality metrics]{Offline model quality metrics (\acs{COG}, \acs{RVC}, \acs{MMD}, \acs{DPP}) for agent-trained \acs{cGAN} models vs.\ EngiBench baselines. Arrows indicate desired direction. Values averaged across available seeds.}
    \label{fig:hpc_baseline_grid}
\end{figure*}

\subsection{Workflow Evaluation}
Table~\ref{tab:workflow_results} reports task completion (TC) and combined overall (CO) scores for the seven Beams2D prompt styles and four \acs{LLM} backends.\footnote{A small number of Qwen3.5-4B runs (1--2 per affected style) were lost because of JSON parsing errors, giving $n{=}13$--$14$ instead of $15$ for three workflow styles. All other model--style cells contain the full $n{=}15$.} GPT-5-mini and Gemini-3-Flash achieve near-perfect task completion on most styles. Average TC is lower for Qwen3-4B (55\%) but increases to 78\% for Qwen3.5-4B.

On Beams2D, among the five W-styles, \textsc{W-Cond} has the lowest task-completion across the four backends. The highest completion rate is 0.93, obtained by GPT-5-mini. This prompt requires the agent to select an operation using the output of a preceding tool call. Figure~\ref{fig:design_comparison} shows one example. Gemini-3-Flash selects the correct branch and produces a design that passes parameter validation (\acs{IoU}\,=\,0.58). Qwen3-4B selects the wrong branch, fails parameter validation, and produces a different topology (\acs{IoU}\,=\,0.37).

The combined overall score is the hierarchical workflow score $S_{\text{workflow}}$ defined in Appendix~\ref{app:metrics}. The two proprietary models obtain similar scores, whereas the average CO increases from 0.53 for Qwen3-4B to 0.63 for Qwen3.5-4B. No pairwise model difference remains significant after correction on Beams2D. The scores contain many ties, so this result indicates limited statistical discrimination rather than model equivalence. On Photonics2D, 14 of the 18 model-pair comparisons remain significant after correction. Gemini-3-Flash outperforms the other backends on most styles with large effects (rank-biserial ${\geq}\,0.75$), and the improvement from Qwen3-4B to Qwen3.5-4B is significant on the distractor and conditional styles (BH-corrected $q<0.05$). Appendix~\ref{app:stats} reports the complete tests and effect sizes.

Task completion and CO measure different outcomes. A failed workflow can receive a nonzero CO when the agent still produces a partial design or makes some valid tool calls. For example, Qwen3-4B obtains CO\,=\,0.57 with TC\,=\,0.40 on \textsc{W-Cond}. On \textsc{Full} and \textsc{Natural}, it obtains TC\,=\,0 but CO\,=\,0.45 and 0.29, respectively.

These differences follow from the task-completion criteria. On \textsc{Full}, Qwen3-4B completes the earlier operations but consistently omits the final \emph{render\_design} call. The W-style prompts explicitly name each operation, including \acs{STL} export, and therefore provide more guidance for task sequencing. On \textsc{Natural}, Qwen3-4B proceeds with optimization instead of recognizing the missing parameters and calling \emph{ask\_human\_for\_clarification}. Qwen3.5-4B improves completion on \textsc{Full} to TC\,=\,0.73, but reaches only TC\,=\,0.33 on \textsc{Natural}. The newer model therefore follows the complete explicit workflow more reliably, while detecting underspecified requests remains difficult.

Table~\ref{tab:tool_usage} shows the corresponding tool-call counts. On \textsc{Full}, GPT-5-mini and Gemini-3-Flash call each expected tool once. Qwen3-4B makes redundant calls and omits \emph{render\_design}, whereas Qwen3.5-4B follows the expected trace more closely. Figure~\ref{fig:co_violin} shows the resulting CO distributions for \textsc{Full} and \textsc{W-Cond}. On \textsc{W-Rand}, additional tool calls reduce CO without improving design quality, which remains near 0.5 (Figure~\ref{fig:co_vs_dq_tools}). Section~\ref{sec:discussion} discusses this tool-use inefficiency.

\textit{Photonics2D evaluation.} Table~\ref{tab:workflow_results_photonics} reports the results for all four backends on three Photonics2D prompt styles. Several design-quality components provide little discrimination on this problem: background pixels inflate pixel accuracy, the optimizer directly enforces the constraint term, and the multi-component photonic designs receive uniformly low 2D-connectivity scores. We therefore emphasize task completion and tool use.

All four models complete every \textsc{W-Rand} and \textsc{W-Distract} trial (TC\,=\,1.00). Performance is substantially lower on \textsc{W-Cond}. Gemini-3-Flash achieves the highest completion rate at 53\%, followed by Qwen3.5-4B at 47\%, GPT-5-mini at 40\%, and Qwen3-4B at 20\%. For comparison, the highest completion rate on Beams2D \textsc{W-Cond} is 93\%.

Most Photonics2D \textsc{W-Cond} failures result from selecting the wrong conditional branch rather than omitting the export operation. Three models call \emph{convert\_design\_to\_stl} in every trial. Qwen3-4B calls it in 8 of 15 trials, accounting for seven failures caused by a missing export. Among the 29 failed trials that perform the export, 28 select the opposite branch. These trials exchange the branch-specific parameters (\emph{threshold} and \emph{mirror\_y}) while retaining the correct shared parameters (\emph{scale\_xy} and \emph{scale\_z}). The agents therefore usually preserve the parameter values but apply the conditional comparison incorrectly. Appendix~\ref{app:photonics} provides the tool-call heatmaps and score distributions. The present experiment does not establish why branch inversion occurs more frequently on Photonics2D.

\subsection{\acs{RAG} Evaluation}

Having established workflow performance, we next evaluate whether document retrieval improves engineering parameter selection. Figure~\ref{fig:rag_score_components} reports the weighted score contributions for four prompts and three \acs{LLM} backends: GPT-5-mini, Gemini-3-Flash, and Qwen3-4B.\footnote{Qwen3.5-4B is excluded from the \acs{RAG} and \acs{HPC} evaluations because it was added after these experiments were completed.} Each model is evaluated three times under the three retrieval conditions defined in Section~\ref{sec:rag_eval}.

The \acs{RAG}-off condition receives a score of zero by definition because the gated scoring mechanism awards parameter credit only when the value is supported by retrieved content. We therefore omit this condition from the figure. Scores are generally highest with \acs{RAG}-on, when the reference documents are indexed, and decrease substantially with Empty \acs{RAG}. This comparison shows that providing access to the retrieval tool without the relevant documents does not reproduce the \acs{RAG}-on results.

Prompt P0 is an exception for Gemini-3-Flash, which returns the expected volume fraction even when the index is empty. The requested value, 0.35, is common in topology optimization, appears in Sigmund's 99-line benchmark~\cite{sigmund_99_2001}, and is also the EngiBench default. The result may therefore reflect prior model knowledge or a common default, although the experiment cannot establish its source. Prompts P2 and P3 instead request parameters from the more recent paper~\cite{he_soptx_2025}. No model returns the complete expected parameter set for these prompts without the indexed documents.

Performance also varies by model and prompt. Gemini-3-Flash obtains higher scores than GPT-5-mini on P0 and P3, whereas GPT-5-mini scores higher on P1. Qwen3-4B generally obtains lower scores than the proprietary models, except on P2, where all three models receive the same score.

\subsection{\acs{HPC} Training Evaluation}

The third evaluation examines whether agents can execute long-running training workflows on remote infrastructure. We evaluate only GPT-5-mini and Gemini-3-Flash because each trial requires substantial wall-clock time for remote submission, training, and monitoring.

Table~\ref{tab:hpc_step_completion} reports completion of script generation, job submission, monitoring, and model evaluation under two prompt styles. The \textsc{Explicit} prompts enumerate the required tool calls, whereas the \textsc{Natural} prompts describe the same workflow without naming each step. Gemini-3-Flash completes all four steps in every tested trial under both prompt styles. GPT-5-mini reaches the final evaluation step in 70\% of \textsc{Explicit} trials and 50\% of \textsc{Natural} trials. Its job-submission and monitoring completion rates also decrease from 90\% to 80\%.

Inspection of the tool traces shows that GPT-5-mini most often stops before calling \emph{evaluate\_model}. The affected runs contain no recorded timeout or tool exception; the agent simply issues no subsequent call. Explicit step enumeration improves completion in these experiments, but the evaluation does not isolate why GPT-5-mini stops during some longer workflows. Gemini-3-Flash completes every tested workflow under both prompt styles.

Figure~\ref{fig:hpc_baseline_grid} compares the offline quality metrics of the models produced by completed \acs{cGAN} runs with the corresponding EngiBench references. The agent-trained models obtain similar metric values in the completed trials, with some variation across runs.

\subsection{Training-Code Authoring}
\label{sec:tc_results}

\begin{figure*}[t]
    \centering
    \includegraphics[width=.9\linewidth]{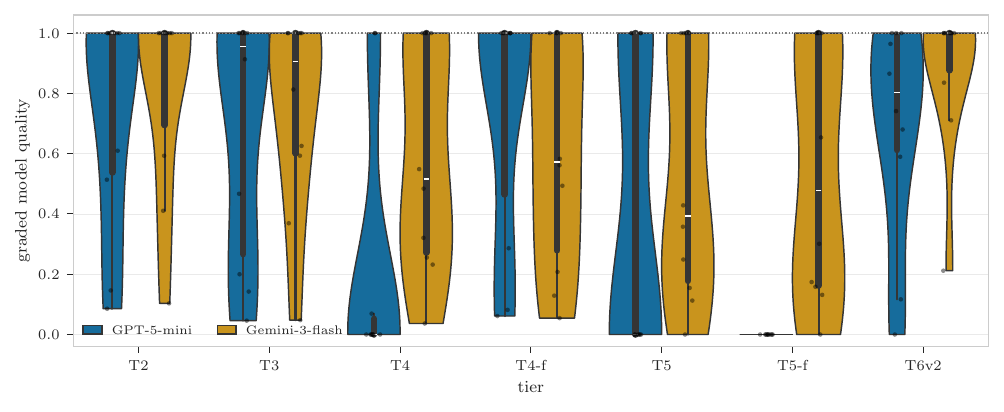}
    \caption[Per-seed model quality by tier]{Per-seed model-quality scores by training-code tier and model backend. Violin plots summarize ten seeds; boxes show the median and interquartile range, and points show individual seeds.}
    \label{fig:tc_score_grid}
\end{figure*}

\begin{figure}[t]
    \centering
    \includegraphics[width=0.85\linewidth]{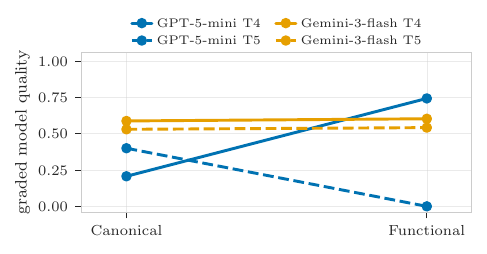}
    \caption[Canonical and functional model quality]{Mean model-quality scores at T4 and T5 under the canonical and functional evaluators. The canonical evaluator loads the checkpoint using the reference cGAN class; the functional evaluator uses the agent-authored model class and sampling function.}
    \label{fig:tc_two_track}
\end{figure}

We evaluate seven training-code variants on GPT-5-mini and Gemini-3-Flash: T2--T5, the functional variants of T4 and T5, and T6v2. Each variant is evaluated over ten seeds using the prompts reproduced in Appendix~\ref{app:tc_prompts}. Section~\ref{sec:tc_eval} and Figure~\ref{fig:tc_tiers} define the tiers. Table~\ref{tab:tc_results} reports the composite score $S_{\mathrm{tc}}$ and its model-quality component. Figures~\ref{fig:tc_score_grid} and~\ref{fig:tc_two_track} show variation across tiers and compare the canonical and functional evaluators. Appendix~\ref{app:stats} provides the statistical tests and weight-sensitivity analysis.

\begin{table}[t]
\centering
\caption[Training-code results]{Training-code results (mean $\pm$ std over ten seeds). $S_{\text{tc}}$ = composite score; MQ = graded model-quality term ($\min(1,\ \mathrm{MMD}_{\text{ref}}/\mathrm{MMD}_{\text{agent}})$, per seed). \textbf{Bold} = better model per tier per metric.}
\label{tab:tc_results}
\footnotesize
\setlength{\tabcolsep}{4pt}
\begin{tabular*}{\columnwidth}{@{\extracolsep{\fill}}lcccc@{}}
\toprule
 & \multicolumn{2}{c}{GPT-5-mini} & \multicolumn{2}{c}{Gemini-3-Flash} \\
\cmidrule(lr){2-3} \cmidrule(lr){4-5}
\textbf{Tier} & \textbf{$S_{\text{tc}}\,\uparrow$} & \textbf{MQ\,$\uparrow$} & \textbf{$S_{\text{tc}}\,\uparrow$} & \textbf{MQ\,$\uparrow$} \\
\midrule
T2 (plumbing)       & 0.85\tiny{$\pm$0.20} & 0.74\tiny{$\pm$0.37} & \textbf{0.89}\tiny{$\pm$0.19} & \textbf{0.81}\tiny{$\pm$0.33} \\
T3 (training loop)  & 0.82\tiny{$\pm$0.23} & 0.68\tiny{$\pm$0.41} & \textbf{0.85}\tiny{$\pm$0.18} & \textbf{0.74}\tiny{$\pm$0.33} \\
T4 (architecture)   & 0.55\tiny{$\pm$0.24} & 0.21\tiny{$\pm$0.42} & \textbf{0.77}\tiny{$\pm$0.20} & \textbf{0.59}\tiny{$\pm$0.38} \\
T4-func             & \textbf{0.81}\tiny{$\pm$0.26} & \textbf{0.74}\tiny{$\pm$0.42} & 0.77\tiny{$\pm$0.22} & 0.60\tiny{$\pm$0.38} \\
T5 (whole module)   & 0.65\tiny{$\pm$0.26} & 0.40\tiny{$\pm$0.52} & \textbf{0.74}\tiny{$\pm$0.23} & \textbf{0.53}\tiny{$\pm$0.42} \\
T5-func             & 0.41\tiny{$\pm$0.02} & 0.00\tiny{$\pm$0.00} & \textbf{0.70}\tiny{$\pm$0.32} & \textbf{0.54}\tiny{$\pm$0.43} \\
T6v2 (open synth.)  & 0.79\tiny{$\pm$0.21} & 0.70\tiny{$\pm$0.37} & \textbf{0.89}\tiny{$\pm$0.14} & \textbf{0.88}\tiny{$\pm$0.25} \\
\bottomrule
\end{tabular*}
\end{table}

At T2, the mean model-quality score is 0.74 for GPT-5-mini and 0.81 for Gemini-3-Flash. The difference between the models is larger at T4, where the agent writes the network architecture. Under the canonical evaluator, GPT-5-mini scores 0.21 and Gemini-3-Flash scores 0.59. The GPT-5-mini programs complete training, but several checkpoints cannot be loaded into the reference \acs{cGAN} class because the authored layer names and structure do not match the reference implementation.

The functional evaluator instead loads the class written by the agent. This raises the GPT-5-mini T4 score from 0.21 to 0.74 (Figure~\ref{fig:tc_two_track}), showing that much of the canonical penalty results from interface incompatibility rather than training failure. The functional T5 evaluation adds another requirement: the agent must supply a working sampling function. GPT-5-mini scores 0.40 under the canonical evaluator, but its authored sampler produces nearly constant outputs that do not respond to conditioning and therefore receives zero model-quality credit under the functional evaluator. Gemini-3-Flash scores 0.53 and 0.54 on the canonical and functional T5 evaluations, respectively.

\begin{figure*}[t]
\centering
\includegraphics[width=0.75\linewidth]{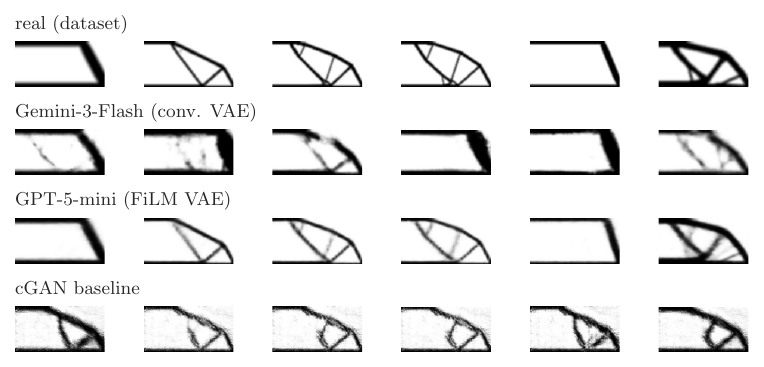}
\caption[Beams2D designs from T6v2 generators]{Beams2D designs produced by the open-synthesis generators for seed~5. Columns correspond to fixed load conditions taken from dataset designs. Rows show the dataset design, the Gemini-3-Flash convolutional VAE, the GPT-5-mini fully connected VAE with FiLM conditioning, and the reference \acs{cGAN}. Designs are shown without thresholding.}
\label{fig:tc_novel_designs}
\end{figure*}

At T6v2, both models select a conditional variational autoencoder rather than the reference \acs{cGAN}. Gemini-3-Flash implements a convolutional encoder and decoder with a reconstruction-plus-KL objective. GPT-5-mini implements a fully connected model with feature-wise linear modulation conditioning. Both programs complete training and provide a working sampling function. Figure~\ref{fig:tc_novel_designs} shows designs produced by the authored generators.

GPT-5-mini obtains a mean model-quality score of 0.70, while Gemini-3-Flash obtains 0.88. Gemini-3-Flash has a mean \acs{MMD} of 0.094, compared with 0.19 for the seed-matched reference \acs{cGAN}, and obtains lower \acs{MMD} on seven of ten seeds. This comparison is specific to these trials and does not establish general superiority of the VAE architecture.

Performance is not monotonic across tiers: model-quality scores decrease at T4 and T5 and increase again at T6v2 (Table~\ref{tab:tc_results}; Appendix~\ref{app:stats}). Some lower scores result from interface incompatibility or sampling-function failure rather than unsuccessful training. The tiers should therefore be interpreted as distinct authoring tasks.

\begin{table}[t]
\centering
\caption[Training-code results on Photonics2D]{Training-code results on Photonics2D (mean $\pm$ std over ten seeds). $S_{\text{tc}}$ = composite score; MQ = graded model-quality term ($\min(1,\ \mathrm{MMD}_{\text{ref}}/\mathrm{MMD}_{\text{agent}})$, per seed). \textbf{Bold} = better model per tier per metric. {\color{black!80}Gray text} shows Beams2D results on the same tiers for comparison.}
\label{tab:tc_results_photonics2d}
\footnotesize
\setlength{\tabcolsep}{4pt}
\begin{tabular*}{\columnwidth}{@{\extracolsep{\fill}}lcccc@{}}
\toprule
 & \multicolumn{2}{c}{GPT-5-mini} & \multicolumn{2}{c}{Gemini-3-Flash} \\
\cmidrule(lr){2-3} \cmidrule(lr){4-5}
\textbf{Tier} & \textbf{$S_{\text{tc}}\,\uparrow$} & \textbf{MQ\,$\uparrow$} & \textbf{$S_{\text{tc}}\,\uparrow$} & \textbf{MQ\,$\uparrow$} \\
\midrule
T2 (plumbing)       & 0.96\tiny{$\pm$0.05} & 0.94\tiny{$\pm$0.08} & \textbf{0.97}\tiny{$\pm$0.04} & \textbf{0.97}\tiny{$\pm$0.07} \\
T4 (architecture)   & 0.59\tiny{$\pm$0.27} & 0.30\tiny{$\pm$0.48} & \textbf{0.78}\tiny{$\pm$0.29} & \textbf{0.60}\tiny{$\pm$0.52} \\
T6v2 (open synth.)  & 0.89\tiny{$\pm$0.17} & 0.90\tiny{$\pm$0.32} & \textbf{0.95}\tiny{$\pm$0.02} & \textbf{1.00}\tiny{$\pm$0.01} \\
\midrule
\multicolumn{5}{l}{{\color{black!80}\textit{Beams2D (same tiers, for comparison)}}} \\
{\color{black!80}T2 (plumbing)} & {\color{black!80}0.85\tiny{$\pm$0.20}} & {\color{black!80}0.74\tiny{$\pm$0.37}} & {\color{black!80}\textbf{0.89}\tiny{$\pm$0.19}} & {\color{black!80}\textbf{0.81}\tiny{$\pm$0.33}} \\
{\color{black!80}T4 (architecture)} & {\color{black!80}0.55\tiny{$\pm$0.24}} & {\color{black!80}0.21\tiny{$\pm$0.42}} & {\color{black!80}\textbf{0.77}\tiny{$\pm$0.20}} & {\color{black!80}\textbf{0.59}\tiny{$\pm$0.38}} \\
{\color{black!80}T6v2 (open synth.)} & {\color{black!80}0.79\tiny{$\pm$0.21}} & {\color{black!80}0.70\tiny{$\pm$0.37}} & {\color{black!80}\textbf{0.89}\tiny{$\pm$0.14}} & {\color{black!80}\textbf{0.88}\tiny{$\pm$0.25}} \\
\bottomrule
\end{tabular*}
\end{table}

\textit{Photonics2D replication.} We repeat T2, T4, and T6v2 on Photonics2D using the same two model backends and ten seeds per tier. Each trial authors code specifically for its assigned problem. Table~\ref{tab:tc_results_photonics2d} reports the results, while Appendix~\ref{app:portability} separately tests whether T6v2 programs can run unchanged on the other problem.

At T2, the mean model-quality scores are 0.94 for GPT-5-mini and 0.97 for Gemini-3-Flash. At T4, the canonical evaluator successfully grades three GPT-5-mini checkpoints and six Gemini-3-Flash checkpoints. The remaining checkpoints cannot be loaded into the reference class. Among the successfully graded checkpoints, mean \acs{MMD} ranges from 0.54 to 0.64, compared with 0.86 for the reference \acs{cGAN}. The low aggregate T4 scores therefore primarily reflect reference-interface incompatibility.

At T6v2, mean \acs{MMD} is 0.45 for GPT-5-mini and 0.22 for Gemini-3-Flash. Their corresponding model-quality scores are 0.90 and 1.00. One GPT-5-mini sampler is rejected by the degeneracy gate. Appendix Figure~\ref{fig:tc_novel_designs_photonics2d} shows designs from both authored generators and the reference \acs{cGAN}.

These results require two qualifications. First, the reference \acs{cGAN} has an \acs{MMD} of 0.86 on Photonics2D, so exceeding this reference does not by itself establish strong absolute model performance. Second, the Page trend test~\cite{4ee01594-9356-33d3-accf-4645c9c68c8c} across the three selected tiers is not significant ($p{=}0.11$). The decrease at T4 followed by an increase at T6v2 is also non-monotonic. The Photonics2D experiment therefore provides a qualitative cross-domain replication, not evidence of a general trend across authoring tiers (Appendix~\ref{app:stats}).

\section{Discussion}
\label{sec:discussion}
\subsection{Key Findings}

We organize the findings around the four benchmark dimensions introduced in Section~\ref{sec:introduction}: workflow execution, retrieval-assisted parameter selection, \acs{HPC} orchestration, and training-code authoring.

\textit{Workflow execution.} GPT-5-mini and Gemini-3-Flash achieve 96--97\% average task completion across the seven Beams2D prompt styles (Table~\ref{tab:workflow_results}). At the same 4B parameter count, task completion increases from 55\% for Qwen3-4B to 78\% for Qwen3.5-4B. The largest improvements occur on \textsc{Full} ($0.00\rightarrow0.73$) and \textsc{W-Derived} ($0.47\rightarrow0.85$). The difference also appears on Photonics2D \textsc{W-Cond}, where completion increases from 0.20 to 0.47, although the highest rate among all models is only 0.53 (Table~\ref{tab:workflow_results_photonics}). Conditional branching and the detection of underspecified requests therefore remain less reliable than explicitly specified workflows.

Tool-call efficiency provides information beyond task completion. On \textsc{W-Rand}, increasing the number of calls from three to four or five reduces the combined score from 0.70 to 0.63 without improving design quality, which remains near 0.5 (Figure~\ref{fig:co_vs_dq_tools}). Additional calls therefore impose a score penalty without improving the resulting design in these trials.

\textit{Retrieval-assisted parameter selection.} Scores are generally highest when the reference documents are indexed and decrease substantially when the retrieval index is empty (Figure~\ref{fig:rag_score_components}). The \acs{RAG}-off condition receives zero parameter credit by construction because the gated scorer requires retrieved evidence. The comparison between \acs{RAG}-on and Empty \acs{RAG} provides the relevant evidence that indexed document content improves parameter selection.

\textit{\acs{HPC} orchestration.} Among the two proprietary models tested, Gemini-3-Flash completes all four stages of the supplied training workflow in every trial. GPT-5-mini reaches the final evaluation stage in 70\% of trials with explicit prompts and 50\% with natural prompts (Table~\ref{tab:hpc_step_completion}). Its incomplete traces most often stop before the final \emph{evaluate\_model} call, without a recorded timeout or tool exception. Thus, both models can submit and monitor remote jobs, but completion of the full workflow depends on the model and prompt format.

\textit{Training-code authoring.} At T2, mean model-quality scores are 0.74 for GPT-5-mini and 0.81 for Gemini-3-Flash (Table~\ref{tab:tc_results}). Results become more model- and tier-dependent when the agent writes the architecture or sampling function. At T4, GPT-5-mini scores 0.21 under the canonical evaluator but 0.74 when the functional evaluator loads the agent's own class (Figure~\ref{fig:tc_two_track}). This difference identifies reference-interface incompatibility rather than unsuccessful training. At T5-func, the GPT-5-mini sampling function ignores conditioning and receives zero model-quality credit, whereas Gemini-3-Flash scores 0.54.

At the open-synthesis tier, both models implement conditional VAEs rather than the reference \acs{cGAN}. Gemini-3-Flash obtains a mean \acs{MMD} of 0.094, compared with 0.19 for the seed-matched reference, and has lower \acs{MMD} on seven of ten Beams2D seeds (Figure~\ref{fig:tc_novel_designs}). These experiments show that the functional evaluator can assess an agent-authored model from a different generative family. They also show that checkpoint compatibility, sampling behavior, and model quality are separate sources of success or failure. These conclusions are limited to the tested models, prompts, seeds, and engineering problems.

\subsection{Comparison to Prior Work}

Table~\ref{tab:literature_comparison} summarizes the capabilities covered by related systems and benchmarks. The following discussion positions our results against the closest work for each of the four evaluation dimensions.

\paragraph{Workflow Performance}
The closest multi-agent engineering system is the 9-agent \acs{MAS} with Design State Graph (DSG) coordination~\cite{massoudi_agentic_2025}. Direct numerical comparison is not meaningful because the two systems address structurally different tasks: DSG-MAS targets full design-state management across coupled requirement--design--test loops, whereas our benchmark evaluates tool-calling workflows on structured engineering pipelines. We therefore position DSG-MAS as a complementary contribution rather than a numerical baseline; matched-task comparison would require a shared evaluation framework that does not currently exist.

Our workflow benchmark extends existing engineering evaluations by testing multi-step tool orchestration. \acs{FDM}-Bench~\cite{eslaminia_fdm-bench_2024} evaluates \acsp{LLM} on G-code anomaly detection and user queries but does not evaluate connected tool workflows. General-purpose benchmarks such as ToolSandbox~\cite{lu_toolsandbox_2025} and ACEBench~\cite{chen_acebench_2025} report performance differences between open-source and proprietary models. We observe a similar difference between the proprietary models and Qwen3-4B, although Qwen3.5-4B improves substantially over Qwen3-4B at the same parameter count. For multi-turn professional tasks, $\tau$-bench~\cite{yao_tau-bench_2024} reports task success below 50\% and $\text{pass}^8$ below 25\%. These values are not directly comparable with ours because the benchmarks use different tasks, tools, and success criteria.

\paragraph{\acs{RAG} Evaluation}
While \acs{RAG} is a well-established technique~\cite{lewis_retrieval-augmented_2021, gao_retrieval-augmented_2024}, its evaluation in engineering contexts has focused on document comprehension rather than actionable parameter extraction. Doris \textit{et al.}~\cite{doris2024designqa} benchmark multimodal \acs{LLM} understanding of engineering documentation (rule comprehension, compliance, and extraction), but do not evaluate whether retrieved information is correctly applied to downstream tool calls. Our evaluation addresses this gap by testing whether retrieved parameters are used in downstream tool calls. The comparison between populated and empty indexes holds retrieval-tool access constant while varying the indexed content. For P2 and P3, which require parameter sets from the reference papers, no model returned the complete expected set when those documents were absent from the index.

\paragraph{\acs{HPC} Orchestration}
Among the studies reviewed, ALL-FEM~\cite{deotale_all-fem_2026} is the closest work on automated simulation execution. It uses multiple \acs{LLM} agents to generate and execute FEniCS code in Google Colab and reports 71.79\% accuracy across 39 \acs{FEA} benchmarks. ALL-FEM evaluates simulation-code generation and execution in a notebook environment. Our \acs{HPC} evaluation instead records whether an agent generates a \acs{SLURM} script, submits and monitors a remote training job, and evaluates the resulting model. Figure~\ref{fig:hpc_baseline_grid} reports the model-quality metrics from completed runs alongside the corresponding EngiBench references.

\paragraph{Training-Code Authoring}
MLGym, MLAgentBench, and MLE-bench evaluate agents that write or modify machine-learning training code~\cite{nathani2025mlgymnewframeworkbenchmark,3692070.3692884,chan2025mlebench}. Their reported success measures are not directly comparable with ours because they use standard machine-learning datasets or competition scores, whereas our functional evaluator computes \acs{MMD} from generated engineering designs. The closest qualitative comparison is MLGym's finding that agents more often tune existing programs than implement a different model architecture.

Our T6v2 experiment removes the reference architecture and supplies only the required data and evaluation interfaces. Both agents select established conditional VAE architectures rather than the reference \acs{cGAN} and complete the \acs{SLURM} training workflow. Gemini-3-Flash implements a convolutional VAE and obtains a model-quality score of 0.88. Its mean \acs{MMD} is 0.094, compared with 0.19 for the seed-matched reference \acs{cGAN}, and it obtains lower \acs{MMD} on seven of ten seeds. GPT-5-mini implements a fully connected VAE, obtains a model-quality score of 0.70, and does not improve on the reference on average. These results show that, in the tested setting, an agent can implement and train a generative model family different from the supplied reference. They do not establish an algorithm that is novel to the research literature. Unlike the benchmarks above, our evaluation additionally requires the authored program to complete a remote \acs{HPC}/\acs{SLURM} workflow.

\paragraph{General-Purpose Coding Agents}
General-purpose coding agents\textemdash such as OpenAI Codex~\cite{openai_codex_2025}, Claude Code~\cite{anthropic_claude_code_2025}, and OpenClaw~\cite{openclaw_2025}\textemdash can write and execute arbitrary code and may call the EngiBench and EngiOpt \acsp{API} without pre-bound tool abstractions. The task definitions and scoring criteria do not conceptually depend on \textsc{EngiAI}'s seven-agent decomposition. The current software evaluates \textsc{EngiAI} directly; evaluating another system requires an adapter that maps its tool traces and artifacts to the schema expected by the scorers.

\subsection{Implications for Engineering-Agent Development}
Results in Table~\ref{tab:evaluation_framework} support five recommendations for developing and evaluating engineering agents. First, workflow execution, retrieval-assisted parameter selection, remote-job orchestration, and code authoring should be reported separately because a single end-to-end success rate hides distinct failure mechanisms. Second, evaluation should assess both the execution trace and the resulting engineering artifact because valid tool calls can still use the wrong branch, parameter, or trained model. Third, prompts should test ambiguity, competing plausible values, derived quantities, conditional decisions, and retention of multiple parameter sets rather than only explicit step-by-step instructions. Fourth, retrieval effectiveness should be assessed by comparing populated and empty indexes while holding retrieval-tool access constant. A no-retrieval condition may be reported separately to verify agent and scorer behavior, but it should not be treated as evidence of retrieval effectiveness when the scorer requires a retrieval call. Fifth, authored training programs should be evaluated functionally using the agent's own model and sampling interfaces, while compatibility with a reference implementation should be reported separately.

\textsc{EngiAI} is intended as a reference implementation and evaluation testbed, not as a required agent architecture. Researchers can adapt the task definitions and scorers to another supervisor, a single agent, or a coding agent by providing compatible tool, trace, and artifact interfaces. Engineering teams can adapt these tests as acceptance criteria for grounding, tool use, long-running execution, and artifact validity before deploying an agent on a new workflow. The present experiments do not establish readiness for unsupervised production use.

\subsection{Limitations and Threats to Validity}
\label{subsec:limitations}

The evaluation covers two EngiBench problems, Beams2D and Photonics2D, and therefore does not represent the full range of engineering problems available in EngiBench. We also did not conduct a study with practicing engineers. The experiments consequently do not assess interactive usability, the quality of human interventions, or how effectively users can redirect an agent after a failure.

Only four \acs{LLM} backends were evaluated, and the \acs{RAG}, \acs{HPC}, and training-code evaluations use subsets of these models. The \acs{HPC} benchmark is further limited to two generative-model families (\acs{cGAN} and diffusion), two proprietary \acs{LLM} backends (GPT-5-mini and Gemini-3-Flash), one cluster, and one training setup. The corresponding orchestration results should therefore be interpreted within these experimental conditions. Finally, we do not include a matched single-agent baseline with identical tool access. The experiments therefore do not isolate the effect of the multi-agent supervisor decomposition.

Appendix~\ref{app:stats} reports bootstrap confidence intervals and the applicable non-parametric tests for all four evaluations. It also tests the sensitivity of the training-code ranking to the composite weights. We did not perform corresponding weight-sensitivity analyses for the other composite scores, and we did not vary temperature or prompt wording systematically.

The training-code experiments cover one reference model family, two engineering problems, two proprietary \acs{LLM} backends, and one prompt wording. The Photonics2D results provide a second problem instance, but they do not establish generalization to other engineering domains or generative-model families.

Model quality is compared with one reference training run for each seed. Reference \acs{cGAN} \acs{MMD} ranges from approximately 0.03 to 0.59 across the ten Beams2D seeds. An authored model can therefore outperform a weak reference run without outperforming the reference method in general. Repeated reference runs would support a stronger comparison.

T2--T5 retain parts of the public EngiOpt implementation. An agent could reproduce these regions from prior exposure rather than derive them during the trial. T6v2 reduces this concern by withholding the reference architecture, but the selected VAE architectures are established methods. We therefore interpret T6v2 as evidence that the agent can implement a model not constrained to the supplied \acs{cGAN}, not as evidence of a new generative algorithm.

API responses varied slightly across repeated runs despite a sampling temperature of zero. Tables~\ref{tab:workflow_results} and~\ref{tab:workflow_results_photonics} report the resulting standard deviations.


\section{Conclusion and Future Work}
\label{sec:conclusion}
We evaluated four capabilities of \acs{LLM} agents in engineering design workflows: tool execution, retrieval-assisted parameter selection, remote training-job orchestration, and training-code authoring. \textsc{EngiAI} provides the multi-agent, tool-connected implementation used in these experiments.

The experiments yield four main findings. First, GPT-5-mini and Gemini-3-Flash complete 96--97\% of the structured Beams2D workflows, compared with 55\% for Qwen3-4B and 78\% for Qwen3.5-4B. Second, indexed documents improve numerical parameter selection relative to an empty retrieval index. Third, Gemini-3-Flash completes every tested \acs{HPC} pipeline, whereas GPT-5-mini reaches the final evaluation step in 70\% of explicit-prompt trials and 50\% of natural-prompt trials. Fourth, successful orchestration of a supplied training program does not imply uniform success when the agent must write that program. The authoring benchmark identifies separate failures in checkpoint compatibility, sampling behavior, and model quality. At T6v2, both proprietary models implement conditional VAEs rather than the supplied \acs{cGAN}; Gemini-3-Flash obtains lower \acs{MMD} than the seed-matched Beams2D reference in seven of ten trials.

The study covers two engineering problems and four \acs{LLM} backends, but the \acs{RAG}, \acs{HPC}, and authoring evaluations use subsets of those backends. The retrieval index contains only two papers, the authoring benchmark uses two proprietary models and one prompt formulation, and model quality is compared with a single reference run per seed. The results should therefore not be generalized beyond the tested models, prompts, and problems.

Future work should evaluate additional engineering problems, model backends, prompt formulations, and sampling temperatures. Retrieval should be tested with larger and noisier document collections. The authoring benchmark should include other generative-model families, including diffusion models, and compare authored models against distributions of repeated reference runs. A further extension would raise the authoring target from the training program to the simulator itself, supplying a multi-page problem specification as input and scoring the generated simulator against reference solutions and physical validity checks. A matched single-agent baseline is also needed to determine whether the multi-agent decomposition itself improves performance. Finally, experiments with larger tool sets exposed through tool \acs{API}s or \acs{MCP} would test how workflow reliability changes as the system expands.

The principal implication is that engineering agents should not be assessed from successful demonstrations alone. Reliable deployment requires capability-specific tests, controlled information conditions, inspection of intermediate artifacts, and validation of the resulting engineering artifact. \textsc{EngiAI} provides one implementation of this evaluation approach, while the four evaluation dimensions define a basis for future comparisons among systems that expose compatible evidence.

\section*{Conflict of Interest}
There are no conflicts of interest.

\section*{Data Availability Statement}
The code of this project is publicly available on \href{https://github.com/gioelemo/EngiAI-Public}{GitHub}.

\bibliographystyle{plainnat}
\bibliography{references}

\appendix

\section{Compute Environment}
\label{app:compute}

The Ollama-served models use 8-bit quantization with a $32{,}768$-token context window. The multi-agent system and all local (Ollama) inference run on a MacBook Pro (Apple M4~Pro, 24\,GB unified memory); \acs{HPC} training jobs are submitted to the ETH Zurich Euler cluster on NVIDIA RTX~4090 GPU nodes, whose per-node CPU and memory allocations vary across runs.

The \acs{RAG} service (MMORE) combines dense and sparse retrieval. Dense retrieval uses 384-dimensional \texttt{all-MiniLM-L6-v2} embeddings with cosine similarity~\cite{reimers_sentence-bert_2019}; sparse retrieval uses SPLADE, a learned sparse lexical-expansion retriever~\cite{formal2021spladesparselexicalexpansion}, weighted equally ($\alpha=0.5$). The system retrieves five chunks, reranks them with \texttt{bge-reranker-base}, stores vectors in Milvus~\cite{wang_milvus_2021}, and synthesizes answers with Qwen3 8B~\cite{yang_qwen3_2025}.

\section{Scoring Methodology}
\label{app:metrics}

\subsection{Design Quality Metrics}

The design quality score is a weighted combination of six per-design metrics:

\begin{equation}
\begin{split}
S_{\text{design}} = {} & 0.31 \cdot \text{\acs{IoU}} + 0.19 \cdot \text{PA} + 0.15 \cdot \text{Obj} \\
& + 0.12 \cdot \text{Constr} + 0.12 \cdot \text{Conn} + 0.11 \cdot \text{WT}
\end{split}
\end{equation}

\noindent where:
\begin{itemize}
    \item \textbf{\acs{IoU}} (Intersection over Union): overlap between the binarized agent design and ground-truth design, $\text{\acs{IoU}} = |A \cap G| / |A \cup G|$, with threshold 0.5.
    \item \textbf{PA} (Pixel Accuracy): fraction of correctly classified pixels, $\text{PA} = \frac{1}{N}\sum_i \mathbf{1}(A_i = G_i)$, where $N$ is the total number of pixels.
    \item \textbf{Obj} (Objective Score): normalized score based on problem-specific objectives (\textit{e.g.}, compliance, field overlap).
    \item \textbf{Constr} (Constraint Score): smooth partial credit for constraint satisfaction, $\exp(-|v_{\text{actual}} - v_{\text{target}}| / \tau)$ with temperature $\tau$, averaged across all constraints.
    \item \textbf{Conn} (2D Connectivity): validates topological connectivity of the 2D design.
    \item \textbf{WT} (3D Watertightness): validates that the exported 3D mesh forms a closed, watertight solid. In practice, WT is uniformly 0.00 across all configurations because the threshold-and-extrude \acs{STL} export produces non-manifold meshes; a post-processing repair step would be needed for watertight output.
\end{itemize}

Weights were assigned by engineering judgment, prioritizing geometric fidelity over engineering performance metrics and printability. A post-hoc \ac{AHP} analysis \cite{saaty_analytic_1987} confirmed consistency ($\text{CR = 0.012}$) and produced weights within $\pm0.03$ of the chosen values.

The hierarchical workflow score combines design quality with operational metrics:
\begin{equation}
S_{\text{workflow}} = 0.65 \cdot S_{\text{design}} + 0.20 \cdot S_{\text{tool}}  + 0.15 \cdot S_{\text{completion}}
\end{equation}
where $S_{\text{tool}}$ is the tool-call efficiency, defined as the ratio of correctly matched calls to the maximum of optimal and actual call counts, and $S_{\text{completion}} \in \{0, 1\}$ indicates whether all required tools were called successfully. For styles with \acs{STL} export (\textsc{W-Rand} through \textsc{W-Multi}), task completion additionally requires that the export parameters match the prompt specifications within a tolerance of $\pm 0.05$ for floats and exact match for booleans; a successful tool call with incorrect parameters counts as failed. For the \textsc{Natural} style, when the agent correctly requests clarification ($S_{\text{completion}} = 1$) and does not produce a design, $S_{\text{design}}$ is set to $1.0$, reflecting that abstention is the correct behavior for underspecified prompts.

\subsection{\acs{RAG} Scoring Details}
\label{app:rag_metrics}

The \acs{RAG} score is a weighted sum over $n$ evaluated parameters plus a retrieval indicator:
\begin{equation}
S_{\text{RAG}} = \sum_{i=1}^{n} w_i \cdot a_{i}^{\text{eff}} + w_{\text{RAG}} \cdot \mathbf{1}(\text{RAG called})
\end{equation}
where effective accuracy $a_{i}^{\text{eff}} = a_{i}$ if the agent called \emph{search\_documents} and $0$ otherwise, and raw accuracy $a_{i} = \mathbf{1}(|v_{\text{actual}} - v_{\text{expected}}| \leq \tau)$ with tolerance $\tau = 0.05$ for volume fraction and force distance, $\tau = 0.5$ for filter radius. The per-prompt scoring weights (Table~\ref{tab:rag_weights}) decrease per-parameter as more parameters are evaluated (1 to 3 across prompts), while $w_{\text{RAG}}$ shrinks from $0.40$ to $0.10$ since the gating mechanism becomes increasingly redundant when multiple parameters must all be correct.

\begin{table}[ht]
\centering
\caption[\acs{RAG} scoring weights per prompt.]{\acs{RAG} scoring weights per prompt. Target parameters are given in Table~\ref{tab:rag_prompts}.}
\label{tab:rag_weights}
\footnotesize
\begin{tabular}{@{}ccccc@{}}
\toprule
\textbf{ID} & $w_{\text{vol}}$ & $w_{\text{force}}$ & $w_{\text{rmin}}$ & $w_{\text{rag}}$ \\
\midrule
\textsc{P0} & 0.60 & \textemdash & \textemdash & 0.40 \\
\textsc{P1} & 0.40 & 0.40 & \textemdash & 0.20 \\
\textsc{P2} & 0.40 & \textemdash & 0.40 & 0.20 \\
\textsc{P3} & 0.30 & 0.30 & 0.30 & 0.10 \\
\bottomrule
\end{tabular}
\end{table}

\subsection{\acs{HPC} Workflow Score}
\label{app:hpc_metrics}

The primary metric is a weighted composite:

\begin{equation}
\begin{split}
S_{\text{HPC}} = {} & w_{\text{step}} \cdot \frac{n_{\text{completed}}}{4} + w_{\text{config}} \cdot c_{\text{config}} \\
& + w_{\text{eval}} \cdot \min\!\left(1, \frac{n_{\text{metrics}}}{6}\right)
\end{split}
\end{equation}

where $n_{\text{completed}}$ is the number of completed workflow steps (out of 4: generate, submit, monitor, evaluate), $c_{\text{config}} \in \{0, 1\}$ indicates whether the training configuration matches the prompt specifications, and $n_{\text{metrics}}$ counts the evaluation metrics successfully extracted. The base weights are $w_{\text{step}} = 0.70$, $w_{\text{config}} = w_{\text{eval}} = 0.15$.

Each secondary weight ($w_{\text{config}}$, $w_{\text{eval}}$) is set to zero when its corresponding step was not called, with the unused weight redistributed to $w_{\text{step}}$ (\textit{i.e.}, $w_{\text{step}} = 1.0 - w_{\text{config}} - w_{\text{eval}}$). This avoids double penalization: a missing step is already penalized in the step completion rate. The evaluation metrics counted in $n_{\text{metrics}}$ are \acs{IOG}, \acs{COG}, \acs{FOG}, \acs{MMD}, \acs{DPP}, and \acs{RVC}.

\subsection{Training-Code Score}
\label{app:tc_metrics}

\begin{figure*}[t]
  \centering
  \includegraphics[width=\textwidth]{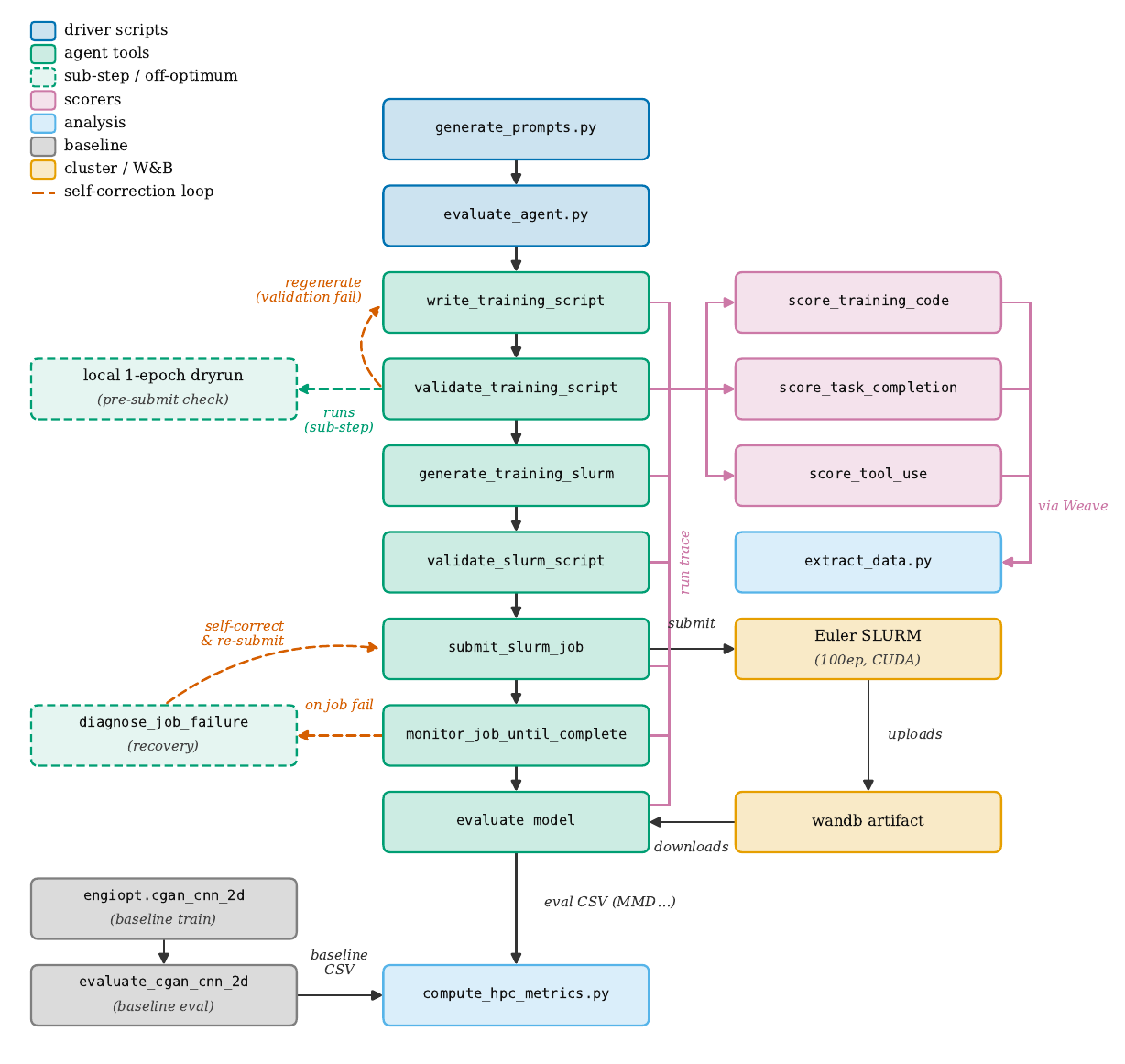}
  \caption[Training-code benchmark pipeline]{Training-code benchmark pipeline. The agent authors and validates a training script, submits and monitors a SLURM job, and evaluates the trained model. Dashed arrows indicate self-correction; gray and purple edges denote execution and scoring flows, respectively.}
  \label{fig:tc_flow}
\end{figure*}

Figure~\ref{fig:tc_flow} gives an overview of the benchmark pipeline end to end. The training-code composite for a single trial is the weighted sum
\begin{equation}
S_{\text{tc}} = 0.55\,q + 0.15\,c + 0.10\,j + 0.10\,d + 0.05\,s + 0.05\,l,
\end{equation}
where $q$ is model quality, $c$ code quality, $j$ job completion, $d$ the one-epoch local dry run, $s$ syntax validity, and $l$ \acs{SLURM} acceptance. All terms lie in $[0,1]$, and every term except $q$ is binary. Code quality $c=1$ when the static validator's structural contracts\textemdash the cross-file symbol, signature, checkpoint-key, and artifact-name checks\textemdash together with its forward-pass check all pass; $s$ marks a syntactically valid script; $l$ marks a submission script accepted by \texttt{sbatch -{}-test-only}; and $j$ and $d$ mark, respectively, a completed full training job on the cluster and a successful one-epoch local execution of the authored script. The weights prioritize the graded design-quality term $q$ over the binary operational gates; a post-hoc \acs{AHP} analysis~\cite{saaty_analytic_1987} confirmed their consistency ($\text{CR = 0.004}$) and produced weights within $\pm0.03$ of the chosen values.

The model-quality term $q$ compares the agent-trained model with the reference EngiOpt model trained using the same seed. If the trial produces a scorable model and a matched reference is available, then
\[
q = \min\!\left(1,\frac{\mathrm{MMD}_{\text{ref}}}
{\mathrm{MMD}_{\text{agent}}}\right).
\]
If no matched reference is available, $q$ and the corresponding composite score are treated as missing. All other trials receive $q=0$. For scoring purposes, a model is considered produced only if its training job completes and its evaluation returns at least one metric. On the functional track, its sampler must also pass the degeneracy check. Requiring completion of the current training job prevents a failed trial from receiving credit for an artifact left by an earlier run.

The comparison is performed separately for each seed because the reference \acs{cGAN} MMD ranges from approximately $0.03$ to $0.59$ across the ten Beams2D seeds. A single reference threshold would therefore not represent the variation among seeds. We record both $\mathrm{MMD}_{\text{ref}}$ and $\mathrm{MMD}_{\text{agent}}$ so that $q$ can be recalculated using another mapping or composite weighting.

On the functional track (the \textsc{-func} tier variants and T6v2), the evaluator instantiates the generative-model class written by the agent and loads the corresponding checkpoint. It then generates conditioned designs using the agent-provided sampling function and computes $\mathrm{MMD}_{\text{agent}}$ with the fixed EngiBench \acs{MMD} evaluator described in Appendix~\ref{app:gen_metrics}. The authored architecture is therefore not required to match the reference implementation. This procedure permits evaluation of the agent-selected architecture at T6v2.

Before model quality is scored, the sampling function is tested for degeneracy. A sampler receives $q=0$ only when its outputs are both nearly constant across latent samples and insensitive to changes in conditioning. Requiring both conditions distinguishes this failure from low variation along only one of these dimensions.

\subsection{Generative Model Quality Metrics}
\label{app:gen_metrics}

For the \acs{HPC} training and training-code benchmarks, we report the following EngiOpt metrics~\cite{felten_engibench_2025} to evaluate trained generative models against the dataset baseline. Let $\mathcal{D}$ denote the dataset designs and $\mathcal{D}_g$ the generated designs.

\begin{itemize}
    \item \textbf{\acs{MMD}} \textemdash Distributional similarity between generated and dataset designs (lower is better):
    \begin{equation}
    \begin{split}
    \text{MMD}^2(k, \mathcal{D}, \mathcal{D}_g) = {} & \mathbb{E}[k(z, z')] + \mathbb{E}[k(\hat{z}, \hat{z}')] \\
    & - 2\mathbb{E}[k(z, \hat{z})]
    \end{split}
    \end{equation}
    where $z, z' \sim \mathcal{D}$, $\hat{z}, \hat{z}' \sim \mathcal{D}_g$, and $k(a, b) = \exp\!\left(-\|a - b\|^2 / (2\sigma^2)\right)$ is a Gaussian kernel with $\sigma = 10.0$.

    \item \textbf{\acs{DPP}} \textemdash Diversity of generated designs (higher is better):
    \begin{equation}
    \text{DPP}(\mathcal{D}_g) = \det(K)
    \end{equation}
    where $K \in \mathbb{R}^{M \times M}$ with $K_{ij} = k(\hat{x}_i, \hat{x}_j)$ for $\hat{x}_i, \hat{x}_j \in \mathcal{D}_g$. In practice, $K$ is regularized as $K + 10^{-6} I$ for numerical stability. In every \acs{HPC} and training-code evaluation, each trained model generates $M{=}50$ designs per seed for the \acs{MMD}/\acs{DPP} computation.

    \item \textbf{\acs{COG}} \textemdash Sum of optimality gaps across the optimization path (lower is better):
    \begin{equation}
    \text{COG}(x_0) = \sum_{t=0}^{T} \left[\tilde{f}(x_t, c) - f^*\right]
    \end{equation}
    where $f^* = \max_{x \in \mathcal{X}} \tilde{f}(x, c)$ is the optimal objective value over the feasible design space $\mathcal{X}$, $c$ denotes the conditioning context (\textit{e.g.}, boundary conditions and loads), $x_0$ is the generated design refined through $T$ optimization steps, and $\tilde{f}$ denotes the simulated objective.

    \item \textbf{\acs{RVC}} \textemdash Fraction of designs violating at least one constraint (lower is better):
    \begin{equation}
    \begin{split}
    \text{RVC}(\mathcal{D}_g) = \frac{1}{M} \sum_{k=1}^{M} \mathbf{1}\big( & \exists\, i: g_i(\hat{x}_k, a(\hat{x}_k), c) > 0 \\
    & \text{or}\; \exists\, j: h_j(\hat{x}_k, a(\hat{x}_k), c) \neq 0 \big)
    \end{split}
    \end{equation}
    where $M = |\mathcal{D}_g|$ is the number of generated designs, $a(\hat{x}_k)$ denotes the attributes of design $\hat{x}_k$, $c$ the conditions, $g_i$ and $h_j$ are inequality and equality constraints, respectively, and $\mathbf{1}(\cdot)$ is the indicator function.

    \item \textbf{\acs{IOG}} \textemdash The $t{=}0$ term of the \acs{COG} sum, \textit{i.e.}, $\tilde{f}(x_0, c) - f^*$.

    \item \textbf{\acs{FOG}} \textemdash The $t{=}T$ term of the \acs{COG} sum, \textit{i.e.}, $\tilde{f}(x_T, c) - f^*$.
\end{itemize}

\section{Benchmark Prompts}
\label{app:prompts}

Representative prompts for each evaluation dimension. Workflow prompts use Beams2D with volfrac$=$0.4, forcedist$=$0.65, rmin$=$4.0, seed$=$42; the seven prompt styles target distinct cognitive demands as described in Section~\ref{sec:workflow_eval}. \acs{HPC} training and \acs{RAG} evaluation prompts follow.

\subsection{Workflow Prompts}

\begin{promptbox}{\textsc{Full}}
\begin{lstlisting}[style=prompt]
Design a 2D beam structure.

Design requirements:
- Use a material volume fraction of 0.4
- Force distance parameter: 0.65
- Minimum filter radius (rmin): 4.0

Optimize the structure and simulate the result to obtain the compliance value.
\end{lstlisting}
\end{promptbox}

\begin{promptbox}{\textsc{Natural}}
\begin{lstlisting}[style=prompt]
Design a 2D beam structure.

Design requirements:
- The design should be moderate material usage
- Apply a force distributed in the upper right region
Optimize the structure and simulate the result to obtain the compliance value.
\end{lstlisting}
\end{promptbox}

\begin{promptbox}{\textsc{W-Rand}}
\begin{lstlisting}[style=prompt]
Execute a 2D topology optimization, simulate the result, and export the geometry as a 3D-printable STL file.

1. Optimization Configuration
   - Volume Fraction: 0.4
   - Force Distance: 0.65
   - Filter Radius (rmin): 4.0
   - Objective: Minimize compliance

2. Simulation
   - After optimization, simulate the design to obtain the compliance value

3. Post-processing & Export
   - Thresholding: Apply a 0.58 density threshold to convert the continuous density map into binary geometry
   - Mirror: Mirror the design across the y-axis for the final geometry
   - XY Scaling: Scale the X and Y dimensions by 2.47
   - Extrusion: Extrude the 2D result by 17.9 units in the Z-axis to create a 3D volume
   - Export: Save the final geometry as an STL file with these exact parameters
\end{lstlisting}
\end{promptbox}

\begin{promptbox}{\textsc{W-Derived}}
\begin{lstlisting}[style=prompt]
Execute a 2D topology optimization, simulate the result, and export the geometry as a 3D-printable STL file.

1. Optimization Configuration
   - Volume Fraction: 0.4
   - Force Distance: 0.65
   - Filter Radius (rmin): 4.0
   - Objective: Minimize compliance

2. Simulation
   - After optimization, simulate the design to obtain the compliance value

3. Post-processing & Export
   The STL export parameters must be derived from the optimization inputs:
   - Thresholding: Use the volume fraction value as the density threshold
   - Mirror: Mirror the design across the y-axis only if the volume fraction is greater than 0.4
   - XY Scaling: Scale the X and Y dimensions by twice the filter radius
   - Extrusion: Extrude the 2D result in the Z-axis by the threshold value multiplied by 40
   - Export: Save the final geometry as an STL file with these derived parameters
\end{lstlisting}
\end{promptbox}

\begin{promptbox}{\textsc{W-Distract}}
\begin{lstlisting}[style=prompt]
Execute a 2D topology optimization, simulate the result, and export the geometry as a 3D-printable STL file.

1. Optimization Configuration
   - Volume Fraction: 0.4
   - Force Distance: 0.65
   - Filter Radius (rmin): 4.0
   - Objective: Minimize compliance

2. Simulation
   - After optimization, simulate the design to obtain the compliance value

3. Post-processing & Export
   - Threshold the density field at 0.42 to preview the design topology
   - Apply a 0.58 density threshold to produce the final solid/void geometry
   - Scale the preview display by 1.76x in XY for quick inspection
   - Scale the X and Y dimensions of the part by 2.47 for manufacturing
   - Mirror the design across the y-axis for the final geometry
   - Extrude the 2D result by 17.9 units in the Z-axis to create a 3D volume
   - Export: Save the final geometry as an STL file with these exact parameters
\end{lstlisting}
\end{promptbox}

\begin{promptbox}{\textsc{W-Cond}}
\begin{lstlisting}[style=prompt]
Execute a 2D topology optimization, simulate the result, then export the geometry as a 3D-printable STL file with parameters that depend on the simulation outcome.

1. Optimization Configuration
   - Volume Fraction: 0.4
   - Force Distance: 0.65
   - Filter Radius (rmin): 4.0
   - Objective: Minimize compliance

2. Simulation
   - After optimization, simulate the design to obtain the compliance value

3. Post-processing & Export (conditional on compliance)
   - If compliance > 254.8:
     - Thresholding: Apply a 0.48 density threshold to convert the continuous density map into binary geometry
     - Mirror: Mirror the design across the y-axis for the final geometry
   - If compliance <= 254.8:
     - Thresholding: Apply a 0.64 density threshold to convert the continuous density map into binary geometry
     - Mirror: Do NOT mirror the design for the final geometry
   - In both cases:
     - XY Scaling: Scale the X and Y dimensions by 0.92
     - Extrusion: Extrude the 2D result by 19.6 units in the Z-axis to create a 3D volume
   - Export: Save the final geometry as an STL file with these exact parameters
\end{lstlisting}
\end{promptbox}

\begin{promptbox}{\textsc{W-Multi}}
\begin{lstlisting}[style=prompt]
Execute a 2D topology optimization, simulate the result, and export the geometry as TWO separate 3D-printable STL files with different parameters.

1. Optimization Configuration
   - Volume Fraction: 0.4
   - Force Distance: 0.65
   - Filter Radius (rmin): 4.0
   - Objective: Minimize compliance

2. Simulation
   - After optimization, simulate the design to obtain the compliance value

3. Post-processing & Export

   Export A:
   - Thresholding: Apply a 0.48 density threshold to convert the continuous density map into binary geometry
   - Mirror: Mirror the design across the y-axis for the final geometry
   - XY Scaling: Scale the X and Y dimensions by 3.64
   - Extrusion: Extrude the 2D result by 19.6 units in the Z-axis to create a 3D volume
   - Export: Save the final geometry as an STL file with these exact parameters

   Export B:
   - Thresholding: Apply a 0.64 density threshold to convert the continuous density map into binary geometry
   - Mirror: Do NOT mirror the design for the final geometry
   - XY Scaling: Scale the X and Y dimensions by 0.92
   - Extrusion: Extrude the 2D result by 16.4 units in the Z-axis to create a 3D volume
   - Export: Save the final geometry as an STL file with these exact parameters
\end{lstlisting}
\end{promptbox}

\subsection{RAG Evaluation Prompts}

\begin{promptbox}{\textsc{P0}}
\begin{lstlisting}[style=prompt]
The EngiBench paper documents the default design conditions for the Beams2D problem in its API walkthrough.

Search the paper to find the default volume fraction (volfrac) listed for the Beams2D problem. Then generate a 2D beam design using exactly that volume fraction. Use default values for all other parameters (do NOT ask for clarification -- proceed directly with defaults).
\end{lstlisting}
\end{promptbox}

\begin{promptbox}{\textsc{P1}}
\begin{lstlisting}[style=prompt]
In the EngiBench paper's Section 3.1 API walkthrough, a code example runs a Beams2D optimization using non-default design conditions. Search the paper to find both the volume fraction and force distance from that example, then generate a 2D beam design with those exact values. Use default values for all other parameters and do not ask for clarification.
\end{lstlisting}
\end{promptbox}

\begin{promptbox}{\textsc{P2}}
\begin{lstlisting}[style=prompt]
The SOPTX paper by He et al. (2025) benchmarks its topology optimization framework on a 2D cantilever beam problem.

Search the paper to find both the volume fraction (volfrac) and the filter radius (rmin) used for that 2D cantilever benchmark. Then generate a 2D beam design using exactly those values. Use default values for all other parameters and do not ask for clarification.
\end{lstlisting}
\end{promptbox}

\begin{promptbox}{\textsc{P3}}
\begin{lstlisting}[style=prompt]
Generate a 2D beam design combining parameters from multiple sources:

1. Use the volume fraction and force distance from the EngiBench paper's API walkthrough example (the non-default values shown in the code snippet).
2. Use the filter radius from the SOPTX paper by He et al. (2025) for their 2D cantilever beam benchmark.

Search the relevant papers to find each value, then generate a 2D beam design using exactly those three parameters. Use default values for all other parameters and do not ask for clarification.
\end{lstlisting}
\end{promptbox}

\subsection{HPC Training Prompts}

\begin{promptbox}{\textsc{HPC-Train (Explicit)}}
\begin{lstlisting}[style=prompt]
Train a cGAN CNN 2D generative model for the Beams2D topology optimization problem on the Euler HPC cluster, then evaluate it against the dataset baseline using the standard EngiOpt evaluation script.

Step 1: Generate Training Script
   - Use the generate_training_command tool with:
     algorithm: cgan_cnn_2d
     problem_id: beams2d
     epochs: 100
     seed: 1

Step 2: Submit to HPC
   - Submit the generated SLURM script to the Euler cluster

Step 3: Monitor Training
   - Monitor the job until it completes
   - Use check_interval=30 and max_checks=200 for the monitoring

Step 4: Evaluate Trained Model
   - Use the evaluate_model tool to evaluate the trained model
     against the dataset baseline:
     problem_id: beams2d
     algorithm: cgan_cnn_2d
     seed: 1
     n_samples: 50
   - This downloads the model from WandB, generates designs, and
     computes metrics (IOG, COG, FOG, MMD, DPP, violation rate)
   - Report the evaluation metrics from the output

Complete all steps in order. Do not ask for clarification.
\end{lstlisting}
\end{promptbox}

\begin{promptbox}{\textsc{HPC-Train (Natural)}}
\begin{lstlisting}[style=prompt]
Train a cGAN CNN 2D model for the Beams2D topology optimization problem on the Euler HPC cluster with seed 1 and 100 epochs. Use the available tools to generate the SLURM training script -- do not write or modify any scripts manually. Submit the job and wait for it to finish. Then use the model evaluation tool to evaluate the trained model against the dataset -- it will download the model from WandB automatically. Report the metrics.

Do not ask for clarification.
\end{lstlisting}
\end{promptbox}

\subsection{Training-Code Authoring Prompts}
\label{app:tc_prompts}

Representative prompts (seed~1) for six training-code authoring tiers and variants (T2, T3, T4, T4-Func, T5, T6v2). The shared blanked EngiOpt \acs{cGAN} training file embedded in each prompt is elided; Figure~\ref{fig:tc_tiers} shows which regions each tier authors.

\begin{promptbox}{\textsc{TC-T2}}
\begin{lstlisting}[style=prompt]
You are given a partially-blanked EngiOpt training file for the Beams2D topology-optimization problem. Complete the blanked regions so the file trains a working conditional generative model, then train and evaluate it on the Euler HPC cluster.

Tier: T2 (plumbing). Blanked regions: B6, B7.

Contract:
Fill the blanked plumbing (sampling helper, logging) so the training loop runs end-to-end on Beams2D.

Training file to complete (fill every BLANK; keep all preserved code verbatim, including the `Generator = NetG` alias and the checkpoint block):

[ 380-line embedded training-file scaffold omitted -- see the tier-explainer figure ]

After completing the file, validate it locally (validation runs a 1-epoch local dry-run), then submit a single full training job to the Euler HPC cluster with seed 1 -- do not submit a separate cluster dry-run job. Monitor the job to completion. Then evaluate the trained model by calling the `evaluate_model` tool (which runs the EngiBench metric harness) -- do NOT compute your own metrics. Report the metrics it returns. Do not ask for clarification.
\end{lstlisting}
\end{promptbox}

\begin{promptbox}{\textsc{TC-T3}}
\begin{lstlisting}[style=prompt]
You are given a partially-blanked EngiOpt training file for the Beams2D topology-optimization problem. Complete the blanked regions so the file trains a working conditional generative model, then train and evaluate it on the Euler HPC cluster.

Tier: T3 (training loop). Blanked regions: B4a, B4b, B5.

Contract:
Fill the blanked training loop (BCE loss, two Adam optimizers, the alternating cGAN generator/discriminator update).

Training file to complete (fill every BLANK; keep all preserved code verbatim, including the `Generator = NetG` alias and the checkpoint block):

[ 401-line embedded training-file scaffold omitted -- see the tier-explainer figure ]

After completing the file, validate it locally (validation runs a 1-epoch local dry-run), then submit a single full training job to the Euler HPC cluster with seed 1 -- do not submit a separate cluster dry-run job. Monitor the job to completion. Then evaluate the trained model by calling the `evaluate_model` tool (which runs the EngiBench metric harness) -- do NOT compute your own metrics. Report the metrics it returns. Do not ask for clarification.
\end{lstlisting}
\end{promptbox}

\begin{promptbox}{\textsc{TC-T4}}
\begin{lstlisting}[style=prompt]
You are given a partially-blanked EngiOpt training file for the Beams2D topology-optimization problem. Complete the blanked regions so the file trains a working conditional generative model, then train and evaluate it on the Euler HPC cluster.

Tier: T4 (architecture). Blanked regions: B3.G.init, B3.G.forward, B3.D.init, B3.D.forward.

Contract:
Implement the blanked NetG and NetD CNN class bodies; honour the constructor signatures and forward-pass shape contract that remain.

Training file to complete (fill every BLANK; keep all preserved code verbatim, including the `Generator = NetG` alias and the checkpoint block):

[ 348-line embedded training-file scaffold omitted -- see the tier-explainer figure ]

After completing the file, validate it locally (validation runs a 1-epoch local dry-run), then submit a single full training job to the Euler HPC cluster with seed 1 -- do not submit a separate cluster dry-run job. Monitor the job to completion. Then evaluate the trained model by calling the `evaluate_model` tool (which runs the EngiBench metric harness) -- do NOT compute your own metrics. Report the metrics it returns. Do not ask for clarification.
\end{lstlisting}
\end{promptbox}

\begin{promptbox}{\textsc{TC-T4-Func}}
\begin{lstlisting}[style=prompt]
You are given a partially-blanked EngiOpt training file for the Beams2D topology-optimization problem. Complete the blanked regions so the file trains a working conditional generative model, then train and evaluate it on the Euler HPC cluster.

Tier: T4-FUNC (architecture + sampler). Blanked regions: B3.G.init, B3.G.forward, B3.D.init, B3.D.forward, B9.sample.

Contract:
Implement the blanked NetG and NetD CNN class bodies, AND author the blanked top-level sample(model, conditions_tensor, n_samples, latent_dim, design_shape, device) that returns n_samples designs of shape (n_samples, *design_shape) in [0, 1]. Your design need not reproduce the canonical architecture exactly -- it is scored on the quality of the designs your own sampler produces.

Training file to complete (fill every BLANK; keep all preserved code verbatim, including the `Generator = NetG` alias and the checkpoint block):

[ 352-line embedded training-file scaffold omitted -- see the tier-explainer figure ]

After completing the file, validate it locally with `validate_training_script(require_sample=True)` (validation runs a 1-epoch local dry-run AND checks your sample() adapter), then submit a single full training job to the Euler HPC cluster with seed 1 -- do not submit a separate cluster dry-run job. Monitor the job to completion. Then evaluate by calling `evaluate_model(eval_mode='functional', candidate_path=<the file you saved with write_training_script>)` -- it loads YOUR Generator class and runs YOUR sample() adapter, then scores the resulting designs with the EngiBench metric oracle -- do NOT compute your own metrics. Report the metrics it returns. Do not ask for clarification.
\end{lstlisting}
\end{promptbox}

\begin{promptbox}{\textsc{TC-T5}}
\begin{lstlisting}[style=prompt]
You are given a partially-blanked EngiOpt training file for the Beams2D topology-optimization problem. Complete the blanked regions so the file trains a working conditional generative model, then train and evaluate it on the Euler HPC cluster.

Tier: T5 (whole module (stretch)). Blanked regions: B1c, B3.G.init, B3.G.forward, B3.D.init, B3.D.forward, B4a, B4b, B5, B6, B7.

Contract:
Complete the whole module from the skeleton into a working conditional CNN-GAN.

Training file to complete (fill every BLANK; keep all preserved code verbatim, including the `Generator = NetG` alias and the checkpoint block):

[ 272-line embedded training-file scaffold omitted -- see the tier-explainer figure ]

After completing the file, validate it locally (validation runs a 1-epoch local dry-run), then submit a single full training job to the Euler HPC cluster with seed 1 -- do not submit a separate cluster dry-run job. Monitor the job to completion. Then evaluate the trained model by calling the `evaluate_model` tool (which runs the EngiBench metric harness) -- do NOT compute your own metrics. Report the metrics it returns. Do not ask for clarification.
\end{lstlisting}
\end{promptbox}

The uppercase term \texttt{NOVEL} in the following prompt is retained verbatim from the prompt used in the evaluation. In this context, it instructs the agent to select a model architecture that is not constrained to the supplied reference \acs{cGAN}. It does not denote or require an algorithm that is novel relative to the research literature.
\begin{promptbox}{\textsc{TC-T6v2}}
\begin{lstlisting}[style=prompt]
Design and implement a NOVEL generative training algorithm for the Beams2D topology-optimization problem, from scratch -- there is NO template to fill in.

Tier: T6V2 (novel algorithm + self-declared interface).

Interface specification (your training module MUST satisfy this):
Design a NOVEL conditional generative model for Beams2D from scratch -- ANY family (GAN / VAE / normalizing flow / diffusion / energy-based / your own). Unlike the fixed-cGAN tier, you DECLARE your OWN interface via a top-level `CONTRACT = {...}` literal dict the eval reads; your module MUST satisfy what it declares:
  - CONTRACT (a literal dict, values must be plain literals) with fields: family, model_class, conditioning_layout, checkpoint_filename, checkpoint_key, artifact_suffix, sample_args, latent_dim.
  - a top-level class named by CONTRACT['model_class'] (an nn.Module) with `__init__(latent_dim, n_conds, design_shape)` -- this EXACT positional signature, because the eval builds it as model_class(latent_dim, n_conds, design_shape).
  - a top-level `sample(...)` returning (n_samples, *design_shape) in [0, 1]; its parameter names (declared, in order, in CONTRACT['sample_args']) are drawn from {conditions, conditions_tensor, design_shape, device, latent_dim, model, n_samples} -- the eval binds them BY NAME (a subset / reordering is fine; `conditions` and `conditions_tensor` are the SAME routed tensor, so pick ONE).
  - CONTRACT['conditioning_layout'] is one of {flat_2d, none, seq_3d, spatial_4d}: the eval reshapes the raw (B, n_conds) conditions into it before your sample() (spatial_4d=(B,n,1,1), flat_2d=(B,n), seq_3d=(B,1,n), none=unchanged).
  - checkpoint (when --save-model): `th.save({"<checkpoint_key>": model.state_dict()}, "<checkpoint_filename>")`, then log a wandb artifact `f"{args.problem_id}_{args.algo}_<artifact_suffix>"` with alias `f"seed_{args.seed}"`. Use the STRING LITERALS matching your CONTRACT (checkpoint_key / checkpoint_filename / artifact_suffix) -- the validator scans the source statically. Everything sample() needs must live in that ONE checkpointed nn.Module.
  - log `latent_dim` to the wandb run config (the eval reads it back from run.config, not the checkpoint).
Train it on Beams2D. You are scored on the QUALITY of the designs your sampler produces (via evaluate_model), NOT on matching any reference.

Provided I/O harness -- use this CLI / dataset-loading / checkpoint plumbing VERBATIM (it is the boilerplate every engiopt script shares; do NOT reinvent or reverse-engineer the EngiBench dataset API). Design your OWN Generator, sample(), loss, and training loop in the marked section -- that is where the novelty must be:

[ 194-line embedded training-file scaffold omitted -- see the tier-explainer figure ]

Write the complete training script with `write_training_script`, then validate it locally with `validate_training_script(require_sample=True)` (validation runs a 1-epoch local dry-run AND checks your sample() adapter), then submit a single full training job to the Euler HPC cluster with seed 1 -- do not submit a separate cluster dry-run job. Monitor the job to completion. Then evaluate by calling `evaluate_model(eval_mode='functional', candidate_path=<the file you saved with write_training_script>)` -- it loads YOUR Generator class and runs YOUR sample() adapter, then scores the resulting designs with the EngiBench metric oracle -- do NOT compute your own metrics. Report the metrics it returns. Do not ask for clarification.
\end{lstlisting}
\end{promptbox}

\section{Additional Results}
\label{app:supplementary}

\subsection{Statistical Analysis}
\label{app:stats}

We summarize the combined-overall scores using medians and 95\% percentile-bootstrap confidence intervals. Each interval is calculated from $9{,}999$ resamples of the trials within that cell. We use non-parametric procedures because the scores are bounded in $[0,1]$ and contain many tied values. Table~\ref{tab:stats_workflow} reports the resulting medians and confidence intervals.

All comparisons preserve the pairing of trials by dataset sample and random seed. Within each prompt style, model pairs are compared using the Wilcoxon signed-rank test. Within each model backend, adjacent prompt styles are compared using the same test. We report the rank-biserial correlation to indicate the direction and magnitude of each paired difference. The Benjamini--Hochberg procedure controls for multiple comparisons separately within the model-comparison and style-comparison families. We additionally use the Friedman test~\cite{cad058f0-09f2-35e2-b8ad-9038985e0961} as an uncorrected screening test for differences among prompt styles within each backend. Where a prompt ordering was specified in advance, the Page test evaluates whether the scores follow that ordering.

\begin{table}[htbp]
\centering
\caption[Workflow per-cell medians and 95\% CIs]{Per-cell median combined-overall score with 95\% bootstrap confidence intervals for the workflow benchmark (percentile bootstrap over the $n$ trials in each cell; $n$ below 15 reflects the lost Qwen3.5-4B runs noted in Section~4).}
\label{tab:stats_workflow}
\small
\begin{tabular}{llccc}
\toprule
Model & Style & $n$ & Median & 95\% CI \\
\midrule
\multicolumn{5}{l}{\emph{Photonics2D}} \\
GPT-5-mini & \textsc{W-Rand} & 15 & $0.528$ & $[0.520,\,0.541]$ \\
GPT-5-mini & \textsc{W-Distract} & 15 & $0.522$ & $[0.511,\,0.535]$ \\
GPT-5-mini & \textsc{W-Cond} & 15 & $0.452$ & $[0.394,\,0.528]$ \\
Gemini-3-Flash & \textsc{W-Rand} & 15 & $0.597$ & $[0.592,\,0.612]$ \\
Gemini-3-Flash & \textsc{W-Distract} & 15 & $0.547$ & $[0.537,\,0.551]$ \\
Gemini-3-Flash & \textsc{W-Cond} & 15 & $0.593$ & $[0.447,\,0.613]$ \\
Qwen3-4B & \textsc{W-Rand} & 15 & $0.550$ & $[0.542,\,0.565]$ \\
Qwen3-4B & \textsc{W-Distract} & 15 & $0.514$ & $[0.511,\,0.531]$ \\
Qwen3-4B & \textsc{W-Cond} & 15 & $0.397$ & $[0.385,\,0.411]$ \\
Qwen3.5-4B & \textsc{W-Rand} & 15 & $0.547$ & $[0.541,\,0.557]$ \\
Qwen3.5-4B & \textsc{W-Distract} & 15 & $0.552$ & $[0.540,\,0.608]$ \\
Qwen3.5-4B & \textsc{W-Cond} & 15 & $0.465$ & $[0.402,\,0.559]$ \\
\midrule
\multicolumn{5}{l}{\emph{Beams2D}} \\
GPT-5-mini & \textsc{W-Rand} & 15 & $0.674$ & $[0.599,\,0.717]$ \\
GPT-5-mini & \textsc{W-Derived} & 15 & $0.646$ & $[0.594,\,0.694]$ \\
GPT-5-mini & \textsc{W-Distract} & 15 & $0.637$ & $[0.599,\,0.668]$ \\
GPT-5-mini & \textsc{W-Cond} & 15 & $0.694$ & $[0.606,\,0.739]$ \\
GPT-5-mini & \textsc{W-Multi} & 15 & $0.696$ & $[0.595,\,0.744]$ \\
Gemini-3-Flash & \textsc{W-Rand} & 15 & $0.646$ & $[0.586,\,0.689]$ \\
Gemini-3-Flash & \textsc{W-Derived} & 15 & $0.612$ & $[0.586,\,0.694]$ \\
Gemini-3-Flash & \textsc{W-Distract} & 15 & $0.629$ & $[0.556,\,0.689]$ \\
Gemini-3-Flash & \textsc{W-Cond} & 15 & $0.662$ & $[0.622,\,0.739]$ \\
Gemini-3-Flash & \textsc{W-Multi} & 15 & $0.696$ & $[0.674,\,0.759]$ \\
Qwen3-4B & \textsc{W-Rand} & 15 & $0.637$ & $[0.599,\,0.700]$ \\
Qwen3-4B & \textsc{W-Derived} & 15 & $0.518$ & $[0.487,\,0.582]$ \\
Qwen3-4B & \textsc{W-Distract} & 15 & $0.599$ & $[0.557,\,0.664]$ \\
Qwen3-4B & \textsc{W-Cond} & 15 & $0.556$ & $[0.475,\,0.700]$ \\
Qwen3-4B & \textsc{W-Multi} & 15 & $0.647$ & $[0.566,\,0.740]$ \\
Qwen3.5-4B & \textsc{W-Rand} & 14 & $0.684$ & $[0.662,\,0.733]$ \\
Qwen3.5-4B & \textsc{W-Derived} & 13 & $0.681$ & $[0.606,\,0.697]$ \\
Qwen3.5-4B & \textsc{W-Distract} & 15 & $0.637$ & $[0.624,\,0.748]$ \\
Qwen3.5-4B & \textsc{W-Cond} & 15 & $0.662$ & $[0.486,\,0.739]$ \\
Qwen3.5-4B & \textsc{W-Multi} & 14 & $0.683$ & $[0.662,\,0.742]$ \\
\bottomrule
\end{tabular}
\end{table}

On Photonics2D, 14 of the 18 pairwise model comparisons remain significant after BH correction ($q<0.05$). Gemini-3-Flash has higher scores than the other three backends on the random and conditional styles and higher scores than GPT-5-mini and Qwen3-4B on the distractor style (rank-biserial correlation ${\geq}\,0.75$, $q<0.02$). Qwen3.5-4B has higher scores than Qwen3-4B on the distractor and conditional styles ($q<0.02$) and higher scores than GPT-5-mini on the distractor and random styles ($q<0.004$).

The conditional style produces lower scores than the distractor style for GPT-5-mini and both Qwen models ($q<0.04$). The Friedman screening test also identifies an effect of prompt style for three of the four backends ($p<0.05$).

On Beams2D, none of the 30 pairwise model comparisons remains significant after correction. Most median scores are near $0.6$--$0.7$, and the observations contain many ties, which limits the power of the paired tests. The absence of significant pairwise differences should therefore not be interpreted as evidence that the models are equivalent. The Friedman test identifies an effect of prompt style for three backends, but the Page test does not support the predefined monotonic ordering of prompt difficulty ($p>0.7$).

Table~\ref{tab:stats_rag} reports the retrieval results. With a populated index (\acs{RAG}-on), the median score is $1.00$ for GPT-5-mini and Gemini-3-Flash and $0.40$ for Qwen3-4B. With an empty index (Empty \acs{RAG}), the median score is $0.20$ for all three backends. Because both conditions provide access to the retrieval tool, their comparison isolates the effect of making the indexed documents available.

\acs{RAG}-off has a median score of zero by construction because the scorer requires a retrieval call before awarding parameter credit. This condition verifies the scoring gate but does not provide statistical evidence that indexed retrieval improves parameter selection.

Each model--condition cell was designed to contain 12 trials: four prompts with three repetitions each. One GPT-5-mini \acs{RAG}-off trial is missing, leaving $n=11$ in that cell. The pooled summaries combine prompts with different parameter sources and retrieval requirements. They should therefore be interpreted descriptively rather than as estimates for a homogeneous task population.

\begin{table}[htbp]
\centering
\caption[RAG per-cell medians and 95\% CIs]{Median combined-overall score with 95\% bootstrap confidence intervals for the \acs{RAG} benchmark. The planned design contains twelve trials per cell (four prompts with three repetitions); GPT-5-mini \acs{RAG}-off contains eleven. \acs{RAG}-off is identically zero because the gated scorer denies parameter credit without a retrieval call.}
\label{tab:stats_rag}
\small
\begin{tabular}{llccc}
\toprule
Model & Retrieval & $n$ & Median & 95\% CI \\
\midrule
\multicolumn{5}{l}{\emph{GPT-5-mini}} \\
GPT-5-mini & RAG-on & 12 & $1.000$ & $[0.850,\,1.000]$ \\
GPT-5-mini & Empty-RAG & 12 & $0.200$ & $[0.150,\,0.300]$ \\
GPT-5-mini & RAG-off & 11 & $0.000$ & $[0.000,\,0.000]$ \\
\midrule
\multicolumn{5}{l}{\emph{Gemini-3-Flash}} \\
Gemini-3-Flash & RAG-on & 12 & $1.000$ & $[0.600,\,1.000]$ \\
Gemini-3-Flash & Empty-RAG & 12 & $0.200$ & $[0.150,\,0.600]$ \\
Gemini-3-Flash & RAG-off & 12 & $0.000$ & $[0.000,\,0.000]$ \\
\midrule
\multicolumn{5}{l}{\emph{Qwen3-4B}} \\
Qwen3-4B & RAG-on & 12 & $0.400$ & $[0.300,\,0.700]$ \\
Qwen3-4B & Empty-RAG & 12 & $0.200$ & $[0.150,\,0.300]$ \\
Qwen3-4B & RAG-off & 12 & $0.000$ & $[0.000,\,0.000]$ \\
\bottomrule
\end{tabular}
\end{table}

Table~\ref{tab:stats_hpc} reports the \acs{HPC} results over ten trials per cell. Gemini-3-Flash has a median $S_{\mathrm{HPC}}$ of $1.00$ for all three evaluated prompt conditions. For GPT-5-mini, the median decreases from $1.00$ with the explicit \acs{cGAN} prompt to $0.89$ with the natural \acs{cGAN} prompt. Its median is $0.81$ for the natural diffusion prompt; diffusion was not evaluated with an explicit prompt. The wider GPT-5-mini intervals reflect greater variation among trials.

\begin{table}[htbp]
\centering
\caption[HPC per-cell medians and 95\% CIs]{Per-cell median \acs{HPC} workflow score $S_{\text{HPC}}$ with 95\% bootstrap confidence intervals (percentile bootstrap over the ten samples per cell).}
\label{tab:stats_hpc}
\small
\begin{tabular}{llccc}
\toprule
Model & Prompt & $n$ & Median & 95\% CI \\
\midrule
\multicolumn{5}{l}{\emph{GPT-5-mini}} \\
GPT-5-mini & Explicit, cGAN & 10 & $1.000$ & $[0.787,\,1.000]$ \\
GPT-5-mini & Natural, cGAN & 10 & $0.894$ & $[0.575,\,1.000]$ \\
GPT-5-mini & Natural, diffusion & 10 & $0.806$ & $[0.787,\,1.000]$ \\
\midrule
\multicolumn{5}{l}{\emph{Gemini-3-Flash}} \\
Gemini-3-Flash & Explicit, cGAN & 10 & $1.000$ & $[1.000,\,1.000]$ \\
Gemini-3-Flash & Natural, cGAN & 10 & $1.000$ & $[1.000,\,1.000]$ \\
Gemini-3-Flash & Natural, diffusion & 10 & $1.000$ & $[1.000,\,1.000]$ \\
\bottomrule
\end{tabular}
\end{table}

Table~\ref{tab:stats_tc} reports the training-code results over ten seeds per cell. Several T4 and T5 model-quality distributions are bimodal: successfully evaluated checkpoints receive scores near the upper end of the range, whereas checkpoints that cannot be loaded receive zero model-quality credit. The medians and bootstrap intervals provide only a partial summary of these distributions; Figure~\ref{fig:tc_score_grid} and Figure~\ref{fig:tc_score_grid_photonics2d} show the individual seed-level results.

For GPT-5-mini, the composite score differs across T2--T5 (Friedman $p=0.046$), and the Page test supports a decrease over these tiers ($p=0.016$). Gemini-3-Flash also decreases from T2 through T5, but its score increases again at T6v2. Although the Page test over T2--T5--T6v2 gives $p=0.031$, the observed sequence is not monotonic. We therefore do not interpret this test as evidence of a general decline with increasing authoring scope.

The code-quality component is constant across the Gemini-3-Flash tiers (pooled standard deviation $0.00$), whereas the model-quality component varies across tiers for both backends (pooled standard deviation $0.36$--$0.45$). For GPT-5-mini, the model-quality component differs significantly among tiers (Friedman $p=0.033$). Thus, most of the variation in the composite score arises from model quality rather than the operational components.

The weight-sensitivity analysis recomputes and ranks the fourteen model--tier cells under alternative composite weights. The rankings remain similar when model quality is increased or decreased (Kendall $\tau\geq0.87$) and when the operational components receive greater weight ($\tau=1.0$). Removing model quality changes the ranking substantially ($\tau=0.24$). The conclusions are therefore stable under the tested weight changes, except when model quality is omitted entirely.

\begin{table}[htbp]
\centering
\caption[Training-code per-cell medians and 95\% CIs]{Per-cell median composite score $S_{\text{tc}}$ with 95\% bootstrap confidence intervals for the training-code benchmark (percentile bootstrap over the ten seeds). Beams2D spans all seven tier variants; Photonics2D covers the plumbing (T2), architecture (T4), and open-synthesis (T6v2) tiers.}
\label{tab:stats_tc}
\small
\begin{tabular}{llccc}
\toprule
Model & Tier & $n$ & Median & 95\% CI \\
\midrule
\multicolumn{5}{l}{\emph{Beams2D}} \\
GPT-5-mini & T2 & 10 & $0.975$ & $[0.732,\,1.000]$ \\
GPT-5-mini & T3 & 10 & $0.976$ & $[0.560,\,1.000]$ \\
GPT-5-mini & T4 & 10 & $0.450$ & $[0.450,\,0.725]$ \\
GPT-5-mini & T4-func & 10 & $0.950$ & $[0.557,\,1.000]$ \\
GPT-5-mini & T5 & 10 & $0.450$ & $[0.450,\,1.000]$ \\
GPT-5-mini & T5-func & 10 & $0.400$ & $[0.400,\,0.425]$ \\
GPT-5-mini & T6v2 & 10 & $0.867$ & $[0.619,\,0.950]$ \\
Gemini-3-Flash & T2 & 10 & $1.000$ & $[0.776,\,1.000]$ \\
Gemini-3-Flash & T3 & 10 & $0.924$ & $[0.724,\,1.000]$ \\
Gemini-3-Flash & T4 & 10 & $0.734$ & $[0.590,\,1.000]$ \\
Gemini-3-Flash & T4-func & 10 & $0.765$ & $[0.564,\,1.000]$ \\
Gemini-3-Flash & T5 & 10 & $0.641$ & $[0.535,\,1.000]$ \\
Gemini-3-Flash & T5-func & 10 & $0.688$ & $[0.516,\,1.000]$ \\
Gemini-3-Flash & T6v2 & 10 & $0.950$ & $[0.860,\,0.950]$ \\
\midrule
\multicolumn{5}{l}{\emph{Photonics2D}} \\
GPT-5-mini & T2 & 10 & $0.969$ & $[0.934,\,0.996]$ \\
GPT-5-mini & T4 & 10 & $0.450$ & $[0.400,\,0.950]$ \\
GPT-5-mini & T6v2 & 10 & $0.950$ & $[0.950,\,0.950]$ \\
Gemini-3-Flash & T2 & 10 & $0.996$ & $[0.950,\,1.000]$ \\
Gemini-3-Flash & T4 & 10 & $1.000$ & $[0.450,\,1.000]$ \\
Gemini-3-Flash & T6v2 & 10 & $0.950$ & $[0.950,\,0.950]$ \\
\bottomrule
\end{tabular}
\end{table}

\subsection{Diffusion Model Results}
\label{app:diffusion}

\begin{figure*}[b]
    \centering
    \includegraphics[width=0.9\linewidth]{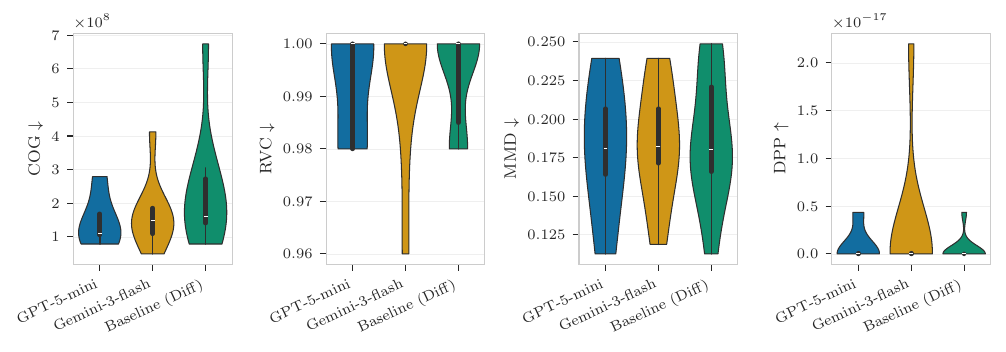}
    \caption[Offline diffusion model-quality metrics]{Offline model quality metrics (\acs{COG}, \acs{RVC}, \acs{MMD}, \acs{DPP}) for agent-trained diffusion models vs.\ EngiBench baselines. Arrows indicate desired direction. Values averaged across available seeds.}
    \label{fig:diff_baseline_grid}
\end{figure*}

Figure~\ref{fig:diff_baseline_grid} shows the offline model quality metrics for agent-trained diffusion models, analogous to the \acs{cGAN} results in Figure~\ref{fig:hpc_baseline_grid}. Agent-trained diffusion models achieve comparable values to the EngiBench baselines.

\subsection{Photonics2D}
\label{app:photonics}

\subsubsection{Workflow Benchmark}

\begin{figure*}[t]
    \centering
    \includegraphics[width=0.6\linewidth]{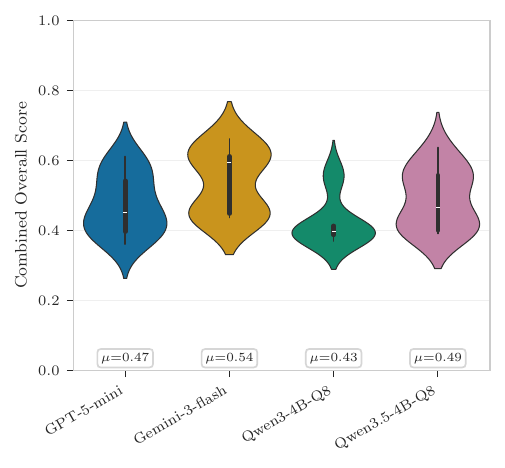}
    \caption[Photonics2D \textsc{W-Cond} score distributions]{Combined overall score distributions for Photonics2D \textsc{W-Cond}. The distributions have lower medians than the corresponding Beams2D results in Figure~\ref{fig:co_violin}.}
    \label{fig:co_violin_wcond_photonics}
\end{figure*}

Table~\ref{tab:tool_usage_photonics} and Figure~\ref{fig:co_violin_wcond_photonics} provide additional Photonics2D \textsc{W-Cond} results for those discussed in Section~\ref{sec:results}.

\begin{table}[t]
\centering
\caption[Tool-calling counts on Photonics2D]{Tool-calling counts for the Photonics2D \textsc{W-Cond} prompt style: average calls per tool per sample (mean $\pm$ std). Cell shading encodes call frequency using a linear scale from the smallest to the largest average in the table.}
\label{tab:tool_usage_photonics}
\scriptsize
\setlength{\tabcolsep}{4pt}
\begin{tabular*}{\columnwidth}{@{\extracolsep{\fill}}lcccc@{}}
\toprule
 & \multicolumn{4}{c}{\textsc{W-Cond}} \\
\cmidrule(lr){2-5}
\textbf{Tool} & GPT-5-mini & Gemini-3-Flash & Qwen3-4B-Q8 & Qwen3.5-4B-Q8 \\
\midrule
\texttt{optimize\_design} & \cellcolor{toolshade!57}1.0\tiny{$\pm$0.0} & \cellcolor{toolshade!57}1.0\tiny{$\pm$0.0} & \cellcolor{toolshade!57}1.0\tiny{$\pm$0.0} & \cellcolor{toolshade!57}1.0\tiny{$\pm$0.0} \\
\texttt{simulate\_design} & \cellcolor{toolshade!57}1.0\tiny{$\pm$0.0} & \cellcolor{toolshade!57}1.0\tiny{$\pm$0.0} & \cellcolor{toolshade!57}1.0\tiny{$\pm$0.0} & \cellcolor{toolshade!57}1.0\tiny{$\pm$0.0} \\
\texttt{convert\_design\_to\_stl} & \cellcolor{toolshade!65}1.1\tiny{$\pm$0.3} & \cellcolor{toolshade!57}1.0\tiny{$\pm$0.0} & \cellcolor{toolshade!31}0.5\tiny{$\pm$0.5} & \cellcolor{toolshade!57}1.0\tiny{$\pm$0.0} \\
\texttt{create\_problem} & \cellcolor{toolshade!34}0.6\tiny{$\pm$0.5} & \cellcolor{toolshade!0}0.0\tiny{$\pm$0.0} & \cellcolor{toolshade!57}1.0\tiny{$\pm$0.0} & \cellcolor{toolshade!42}0.7\tiny{$\pm$0.4} \\
\texttt{render\_design} & \cellcolor{toolshade!15}0.3\tiny{$\pm$0.4} & \cellcolor{toolshade!0}0.0\tiny{$\pm$0.0} & \cellcolor{toolshade!4}0.1\tiny{$\pm$0.2} & \cellcolor{toolshade!0}0.0\tiny{$\pm$0.0} \\
\bottomrule
\end{tabular*}
\end{table}

\subsubsection{Training-code Benchmark}
\begin{figure*}[t]
    \centering
    \includegraphics[width=\linewidth]{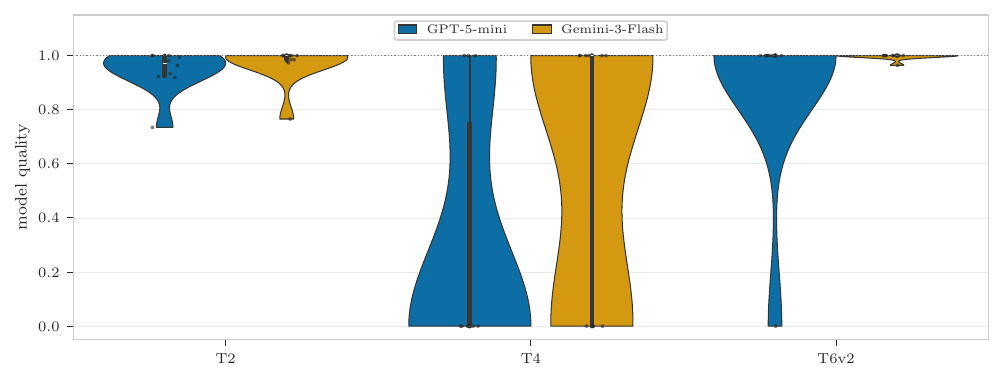}
    \caption[Per-seed model quality by tier on Photonics2D]{Per-seed model-quality scores by training-code tier and model backend on Photonics2D. Violin plots summarize ten seeds; boxes show the median and interquartile range, and points show individual seeds. Zero-valued T4 trials indicate checkpoints that the canonical evaluator could not load.}
    \label{fig:tc_score_grid_photonics2d}
\end{figure*}

Figure~\ref{fig:tc_score_grid_photonics2d} shows the per-seed graded model quality by tier for the training-code benchmark on Photonics2D (Section~\ref{sec:tc_results}).

\begin{figure*}[t]
\centering
\includegraphics[width=0.8\linewidth]{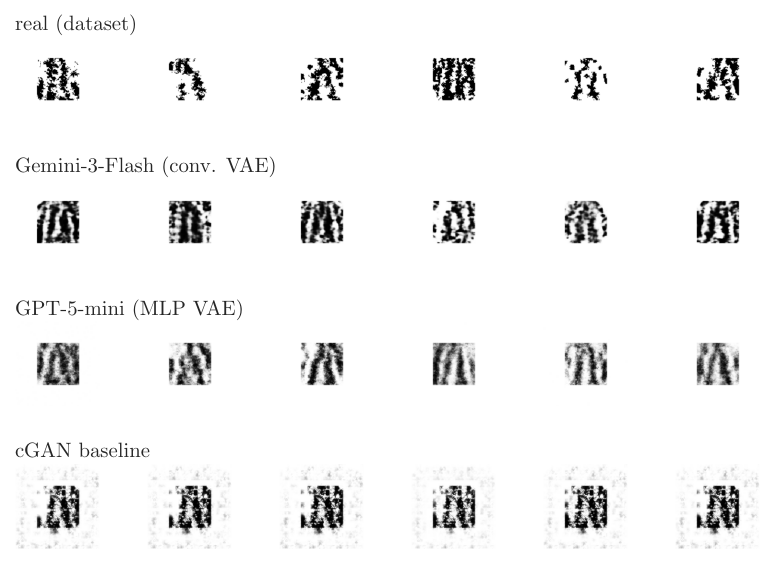}
\caption[Photonics2D designs from T6v2 generators]{Photonics2D designs for seed~9. Columns correspond to fixed conditioning vectors containing the wavelengths and blur radius. Rows show the dataset design, the Gemini-3-Flash convolutional VAE, the GPT-5-mini fully connected VAE, and the reference \acs{cGAN}. Designs are shown without thresholding.}
\label{fig:tc_novel_designs_photonics2d}
\end{figure*}

Figure~\ref{fig:tc_novel_designs_photonics2d} shows designs sampled from the open-synthesis (T6v2) generators authored by the two backends for Photonics2D. This figure is the cross-problem counterpart of the Beams2D results in Figure~\ref{fig:tc_novel_designs}.

For the illustrated seed, both agent-authored generators produce near-binary patterns that vary across the six conditioning vectors. The reference \acs{cGAN} produces visually similar outputs across these conditions. This example is illustrative; Section~\ref{sec:tc_results} reports the corresponding quantitative results across ten seeds.

\subsubsection{Cross-Problem Portability}
\label{app:portability}
We evaluate cross-problem portability using the trained T6v2 scripts. For each trial, we recover the Weave-logged authoring call whose model passes strict loading of that trial's checkpoint. This produces 20 scripts per source problem: ten from each model backend. T2 and T4 are excluded because they complete regions of a shared problem-independent program skeleton.

When executed unchanged on the other problem, 12 of the 20 Beams2D scripts run successfully on Photonics2D, and 14 of the 20 Photonics2D scripts run successfully on Beams2D. Every failure results from a dimension fixed for the source problem: a flattened feature dimension of $9{,}216$ where Photonics2D requires $28{,}800$ in one direction, or a feature-map width of $48$ where Beams2D requires $50$ in the other.

Portability differs by model backend. All 20 GPT-5-mini scripts transfer successfully because their fully connected architectures derive their dimensions from the input design shape. All transfer failures occur in Gemini-3-Flash scripts whose convolutional architectures contain spatial dimensions fixed for the source problem.

\section{Raw Benchmark Scores}
\label{app:raw_scores}

Tables~\ref{tab:raw_scores_beams2d} and~\ref{tab:raw_scores_photonics2d} report all individual sub-scores per model and prompt style, enabling readers to recompute composite scores under alternative weighting schemes.

Tables~\ref{tab:raw_scores_training_code_beams2d} and~\ref{tab:raw_scores_training_code_photonics2d} do the same for the training-code benchmark, reporting the six weighted components of $S_{\mathrm{tc}}$ separately alongside the composite. Any alternative weighting of those components can therefore be evaluated without re-running the grid.

\begin{landscape}
\begin{table*}[t]
\centering
\caption[Raw per-metric scores on Beams2D.]{Raw per-metric scores (mean $\pm$ std) for all models and prompt styles on Beams2D. Most cells aggregate 15 runs (3 seeds $\times$ 5 samples); a few Qwen3.5-4B configurations have 13--14 runs due to JSON parsing errors. \textbf{Bold} = best model per metric within each prompt style. Metric abbreviations are defined in Appendix~A of the main paper. The WT column is identically $0.00$ across all models because our threshold-based STL extraction produces non-manifold meshes; this reflects a limitation of the post-processing pipeline rather than agent capability.}
\label{tab:raw_scores_beams2d}
\footnotesize
\setlength{\tabcolsep}{3pt}
\begin{tabular}{@{}llrrrrrrrrrr@{}}
\toprule
\textbf{Style} & \textbf{Model} & \textbf{IoU\,$\uparrow$} & \textbf{PA\,$\uparrow$} & \textbf{Obj\,$\uparrow$} & \textbf{Constr\,$\uparrow$} & \textbf{Conn\,$\uparrow$} & \textbf{WT\,$\uparrow$} & \textbf{Tool Eff.\,$\uparrow$} & \textbf{TC\,$\uparrow$} & \textbf{DQ\,$\uparrow$} & \textbf{CO\,$\uparrow$} \\
\midrule
\textsc{Full} & GPT-5-mini & \textbf{0.40}\scriptsize{$\pm$0.15} & \textbf{0.73}\scriptsize{$\pm$0.08} & \textbf{0.52}\scriptsize{$\pm$0.43} & 1.00\scriptsize{$\pm$0.00} & 0.67\scriptsize{$\pm$0.47} & 0.00\scriptsize{$\pm$0.00} & 0.72\scriptsize{$\pm$0.12} & \textbf{1.00}\scriptsize{$\pm$0.00} & \textbf{0.54}\scriptsize{$\pm$0.10} & 0.65\scriptsize{$\pm$0.07} \\
 & Gemini-3-Flash & 0.39\scriptsize{$\pm$0.15} & 0.72\scriptsize{$\pm$0.08} & 0.49\scriptsize{$\pm$0.45} & 1.00\scriptsize{$\pm$0.00} & \textbf{0.73}\scriptsize{$\pm$0.44} & --- & \textbf{0.98}\scriptsize{$\pm$0.08} & 0.93\scriptsize{$\pm$0.25} & \textbf{0.54}\scriptsize{$\pm$0.09} & \textbf{0.69}\scriptsize{$\pm$0.10} \\
 & Qwen3-4B & \textbf{0.40}\scriptsize{$\pm$0.15} & \textbf{0.73}\scriptsize{$\pm$0.08} & 0.16\scriptsize{$\pm$0.33} & 1.00\scriptsize{$\pm$0.00} & 0.67\scriptsize{$\pm$0.47} & --- & 0.67\scriptsize{$\pm$0.00} & 0.00\scriptsize{$\pm$0.00} & 0.49\scriptsize{$\pm$0.10} & 0.45\scriptsize{$\pm$0.06} \\
 & Qwen3.5-4B & \textbf{0.40}\scriptsize{$\pm$0.15} & \textbf{0.73}\scriptsize{$\pm$0.08} & 0.37\scriptsize{$\pm$0.46} & 1.00\scriptsize{$\pm$0.00} & 0.67\scriptsize{$\pm$0.47} & --- & 0.72\scriptsize{$\pm$0.05} & 0.73\scriptsize{$\pm$0.44} & 0.52\scriptsize{$\pm$0.09} & 0.59\scriptsize{$\pm$0.11} \\
\midrule
\textsc{Natural} & GPT-5-mini & 0.39\scriptsize{$\pm$0.14} & 0.73\scriptsize{$\pm$0.05} & 0.04\scriptsize{$\pm$0.06} & 0.09\scriptsize{$\pm$0.12} & 0.75\scriptsize{$\pm$0.43} & 0.00\scriptsize{$\pm$0.00} & 0.75\scriptsize{$\pm$0.41} & 0.87\scriptsize{$\pm$0.34} & 0.37\scriptsize{$\pm$0.11} & 0.82\scriptsize{$\pm$0.30} \\
 & Gemini-3-Flash & 0.34\scriptsize{$\pm$0.07} & 0.69\scriptsize{$\pm$0.05} & 0.10\scriptsize{$\pm$0.14} & \textbf{0.13}\scriptsize{$\pm$0.11} & 0.67\scriptsize{$\pm$0.47} & --- & \textbf{0.81}\scriptsize{$\pm$0.33} & \textbf{1.00}\scriptsize{$\pm$0.00} & 0.35\scriptsize{$\pm$0.07} & \textbf{0.88}\scriptsize{$\pm$0.23} \\
 & Qwen3-4B & 0.46\scriptsize{$\pm$0.18} & 0.72\scriptsize{$\pm$0.12} & \textbf{0.24}\scriptsize{$\pm$0.32} & 0.05\scriptsize{$\pm$0.10} & \textbf{1.00}\scriptsize{$\pm$0.00} & --- & 0.00\scriptsize{$\pm$0.00} & 0.00\scriptsize{$\pm$0.00} & \textbf{0.44}\scriptsize{$\pm$0.07} & 0.29\scriptsize{$\pm$0.05} \\
 & Qwen3.5-4B & \textbf{0.47}\scriptsize{$\pm$0.13} & \textbf{0.74}\scriptsize{$\pm$0.09} & 0.13\scriptsize{$\pm$0.27} & 0.06\scriptsize{$\pm$0.11} & 0.91\scriptsize{$\pm$0.29} & --- & 0.28\scriptsize{$\pm$0.44} & 0.33\scriptsize{$\pm$0.47} & 0.42\scriptsize{$\pm$0.07} & 0.48\scriptsize{$\pm$0.32} \\
\midrule
\textsc{W-Rand} & GPT-5-mini & \textbf{0.40}\scriptsize{$\pm$0.15} & \textbf{0.73}\scriptsize{$\pm$0.08} & \textbf{0.52}\scriptsize{$\pm$0.43} & 1.00\scriptsize{$\pm$0.00} & 0.67\scriptsize{$\pm$0.47} & 0.00\scriptsize{$\pm$0.00} & 0.77\scriptsize{$\pm$0.17} & 1.00\scriptsize{$\pm$0.00} & 0.54\scriptsize{$\pm$0.10} & 0.66\scriptsize{$\pm$0.07} \\
 & Gemini-3-Flash & \textbf{0.40}\scriptsize{$\pm$0.11} & \textbf{0.73}\scriptsize{$\pm$0.06} & 0.35\scriptsize{$\pm$0.44} & 1.00\scriptsize{$\pm$0.00} & 0.73\scriptsize{$\pm$0.44} & 0.00\scriptsize{$\pm$0.00} & 0.80\scriptsize{$\pm$0.18} & 1.00\scriptsize{$\pm$0.00} & 0.52\scriptsize{$\pm$0.09} & 0.65\scriptsize{$\pm$0.09} \\
 & Qwen3-4B & 0.39\scriptsize{$\pm$0.15} & 0.72\scriptsize{$\pm$0.08} & 0.48\scriptsize{$\pm$0.45} & 1.00\scriptsize{$\pm$0.00} & \textbf{0.87}\scriptsize{$\pm$0.34} & 0.00\scriptsize{$\pm$0.00} & 0.63\scriptsize{$\pm$0.06} & 1.00\scriptsize{$\pm$0.00} & \textbf{0.55}\scriptsize{$\pm$0.11} & 0.64\scriptsize{$\pm$0.07} \\
 & Qwen3.5-4B & \textbf{0.40}\scriptsize{$\pm$0.16} & \textbf{0.73}\scriptsize{$\pm$0.08} & 0.38\scriptsize{$\pm$0.44} & 1.00\scriptsize{$\pm$0.00} & 0.71\scriptsize{$\pm$0.45} & 0.00\scriptsize{$\pm$0.00} & \textbf{1.00}\scriptsize{$\pm$0.00} & 1.00\scriptsize{$\pm$0.00} & 0.53\scriptsize{$\pm$0.08} & \textbf{0.69}\scriptsize{$\pm$0.06} \\
\midrule
\textsc{W-Derived} & GPT-5-mini & 0.40\scriptsize{$\pm$0.15} & 0.73\scriptsize{$\pm$0.08} & \textbf{0.52}\scriptsize{$\pm$0.43} & \textbf{1.00}\scriptsize{$\pm$0.00} & 0.67\scriptsize{$\pm$0.47} & 0.00\scriptsize{$\pm$0.00} & 0.76\scriptsize{$\pm$0.14} & \textbf{1.00}\scriptsize{$\pm$0.00} & \textbf{0.54}\scriptsize{$\pm$0.10} & 0.65\scriptsize{$\pm$0.07} \\
 & Gemini-3-Flash & \textbf{0.41}\scriptsize{$\pm$0.13} & \textbf{0.74}\scriptsize{$\pm$0.07} & 0.36\scriptsize{$\pm$0.43} & 0.93\scriptsize{$\pm$0.25} & 0.53\scriptsize{$\pm$0.50} & 0.00\scriptsize{$\pm$0.00} & 0.81\scriptsize{$\pm$0.14} & \textbf{1.00}\scriptsize{$\pm$0.00} & 0.50\scriptsize{$\pm$0.11} & 0.64\scriptsize{$\pm$0.07} \\
 & Qwen3-4B & 0.40\scriptsize{$\pm$0.15} & 0.73\scriptsize{$\pm$0.08} & 0.28\scriptsize{$\pm$0.40} & \textbf{1.00}\scriptsize{$\pm$0.00} & \textbf{0.73}\scriptsize{$\pm$0.44} & 0.00\scriptsize{$\pm$0.00} & 0.64\scriptsize{$\pm$0.07} & 0.47\scriptsize{$\pm$0.50} & 0.51\scriptsize{$\pm$0.08} & 0.53\scriptsize{$\pm$0.08} \\
 & Qwen3.5-4B & \textbf{0.41}\scriptsize{$\pm$0.16} & 0.73\scriptsize{$\pm$0.09} & 0.43\scriptsize{$\pm$0.46} & \textbf{1.00}\scriptsize{$\pm$0.00} & 0.62\scriptsize{$\pm$0.49} & 0.00\scriptsize{$\pm$0.00} & \textbf{0.95}\scriptsize{$\pm$0.18} & 0.85\scriptsize{$\pm$0.36} & 0.52\scriptsize{$\pm$0.09} & \textbf{0.66}\scriptsize{$\pm$0.10} \\
\midrule
\textsc{W-Distract} & GPT-5-mini & 0.40\scriptsize{$\pm$0.15} & 0.73\scriptsize{$\pm$0.08} & \textbf{0.52}\scriptsize{$\pm$0.43} & 1.00\scriptsize{$\pm$0.00} & 0.67\scriptsize{$\pm$0.47} & 0.00\scriptsize{$\pm$0.00} & 0.63\scriptsize{$\pm$0.08} & \textbf{1.00}\scriptsize{$\pm$0.00} & \textbf{0.54}\scriptsize{$\pm$0.10} & 0.63\scriptsize{$\pm$0.06} \\
 & Gemini-3-Flash & \textbf{0.41}\scriptsize{$\pm$0.14} & \textbf{0.74}\scriptsize{$\pm$0.07} & 0.40\scriptsize{$\pm$0.43} & 1.00\scriptsize{$\pm$0.00} & 0.53\scriptsize{$\pm$0.50} & 0.00\scriptsize{$\pm$0.00} & 0.72\scriptsize{$\pm$0.06} & \textbf{1.00}\scriptsize{$\pm$0.00} & 0.51\scriptsize{$\pm$0.11} & 0.63\scriptsize{$\pm$0.08} \\
 & Qwen3-4B & 0.39\scriptsize{$\pm$0.15} & 0.73\scriptsize{$\pm$0.08} & 0.41\scriptsize{$\pm$0.44} & 1.00\scriptsize{$\pm$0.00} & \textbf{0.73}\scriptsize{$\pm$0.44} & 0.00\scriptsize{$\pm$0.00} & 0.60\scriptsize{$\pm$0.00} & \textbf{1.00}\scriptsize{$\pm$0.00} & 0.53\scriptsize{$\pm$0.10} & 0.61\scriptsize{$\pm$0.06} \\
 & Qwen3.5-4B & 0.40\scriptsize{$\pm$0.15} & 0.73\scriptsize{$\pm$0.08} & 0.42\scriptsize{$\pm$0.45} & 1.00\scriptsize{$\pm$0.00} & 0.67\scriptsize{$\pm$0.47} & 0.00\scriptsize{$\pm$0.00} & \textbf{0.88}\scriptsize{$\pm$0.12} & 0.93\scriptsize{$\pm$0.25} & 0.53\scriptsize{$\pm$0.10} & \textbf{0.66}\scriptsize{$\pm$0.09} \\
\midrule
\textsc{W-Cond} & GPT-5-mini & \textbf{0.40}\scriptsize{$\pm$0.15} & \textbf{0.73}\scriptsize{$\pm$0.08} & \textbf{0.52}\scriptsize{$\pm$0.43} & \textbf{1.00}\scriptsize{$\pm$0.00} & 0.67\scriptsize{$\pm$0.47} & 0.00\scriptsize{$\pm$0.00} & 0.84\scriptsize{$\pm$0.16} & \textbf{0.93}\scriptsize{$\pm$0.25} & 0.54\scriptsize{$\pm$0.10} & \textbf{0.66}\scriptsize{$\pm$0.10} \\
 & Gemini-3-Flash & \textbf{0.40}\scriptsize{$\pm$0.16} & \textbf{0.73}\scriptsize{$\pm$0.08} & 0.40\scriptsize{$\pm$0.43} & 0.93\scriptsize{$\pm$0.25} & \textbf{0.87}\scriptsize{$\pm$0.34} & 0.00\scriptsize{$\pm$0.00} & 0.86\scriptsize{$\pm$0.17} & 0.87\scriptsize{$\pm$0.34} & 0.54\scriptsize{$\pm$0.10} & 0.65\scriptsize{$\pm$0.10} \\
 & Qwen3-4B & 0.39\scriptsize{$\pm$0.15} & 0.72\scriptsize{$\pm$0.09} & 0.48\scriptsize{$\pm$0.44} & \textbf{1.00}\scriptsize{$\pm$0.00} & \textbf{0.87}\scriptsize{$\pm$0.34} & 0.00\scriptsize{$\pm$0.00} & 0.76\scriptsize{$\pm$0.15} & 0.40\scriptsize{$\pm$0.49} & \textbf{0.55}\scriptsize{$\pm$0.10} & 0.57\scriptsize{$\pm$0.12} \\
 & Qwen3.5-4B & \textbf{0.40}\scriptsize{$\pm$0.15} & \textbf{0.73}\scriptsize{$\pm$0.08} & 0.33\scriptsize{$\pm$0.41} & \textbf{1.00}\scriptsize{$\pm$0.00} & 0.67\scriptsize{$\pm$0.47} & 0.00\scriptsize{$\pm$0.00} & \textbf{1.00}\scriptsize{$\pm$0.00} & 0.60\scriptsize{$\pm$0.49} & 0.51\scriptsize{$\pm$0.10} & 0.62\scriptsize{$\pm$0.13} \\
\midrule
\textsc{W-Multi} & GPT-5-mini & 0.40\scriptsize{$\pm$0.15} & 0.73\scriptsize{$\pm$0.08} & \textbf{0.52}\scriptsize{$\pm$0.43} & 1.00\scriptsize{$\pm$0.00} & 0.67\scriptsize{$\pm$0.47} & 0.00\scriptsize{$\pm$0.00} & 0.87\scriptsize{$\pm$0.13} & 0.93\scriptsize{$\pm$0.25} & \textbf{0.54}\scriptsize{$\pm$0.10} & 0.66\scriptsize{$\pm$0.09} \\
 & Gemini-3-Flash & 0.40\scriptsize{$\pm$0.15} & 0.73\scriptsize{$\pm$0.08} & \textbf{0.52}\scriptsize{$\pm$0.43} & 1.00\scriptsize{$\pm$0.00} & 0.67\scriptsize{$\pm$0.47} & 0.00\scriptsize{$\pm$0.00} & 0.99\scriptsize{$\pm$0.05} & \textbf{1.00}\scriptsize{$\pm$0.00} & \textbf{0.54}\scriptsize{$\pm$0.10} & \textbf{0.70}\scriptsize{$\pm$0.07} \\
 & Qwen3-4B & 0.40\scriptsize{$\pm$0.15} & 0.73\scriptsize{$\pm$0.08} & 0.34\scriptsize{$\pm$0.41} & 1.00\scriptsize{$\pm$0.00} & \textbf{0.80}\scriptsize{$\pm$0.40} & 0.00\scriptsize{$\pm$0.00} & 0.78\scriptsize{$\pm$0.05} & \textbf{1.00}\scriptsize{$\pm$0.00} & 0.53\scriptsize{$\pm$0.11} & 0.65\scriptsize{$\pm$0.08} \\
 & Qwen3.5-4B & \textbf{0.42}\scriptsize{$\pm$0.14} & \textbf{0.74}\scriptsize{$\pm$0.07} & 0.42\scriptsize{$\pm$0.42} & 1.00\scriptsize{$\pm$0.00} & 0.64\scriptsize{$\pm$0.48} & 0.00\scriptsize{$\pm$0.00} & \textbf{1.00}\scriptsize{$\pm$0.00} & \textbf{1.00}\scriptsize{$\pm$0.00} & 0.53\scriptsize{$\pm$0.10} & 0.69\scriptsize{$\pm$0.06} \\
\bottomrule
\end{tabular}
\end{table*}

\begin{table*}[t]
\centering
\caption[Raw per-metric scores on Photonics2D.]{Raw per-metric scores (mean $\pm$ std) for all models and prompt styles on Photonics2D. Most cells aggregate 15 runs (3 seeds $\times$ 5 samples); a few Qwen3.5-4B configurations have 13--14 runs due to JSON parsing errors. \textbf{Bold} = best model per metric within each prompt style. Metric abbreviations are defined in Appendix~A of the main paper. The WT column is identically $0.00$ across all models because our threshold-based STL extraction produces non-manifold meshes; this reflects a limitation of the post-processing pipeline rather than agent capability.}
\label{tab:raw_scores_photonics2d}
\footnotesize
\setlength{\tabcolsep}{3pt}
\begin{tabular}{@{}llrrrrrrrrrr@{}}
\toprule
\textbf{Style} & \textbf{Model} & \textbf{IoU\,$\uparrow$} & \textbf{PA\,$\uparrow$} & \textbf{Obj\,$\uparrow$} & \textbf{Constr\,$\uparrow$} & \textbf{Conn\,$\uparrow$} & \textbf{WT\,$\uparrow$} & \textbf{Tool Eff.\,$\uparrow$} & \textbf{TC\,$\uparrow$} & \textbf{DQ\,$\uparrow$} & \textbf{CO\,$\uparrow$} \\
\midrule
\textsc{W-Rand} & GPT-5-mini & \textbf{0.32}\scriptsize{$\pm$0.05} & 0.89\scriptsize{$\pm$0.01} & 0.00\scriptsize{$\pm$0.00} & 1.00\scriptsize{$\pm$0.00} & 0.00\scriptsize{$\pm$0.00} & 0.00\scriptsize{$\pm$0.00} & 0.63\scriptsize{$\pm$0.06} & 1.00\scriptsize{$\pm$0.00} & 0.39\scriptsize{$\pm$0.02} & 0.53\scriptsize{$\pm$0.01} \\
 & Gemini-3-Flash & 0.31\scriptsize{$\pm$0.05} & 0.89\scriptsize{$\pm$0.01} & 0.01\scriptsize{$\pm$0.03} & 1.00\scriptsize{$\pm$0.00} & 0.00\scriptsize{$\pm$0.00} & 0.00\scriptsize{$\pm$0.00} & \textbf{0.97}\scriptsize{$\pm$0.08} & 1.00\scriptsize{$\pm$0.00} & 0.39\scriptsize{$\pm$0.02} & \textbf{0.60}\scriptsize{$\pm$0.02} \\
 & Qwen3-4B & 0.31\scriptsize{$\pm$0.05} & 0.89\scriptsize{$\pm$0.01} & \textbf{0.04}\scriptsize{$\pm$0.14} & 1.00\scriptsize{$\pm$0.00} & 0.00\scriptsize{$\pm$0.00} & 0.00\scriptsize{$\pm$0.00} & 0.75\scriptsize{$\pm$0.00} & 1.00\scriptsize{$\pm$0.00} & 0.39\scriptsize{$\pm$0.03} & 0.56\scriptsize{$\pm$0.02} \\
 & Qwen3.5-4B & 0.31\scriptsize{$\pm$0.06} & 0.89\scriptsize{$\pm$0.01} & 0.03\scriptsize{$\pm$0.10} & 1.00\scriptsize{$\pm$0.00} & 0.00\scriptsize{$\pm$0.00} & 0.00\scriptsize{$\pm$0.00} & 0.73\scriptsize{$\pm$0.05} & 1.00\scriptsize{$\pm$0.00} & 0.39\scriptsize{$\pm$0.02} & 0.55\scriptsize{$\pm$0.01} \\
\midrule
\textsc{W-Distract} & GPT-5-mini & 0.31\scriptsize{$\pm$0.05} & 0.89\scriptsize{$\pm$0.01} & 0.02\scriptsize{$\pm$0.06} & 1.00\scriptsize{$\pm$0.00} & 0.00\scriptsize{$\pm$0.00} & 0.00\scriptsize{$\pm$0.00} & 0.59\scriptsize{$\pm$0.07} & 1.00\scriptsize{$\pm$0.00} & 0.39\scriptsize{$\pm$0.02} & 0.52\scriptsize{$\pm$0.02} \\
 & Gemini-3-Flash & \textbf{0.32}\scriptsize{$\pm$0.05} & 0.89\scriptsize{$\pm$0.01} & \textbf{0.05}\scriptsize{$\pm$0.15} & 1.00\scriptsize{$\pm$0.00} & 0.00\scriptsize{$\pm$0.00} & 0.00\scriptsize{$\pm$0.00} & 0.72\scriptsize{$\pm$0.06} & 1.00\scriptsize{$\pm$0.00} & \textbf{0.40}\scriptsize{$\pm$0.04} & 0.55\scriptsize{$\pm$0.03} \\
 & Qwen3-4B & 0.31\scriptsize{$\pm$0.05} & 0.89\scriptsize{$\pm$0.01} & 0.04\scriptsize{$\pm$0.14} & 1.00\scriptsize{$\pm$0.00} & 0.00\scriptsize{$\pm$0.00} & 0.00\scriptsize{$\pm$0.00} & 0.58\scriptsize{$\pm$0.04} & 1.00\scriptsize{$\pm$0.00} & 0.39\scriptsize{$\pm$0.03} & 0.52\scriptsize{$\pm$0.02} \\
 & Qwen3.5-4B & \textbf{0.32}\scriptsize{$\pm$0.06} & 0.89\scriptsize{$\pm$0.01} & \textbf{0.05}\scriptsize{$\pm$0.19} & 1.00\scriptsize{$\pm$0.00} & 0.00\scriptsize{$\pm$0.00} & 0.00\scriptsize{$\pm$0.00} & \textbf{0.82}\scriptsize{$\pm$0.16} & 1.00\scriptsize{$\pm$0.00} & \textbf{0.40}\scriptsize{$\pm$0.04} & \textbf{0.57}\scriptsize{$\pm$0.05} \\
\midrule
\textsc{W-Cond} & GPT-5-mini & \textbf{0.32}\scriptsize{$\pm$0.05} & 0.89\scriptsize{$\pm$0.01} & 0.06\scriptsize{$\pm$0.16} & 1.00\scriptsize{$\pm$0.00} & 0.00\scriptsize{$\pm$0.00} & 0.00\scriptsize{$\pm$0.00} & 0.78\scriptsize{$\pm$0.16} & 0.40\scriptsize{$\pm$0.49} & \textbf{0.40}\scriptsize{$\pm$0.03} & 0.47\scriptsize{$\pm$0.08} \\
 & Gemini-3-Flash & 0.31\scriptsize{$\pm$0.05} & 0.89\scriptsize{$\pm$0.01} & \textbf{0.08}\scriptsize{$\pm$0.17} & 1.00\scriptsize{$\pm$0.00} & 0.00\scriptsize{$\pm$0.00} & 0.00\scriptsize{$\pm$0.00} & \textbf{1.00}\scriptsize{$\pm$0.00} & \textbf{0.53}\scriptsize{$\pm$0.50} & \textbf{0.40}\scriptsize{$\pm$0.03} & \textbf{0.54}\scriptsize{$\pm$0.09} \\
 & Qwen3-4B & 0.31\scriptsize{$\pm$0.05} & 0.89\scriptsize{$\pm$0.01} & 0.04\scriptsize{$\pm$0.15} & 1.00\scriptsize{$\pm$0.00} & 0.00\scriptsize{$\pm$0.00} & 0.00\scriptsize{$\pm$0.00} & 0.70\scriptsize{$\pm$0.05} & 0.20\scriptsize{$\pm$0.40} & 0.39\scriptsize{$\pm$0.03} & 0.43\scriptsize{$\pm$0.07} \\
 & Qwen3.5-4B & \textbf{0.32}\scriptsize{$\pm$0.05} & 0.89\scriptsize{$\pm$0.01} & 0.04\scriptsize{$\pm$0.15} & 1.00\scriptsize{$\pm$0.00} & 0.00\scriptsize{$\pm$0.00} & 0.00\scriptsize{$\pm$0.00} & 0.82\scriptsize{$\pm$0.11} & 0.47\scriptsize{$\pm$0.50} & 0.39\scriptsize{$\pm$0.04} & 0.49\scriptsize{$\pm$0.08} \\
\bottomrule
\end{tabular}
\end{table*}

\begin{table*}[t]
\centering
\caption[Unweighted training-code components on Beams2D.]{Unweighted components of the training-code score on Beams2D (mean $\pm$ std over ten seeds per cell). Column headings give each term's weight in $S_{\mathrm{tc}}$; the final column is the composite under those weights. The components are reported separately so the composite can be recomputed under any alternative weighting. \textbf{Bold} = better model per component within each tier. Cells where model quality was unscorable are excluded from that column's mean.}
\label{tab:raw_scores_training_code_beams2d}
\footnotesize
\setlength{\tabcolsep}{3pt}
\begin{tabular}{@{}llrrrrrrr@{}}
\toprule
\textbf{Tier} & \textbf{Model} & \textbf{MQ (.55)\,$\uparrow$} & \textbf{Code (.15)\,$\uparrow$} & \textbf{Job (.10)\,$\uparrow$} & \textbf{Dry (.10)\,$\uparrow$} & \textbf{Syn (.05)\,$\uparrow$} & \textbf{SLURM (.05)\,$\uparrow$} & \textbf{$S_{\mathrm{tc}}$\,$\uparrow$} \\
\midrule
T2 (plumbing) & GPT-5-mini & 0.74\scriptsize{$\pm$0.35} & 1.00\scriptsize{$\pm$0.00} & 1.00\scriptsize{$\pm$0.00} & 1.00\scriptsize{$\pm$0.00} & 1.00\scriptsize{$\pm$0.00} & 0.90\scriptsize{$\pm$0.30} & 0.85\scriptsize{$\pm$0.19} \\
 & Gemini-3-Flash & \textbf{0.81}\scriptsize{$\pm$0.31} & 1.00\scriptsize{$\pm$0.00} & 1.00\scriptsize{$\pm$0.00} & 1.00\scriptsize{$\pm$0.00} & 1.00\scriptsize{$\pm$0.00} & 0.90\scriptsize{$\pm$0.30} & \textbf{0.89}\scriptsize{$\pm$0.18} \\
\midrule
T3 (training loop) & GPT-5-mini & 0.68\scriptsize{$\pm$0.39} & 1.00\scriptsize{$\pm$0.00} & 1.00\scriptsize{$\pm$0.00} & 1.00\scriptsize{$\pm$0.00} & 1.00\scriptsize{$\pm$0.00} & \textbf{1.00}\scriptsize{$\pm$0.00} & 0.82\scriptsize{$\pm$0.22} \\
 & Gemini-3-Flash & \textbf{0.74}\scriptsize{$\pm$0.32} & 1.00\scriptsize{$\pm$0.00} & 1.00\scriptsize{$\pm$0.00} & 1.00\scriptsize{$\pm$0.00} & 1.00\scriptsize{$\pm$0.00} & 0.90\scriptsize{$\pm$0.30} & \textbf{0.85}\scriptsize{$\pm$0.17} \\
\midrule
T4 (architecture) & GPT-5-mini & 0.21\scriptsize{$\pm$0.40} & 1.00\scriptsize{$\pm$0.00} & 0.90\scriptsize{$\pm$0.30} & 1.00\scriptsize{$\pm$0.00} & 1.00\scriptsize{$\pm$0.00} & \textbf{1.00}\scriptsize{$\pm$0.00} & 0.55\scriptsize{$\pm$0.23} \\
 & Gemini-3-Flash & \textbf{0.59}\scriptsize{$\pm$0.36} & 1.00\scriptsize{$\pm$0.00} & \textbf{1.00}\scriptsize{$\pm$0.00} & 1.00\scriptsize{$\pm$0.00} & 1.00\scriptsize{$\pm$0.00} & 0.90\scriptsize{$\pm$0.30} & \textbf{0.77}\scriptsize{$\pm$0.19} \\
\midrule
T4-func & GPT-5-mini & \textbf{0.74}\scriptsize{$\pm$0.40} & 1.00\scriptsize{$\pm$0.00} & 1.00\scriptsize{$\pm$0.00} & 0.90\scriptsize{$\pm$0.30} & 1.00\scriptsize{$\pm$0.00} & 0.30\scriptsize{$\pm$0.46} & \textbf{0.81}\scriptsize{$\pm$0.25} \\
 & Gemini-3-Flash & 0.60\scriptsize{$\pm$0.36} & 1.00\scriptsize{$\pm$0.00} & 1.00\scriptsize{$\pm$0.00} & \textbf{1.00}\scriptsize{$\pm$0.00} & 1.00\scriptsize{$\pm$0.00} & \textbf{0.70}\scriptsize{$\pm$0.46} & 0.77\scriptsize{$\pm$0.21} \\
\midrule
T5 (whole module) & GPT-5-mini & 0.40\scriptsize{$\pm$0.49} & 0.90\scriptsize{$\pm$0.30} & 1.00\scriptsize{$\pm$0.00} & 0.90\scriptsize{$\pm$0.30} & 1.00\scriptsize{$\pm$0.00} & \textbf{1.00}\scriptsize{$\pm$0.00} & 0.65\scriptsize{$\pm$0.25} \\
 & Gemini-3-Flash & \textbf{0.53}\scriptsize{$\pm$0.40} & \textbf{1.00}\scriptsize{$\pm$0.00} & 1.00\scriptsize{$\pm$0.00} & \textbf{1.00}\scriptsize{$\pm$0.00} & 1.00\scriptsize{$\pm$0.00} & 0.90\scriptsize{$\pm$0.30} & \textbf{0.74}\scriptsize{$\pm$0.22} \\
\midrule
T5-func & GPT-5-mini & 0.00\scriptsize{$\pm$0.00} & \textbf{1.00}\scriptsize{$\pm$0.00} & \textbf{1.00}\scriptsize{$\pm$0.00} & \textbf{1.00}\scriptsize{$\pm$0.00} & 1.00\scriptsize{$\pm$0.00} & 0.20\scriptsize{$\pm$0.40} & 0.41\scriptsize{$\pm$0.02} \\
 & Gemini-3-Flash & \textbf{0.54}\scriptsize{$\pm$0.41} & 0.90\scriptsize{$\pm$0.30} & 0.90\scriptsize{$\pm$0.30} & 0.90\scriptsize{$\pm$0.30} & 1.00\scriptsize{$\pm$0.00} & \textbf{0.70}\scriptsize{$\pm$0.46} & \textbf{0.70}\scriptsize{$\pm$0.30} \\
\midrule
T6v2 (open synth.) & GPT-5-mini & 0.70\scriptsize{$\pm$0.35} & 1.00\scriptsize{$\pm$0.00} & 1.00\scriptsize{$\pm$0.00} & 1.00\scriptsize{$\pm$0.00} & 1.00\scriptsize{$\pm$0.00} & \textbf{0.20}\scriptsize{$\pm$0.40} & 0.79\scriptsize{$\pm$0.20} \\
 & Gemini-3-Flash & \textbf{0.88}\scriptsize{$\pm$0.24} & 1.00\scriptsize{$\pm$0.00} & 1.00\scriptsize{$\pm$0.00} & 1.00\scriptsize{$\pm$0.00} & 1.00\scriptsize{$\pm$0.00} & 0.10\scriptsize{$\pm$0.30} & \textbf{0.89}\scriptsize{$\pm$0.14} \\
\bottomrule
\end{tabular}
\end{table*}

\begin{table*}[t]
\centering
\caption[Unweighted training-code components on Photonics2D.]{Unweighted components of the training-code score on Photonics2D (mean $\pm$ std over ten seeds per cell). Column headings give each term's weight in $S_{\mathrm{tc}}$; the final column is the composite under those weights. The components are reported separately so the composite can be recomputed under any alternative weighting. \textbf{Bold} = better model per component within each tier. Cells where model quality was unscorable are excluded from that column's mean.}
\label{tab:raw_scores_training_code_photonics2d}
\footnotesize
\setlength{\tabcolsep}{3pt}
\begin{tabular}{@{}llrrrrrrr@{}}
\toprule
\textbf{Tier} & \textbf{Model} & \textbf{MQ (.55)\,$\uparrow$} & \textbf{Code (.15)\,$\uparrow$} & \textbf{Job (.10)\,$\uparrow$} & \textbf{Dry (.10)\,$\uparrow$} & \textbf{Syn (.05)\,$\uparrow$} & \textbf{SLURM (.05)\,$\uparrow$} & \textbf{$S_{\mathrm{tc}}$\,$\uparrow$} \\
\midrule
T2 (plumbing) & GPT-5-mini & 0.94\scriptsize{$\pm$0.08} & 1.00\scriptsize{$\pm$0.00} & 1.00\scriptsize{$\pm$0.00} & 1.00\scriptsize{$\pm$0.00} & 1.00\scriptsize{$\pm$0.00} & 0.80\scriptsize{$\pm$0.40} & 0.96\scriptsize{$\pm$0.04} \\
 & Gemini-3-Flash & \textbf{0.97}\scriptsize{$\pm$0.07} & 1.00\scriptsize{$\pm$0.00} & 1.00\scriptsize{$\pm$0.00} & 1.00\scriptsize{$\pm$0.00} & 1.00\scriptsize{$\pm$0.00} & 0.80\scriptsize{$\pm$0.40} & \textbf{0.97}\scriptsize{$\pm$0.04} \\
\midrule
T4 (architecture) & GPT-5-mini & 0.30\scriptsize{$\pm$0.46} & 1.00\scriptsize{$\pm$0.00} & 1.00\scriptsize{$\pm$0.00} & 1.00\scriptsize{$\pm$0.00} & 1.00\scriptsize{$\pm$0.00} & 0.50\scriptsize{$\pm$0.50} & 0.59\scriptsize{$\pm$0.26} \\
 & Gemini-3-Flash & \textbf{0.60}\scriptsize{$\pm$0.49} & 1.00\scriptsize{$\pm$0.00} & 1.00\scriptsize{$\pm$0.00} & 1.00\scriptsize{$\pm$0.00} & 1.00\scriptsize{$\pm$0.00} & \textbf{0.90}\scriptsize{$\pm$0.30} & \textbf{0.78}\scriptsize{$\pm$0.28} \\
\midrule
T6v2 (open synth.) & GPT-5-mini & 0.90\scriptsize{$\pm$0.30} & 1.00\scriptsize{$\pm$0.00} & 1.00\scriptsize{$\pm$0.00} & 1.00\scriptsize{$\pm$0.00} & 1.00\scriptsize{$\pm$0.00} & 0.00\scriptsize{$\pm$0.00} & 0.89\scriptsize{$\pm$0.17} \\
 & Gemini-3-Flash & \textbf{1.00}\scriptsize{$\pm$0.01} & 1.00\scriptsize{$\pm$0.00} & 1.00\scriptsize{$\pm$0.00} & 1.00\scriptsize{$\pm$0.00} & 1.00\scriptsize{$\pm$0.00} & \textbf{0.10}\scriptsize{$\pm$0.30} & \textbf{0.95}\scriptsize{$\pm$0.02} \\
\bottomrule
\end{tabular}
\end{table*}

\end{landscape}
\end{document}